%% file: main.tex
\documentclass[unnumsec,webpdf,modern,large]{oup-authoring-template}
\usepackage{xurl}
\usepackage{booktabs}
\providecommand{\societylogo}{}
\graphicspath{{figures/}}
\journaltitle{PNAS Nexus}
\copyrightyear{2026}
\pubyear{2026}
\appnotes{Research Report}

\title[Compressed preference geometry in synthetic publics]{Language models compress human preference geometry while political structure remains prompt-sensitive}

\author[1,2,$\ast$]{Adolfo Ignacio Fuentes-Jofre}
\author[1,2,3,$\ast$]{Cristian Candia}

\address[1]{\orgdiv{Computational Research in Social Science Laboratory},
\orgname{School of Engineering, Universidad del Desarrollo},
\orgaddress{Las Condes, \postcode{7610658}, \country{Chile}}}
\address[2]{\orgdiv{Instituto de Data Science, Facultad de Ingenier\'ia},
\orgname{Universidad del Desarrollo},
\orgaddress{Las Condes, \postcode{7610658}, \country{Chile}}}
\address[3]{\orgdiv{Northwestern Institute on Complex Systems (NICO)},
\orgname{Northwestern University},
\orgaddress{Evanston, \state{IL}, \postcode{60208}, \country{USA}}}
\corresp[$\ast$]{Corresponding authors: Adolfo Ignacio Fuentes Jofr\'e, adolfo.fuentes.j@gmail.com; Cristian Candia, crcandiav@gmail.com.}

\abstract{
Large language models are increasingly used to construct synthetic publics. Prior work has documented errors in individual responses, population heterogeneity, and prompt robustness, but whether these failures reflect a common loss of fidelity or distinct properties of synthetic populations remains unclear. We tested this using matched pairwise policy choices from 1,991 Chilean, 843 French, and 434 Brazilian adults replayed through three open-weight model families. Across a 27-condition Chilean grid and cross-country factorial replications, no condition jointly reproduced individual choices, population heterogeneity, and political organization. The most consistent result was compression of the latent preference space: synthetic participant variance was only 0.4--10.5\% of the corresponding human reference in the primary comparisons and remained below human variance in all 270 matched refits. Political structure was much more sensitive to how choices were elicited. Reversing proposal position changed 12.3--31.2\% of decoded choices. Explicitly labeling the endpoints of the numeric left-right self-placement scale changed 20.5--28.1\% of choices and moved an independent proposal-level measure of left-right cleavage toward the human pattern in all three settings. Yet matched agreement improved by only 3.1--4.2 percentage points and participant geometry remained compressed. In Chile, the same intervention compressed the synthetic population further. These results separate political legibility from population fidelity: language models can recover salient political divisions while failing to preserve the breadth of human preference variation. Synthetic-public fidelity is therefore a property of the checkpoint, prompt, and setting, not of a checkpoint alone, and must be evaluated separately at individual, population, and political-structure levels.
}

\keywords[Abbreviations]{LLM, large language model; CI, confidence interval; UUID, universally unique identifier; ROC AUC, receiver-operating-characteristic area under the curve; CV, cross-validation; BT, Bradley--Terry; DYNAMAP, Dynamic System for Mapping of Users and Their Preferences Using Artificial Intelligence}

\otherabstract[Significance statement]{
Synthetic publics generated by language models may reproduce recognizable political patterns without preserving the diversity of the human populations they are intended to represent. Across matched policy choices from adults in Chile, France, and Brazil, synthetic participants occupied much narrower preference spaces than humans, while individual choices and political structure remained sensitive to proposal position and prompt wording. Explicitly labeling the ends of a political self-placement scale made political divisions more human-like but did not restore population heterogeneity. These findings show that political legibility and population fidelity are distinct properties. Evaluations of synthetic publics should therefore distinguish individual-choice agreement, population heterogeneity, positional stability, and political organization rather than treating any single measure as evidence of human-like representation.}

\begin{document}
\maketitle

\section{Introduction}

Large language models are increasingly used to answer surveys, forecast choices, and instantiate synthetic populations. Their value is often framed as \emph{algorithmic fidelity}: conditional model behavior may reproduce patterns observed among corresponding human groups \citep{Argyle2023}. Early work also showed that LLM-based participants can replicate some established behavioral findings \citep{Aher2023}, while political applications have used them to predict pairwise preferences or augment aggregate estimates \citep{GudinoRosero2024}. These successes motivate synthetic-public applications, but they do not by themselves establish population fidelity. A synthetic public may reproduce an aggregate mean while failing to preserve the choices of matched individuals, the heterogeneity among people, or the political divisions that organize those differences.

Existing evidence gives reason to separate these targets. Plausible averages can coexist with too little variation and altered covariate relationships \citep{Bisbee2024}. Identity-conditioned outputs can flatten or misrepresent demographic groups, and increased sampling temperature does not necessarily restore the missing variation \citep{Wang2025}. Other studies similarly report under-dispersion in voting preferences and contraction of persona-conditioned behavioral spaces \citep{Ball2025,Xiao2026,Bhattacharyya2026}. Even methods specialized to reproduce survey-response distributions improve some aggregate targets while continuing to struggle on unseen questions and countries \citep{Cao2025}. Population heterogeneity therefore requires direct validation rather than inference from average agreement or predictive accuracy.

A second challenge is specification dependence. Survey-style LLM responses can change under wording perturbations that do not produce corresponding changes in humans \citep{Tjuatja2024}, and closed-ended survey simulations differ materially across response-generation methods at both individual and subpopulation levels \citep{Ahnert2026}. For synthetic publics, this implies that substantive conclusions should not be inferred from one prompt realization alone. Position stability and prompt specification must be evaluated separately from human agreement: exchanging A/B placement should not alter the underlying proposal preference, while clarifying the meaning of a supplied political scale may reveal whether an apparent political structure depends on how that scale is expressed.

A third challenge is transfer across societies. Some evidence is encouraging: synthetic participants can approximate aggregate human responses in both WEIRD and non-WEIRD samples, although correspondence can be weaker outside the United States \citep{Shrestha2024}. Other work documents unevenly represented opinions, cultural tendencies, and cross-societal differences in LLM-simulated public opinion \citep{Santurkar2023,Atari2023,Brinkmann2023,Lu2025,Abeliuk2026}. Together with the difficulty of generalizing survey-response distributions to unseen countries \citep{Cao2025}, this means that fidelity established in one political and linguistic setting cannot simply be assumed to transfer to another.

Taken together, this literature implies a stricter validation problem than a single accuracy benchmark. A synthetic public should be evaluated on three separable requirements: whether it preserves matched behavior and population heterogeneity, whether its substantive structure is stable to prompt and position specification, and whether the same diagnostic pattern recurs across settings. We operationalize these requirements with seven criteria. \emph{Internal reconstruction} asks whether a spatial model explains an LLM's own repeated choices. \emph{Matched human agreement} asks whether the LLM selects the proposal chosen by the corresponding person. \emph{Self-placement recoverability} asks whether choice-derived coordinates encode supplied political self-placement. \emph{Position stability} asks whether changing A/B placement changes the decoded proposal choice. \emph{Participant geometry} asks whether synthetic people span the human latent space; \emph{proposal topology} asks whether relations among policy items are preserved; and \emph{political cleavage} asks whether the same proposals divide left- and right-identified groups. We estimate participant and proposal geometry using DYNAMAP (Dynamic System for Mapping of Users and Their Preferences Using Artificial Intelligence), a two-dimensional neural comparative-choice model that jointly embeds participants and policy proposals from repeated pairwise choices. Its architecture, estimation, alignment, and validation are specified below and in the SI Appendix (\textit{DYNAMAP architecture and validation}). Aggregate proposal structure and political cleavage are additionally evaluated with embedding-independent Bradley--Terry models.

The empirical design applies these criteria to the same matched decisions among confirmed adults. A Chilean grid replayed 39,820 choices---20 from each of 1,991 participants---with nine instruction-tuned checkpoints from Qwen3, Gemma 4, and Ministral 3 at three temperatures. Open-weight checkpoints were used so exact revisions and local inference settings could be fixed and replayed. We then selected three temperature-zero checkpoints that represented different grid maxima and crossed self-placement present versus ablated with original versus content-and-position-inverted A/B display. Checkpoint selection was outcome-informed; the $2\times2$ intervention was fixed before its new responses were generated. The same factorial was replicated in French and Brazilian Portuguese among 843 and 434 confirmed-adult participants, respectively, because a common data architecture captured pairwise policy choices around the 2022 presidential elections in two additional national and linguistic settings \citep{Navarrete2024}. The design therefore replays native-language policy instruments across three linguistic and national political settings; it provides cross-national replication, not representative coverage of languages or countries. Finally, after observing the bare-numeric results, we directly tested one candidate mechanism by explicitly labeling the endpoints of the self-placement scale for Qwen3-32B at temperature zero in both proposal orders. This outcome-informed probe tests whether clarifying the meaning of the supplied political scale alters political organization, without assuming that any such effect generalizes across checkpoints.

\section{Results}

\subsection{No model condition optimized all fidelity criteria}

All 27 grid conditions used the same 1,991 confirmed-adult participants, 20 decisions per participant, prompt template, and downstream estimators. Comparisons within the grid are therefore matched on people and proposals, although cross-family differences remain descriptive because checkpoint training, serving software, and numerical representation were not experimentally standardized.

\begin{figure*}[t]
\centering
\includegraphics[width=0.98\textwidth]{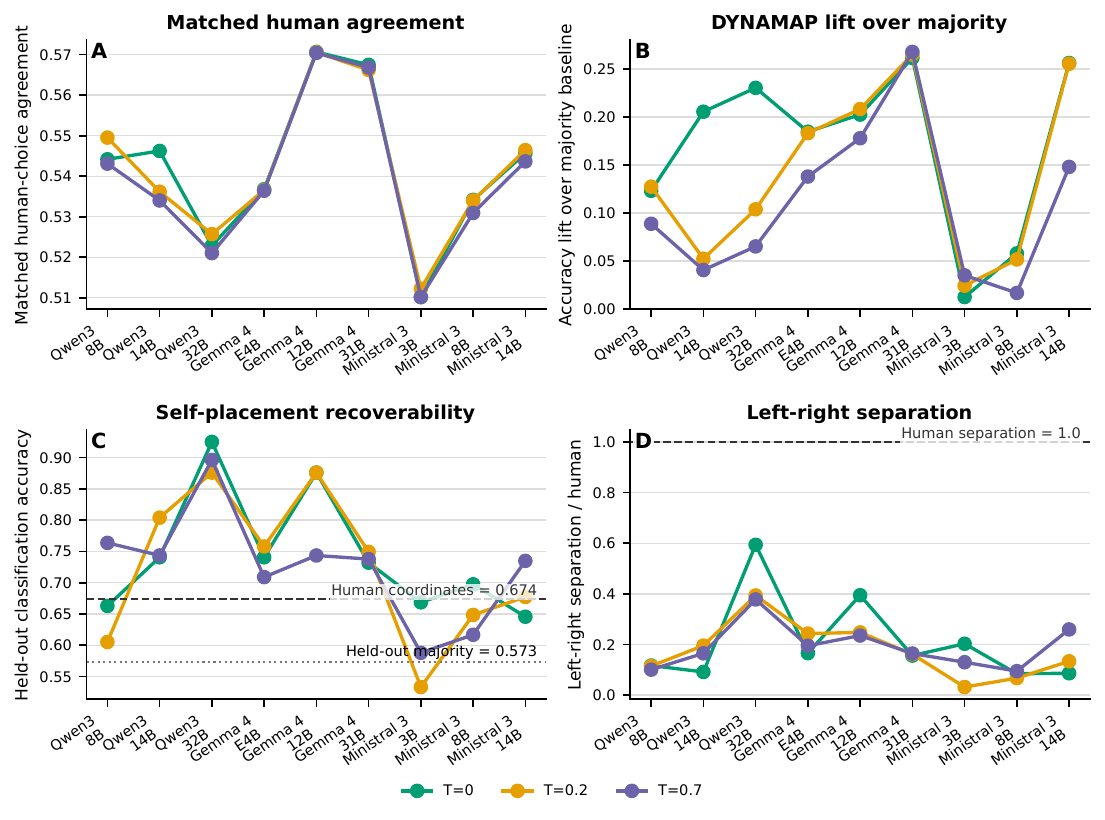}
\caption{Family, scale, and temperature move fidelity criteria differently. Each point summarizes one matched 39,820-decision confirmed-adult condition. Panel A compares model and human choices. Panel B subtracts the condition-specific validation-majority accuracy from internal DYNAMAP reconstruction. Panel C predicts supplied left--right self-placement from sociodemographics plus choice-derived coordinates; the dotted reference is the 0.573 majority rule in the fixed adult holdout. Panel D normalizes left--right separation to the independently fitted adult human space after proposal-anchor alignment. Geometry panels use the primary DYNAMAP initialization (seed 101). Dashed benchmarks are labeled as human references. Lines connect checkpoint sizes only to aid reading; they do not imply a continuous scale law.}
\label{fig:grid}
\figalttext[Four panels of metrics across 27 model conditions]{Lines for three temperatures vary nonmonotonically across nine family-size cells. The maxima for human agreement, DYNAMAP lift, political self-placement classification, and left-right separation occur in different conditions.}
\end{figure*}

Matched human agreement ranged from 0.510 to 0.571 (Fig.~\ref{fig:grid}; SI Appendix, Table~S2). Gemma 4-12B at $T=0$ had the highest agreement (0.571), but even this condition disagreed with the observed person in approximately 43\% of cells. Family averages were 0.558 for Gemma 4, 0.536 for Qwen3, and 0.530 for Ministral 3.

Internal reconstruction favored a different condition. Qwen3-14B at $T=0.7$ reached the highest raw DYNAMAP validation accuracy (0.850), but the condition-specific majority rule already achieved 0.809, leaving only about 0.040 lift. Gemma 4-31B at $T=0.7$ instead produced the largest lift (0.268) with lower raw accuracy. High reconstruction could therefore arise from concentrated responses rather than richer participant--proposal structure.

Recoverability of political self-placement produced another ordering. The human-coordinate benchmark classified held-out left (1--4) versus right (6--10) self-placement with 0.674 accuracy, compared with 0.516 from sociodemographics alone. The fixed adult holdout contained 199 majority-class participants among 347 (0.573). Qwen3-32B at $T=0$ reached 0.925 held-out and 0.912 CV-10 accuracy. These above-human scores reflect consistent use of an explicitly supplied persona cue, not emergent inference or deeper knowledge of individuals. Qwen3-32B at $T=0$, for example, recovered 59.4\% of the human left--right centroid separation but agreed with humans in only 52.3\% of cells. Gemma 4-12B at $T=0$ reached 57.1\% agreement but 39.4\% of human separation. Neither model scale nor temperature improved all criteria monotonically. No condition jointly maximized internal learnability, matched behavior, political organization, and human-like structure.

\subsection{Synthetic participants occupied a small fraction of the human space}

We aligned each independently fitted synthetic space to the human space using only matched proposal coordinates; political self-placement did not enter the transformation. At the primary initialization (seed 101), synthetic participant-coordinate variance was 0.4--6.2\% of the human value across the Chilean grid, while correlations between human and synthetic proposal distances were only 0.023--0.136. Contraction appeared in every family. Because proposal-anchor alignment removes arbitrary orientation and scale before comparing participant coordinates, it cannot be attributed merely to rotation of a latent space.

\begin{figure*}[t]
\centering
\includegraphics[width=0.98\textwidth]{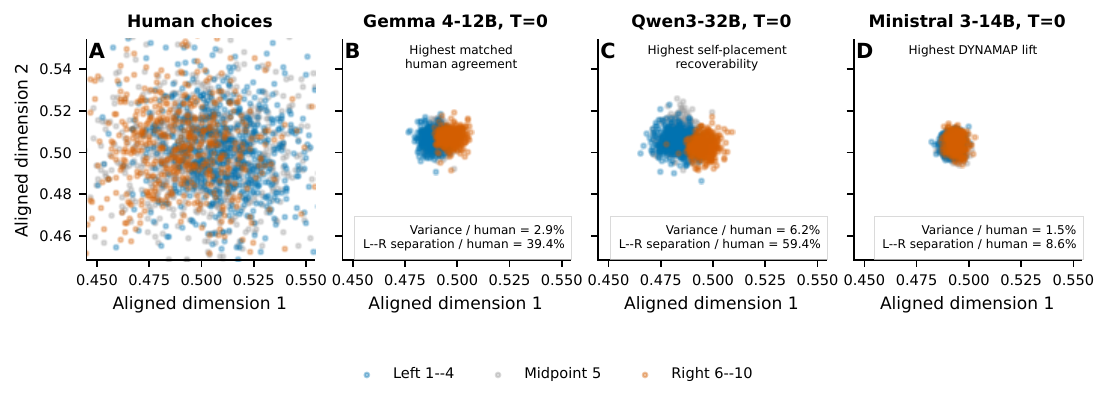}
\caption{Synthetic participant coordinates occupy a small fraction of the human latent space. Panel A shows matched human participant coordinates. Panels B--D show Gemma 4-12B at $T=0$ (highest matched human agreement), Qwen3-32B at $T=0$ (highest self-placement recoverability), and Ministral 3-14B at $T=0$ (highest DYNAMAP lift). Synthetic proposal coordinates were aligned to matched human proposal anchors with a similarity Procrustes transformation, and the fitted transformation was applied unchanged to participant coordinates. Political self-placement was not used for alignment. All panels share axes; annotations report total participant-coordinate variance and left--right centroid separation as percentages of the human reference. Estimates use the primary DYNAMAP initialization (seed 101).}
\label{fig:geometry_compression}
\figalttext[Four aligned latent-space maps with shared axes]{Human participants occupy a broad two-dimensional cloud. The three selected synthetic populations occupy much smaller regions, although Qwen3-32B shows the clearest supplied left-right separation.}
\end{figure*}

Fig.~\ref{fig:geometry_compression} illustrates the human space and the three temperature-zero checkpoints subsequently used in the factorial. The examples were chosen because they represented different grid maxima in the original condition-selection exercise: Gemma 4-12B for matched agreement, Qwen3-32B for self-placement recoverability, and Ministral 3-14B for reconstruction lift. All three synthetic participant clouds occupy a small region of the aligned adult human space. Qwen3-32B exhibits a more visible supplied left--right axis than the other examples, yet its total participant variance remains 6.2\% of human. Recovering one direction did not restore the breadth or topology of the human population. The compression result also survives the 30-seed analysis reported below.

\subsection{Position and self-placement interventions exposed distinct fragilities}

The factorial crossed the three temperature-zero checkpoints with self-placement present versus ablated and original versus inverted A/B order. Each of the 12 cells contained the same 1,991 confirmed-adult participants and 39,820 decisions. DYNAMAP accuracy remained 0.763--0.827 and lift over the validation-majority rule 0.130--0.267, so every intervention cell remained internally learnable.

\begin{figure*}[t]
\centering
\includegraphics[width=0.98\textwidth]{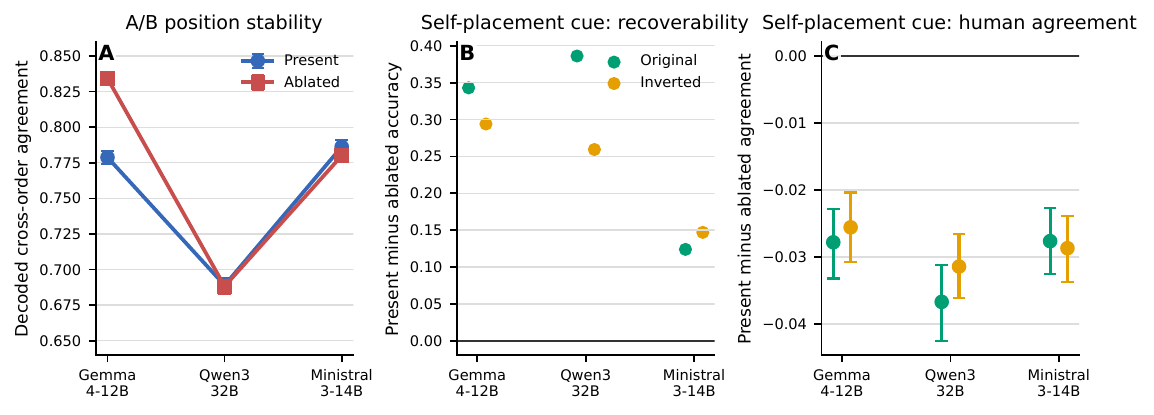}
\caption{Factorial interventions separate position stability, self-placement recoverability, and matched human agreement in the confirmed-adult cohort. Panel A shows agreement between original and inverted A/B displays after responses were decoded to original proposal identity; intervals resample paired participant histories. Panels B--C show the effect of including self-placement (present minus ablated) on held-out political self-placement classification and matched human agreement within each A/B order. Panel B reports deterministic fixed-holdout point contrasts; panel C intervals use paired participant-bootstrap resampling.}
\label{fig:confirmatory}
\figalttext[Three panels summarizing two factorial interventions]{Cross-order agreement ranges from about 0.69 to 0.83. Including self-placement increases held-out political self-placement classification accuracy for every model and order but decreases matched human agreement in every comparison.}
\end{figure*}

Internal learnability did not imply position invariance (Fig.~\ref{fig:confirmatory}A; SI Appendix, Tables~S3--S5). After decoding both displays to original proposal identity, cross-order agreement was 0.688--0.834: reversing content and position changed 16.6--31.2\% of substantive choices. Participant-bootstrap intervals were narrow; intervals that additionally varied proposal composition were wider but preserved the diagnosis (SI Appendix, Fig.~S18).

The self-placement intervention revealed a different dependence (Fig.~\ref{fig:confirmatory}B--C). Including self-placement increased held-out political self-placement classification accuracy by 0.124--0.386 in all six model-by-order contrasts, yet reduced matched human agreement in all six comparisons. With the cue ablated, held-out classification fell to 0.522--0.559 in all six cells, below the 0.573 majority rule; the recovered axis was therefore the supplied one. For Qwen3-32B in original order, held-out self-placement accuracy reached 0.925 while matched agreement was 0.523 and participant-coordinate variance was 6.2\% of human. Self-placement recoverability, matched-choice fidelity, and population heterogeneity therefore moved separately.

\subsection{Compression recurred in France and Brazil}

\begin{figure*}[t]
\centering
\includegraphics[width=0.98\textwidth]{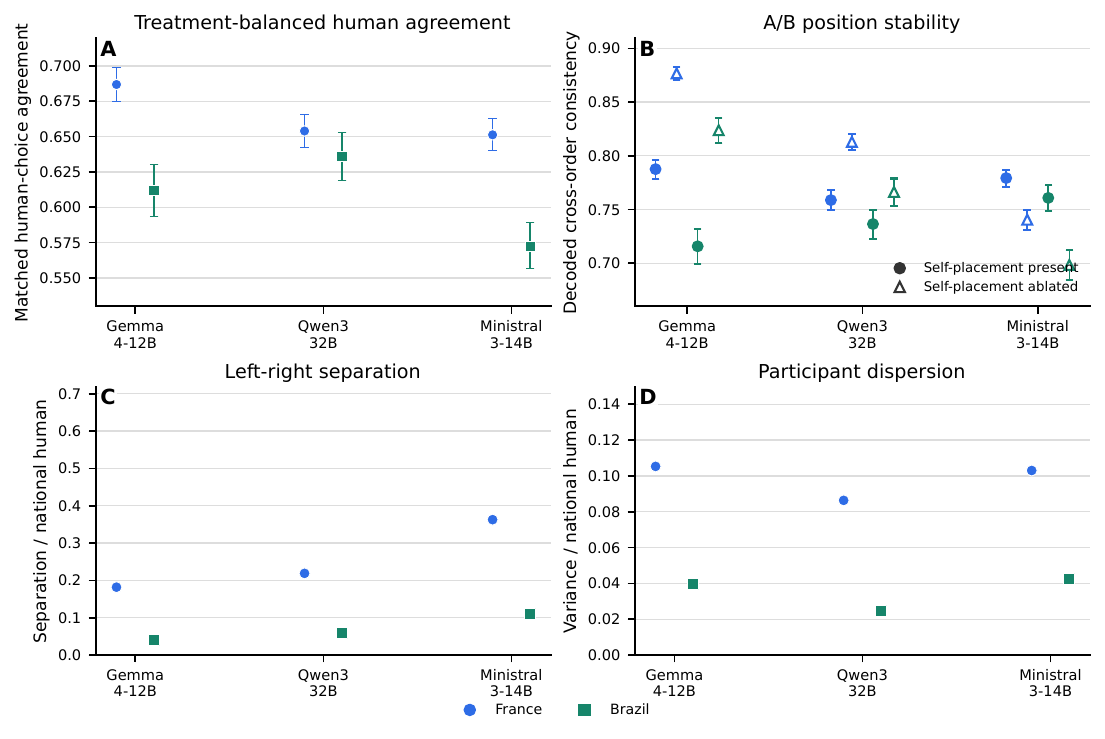}
\caption{Behavioral and geometric replication in the primary confirmed-adult French and Brazilian C20 cohorts. Panel A shows equal-weight decoded matched human agreement with participant-bootstrap 95\% confidence intervals; absolute levels are not compared across countries because instruments and released-file order differ. Panel B shows proposal-identity consistency across original and inverted displays with self-placement present (filled circles) or ablated (open triangles). At the primary DYNAMAP initialization (seed 101), panels C--D normalize left--right centroid separation and total participant-coordinate variance to each country's own adult human space after proposal-anchor alignment. The separation diagnostic is not a checkpoint ranking. French and Brazilian spaces were aligned only to their national reference and never to one another. Reconstruction and lift appear in SI Appendix, Table~S10.}
\label{fig:country_replication}
\figalttext[Four panels comparing behavior and normalized geometry in France and Brazil]{Treatment-balanced matched agreement and A/B identity consistency vary by checkpoint and setting. At seed 101, synthetic left-right separation and especially participant variance remain below the corresponding national human references.}
\end{figure*}

We replicated the factorial in the primary 20-choice cohorts in France and Brazil (Fig.~\ref{fig:country_replication}). Across the 24 country-by-checkpoint-by-prompt-by-order cells, DYNAMAP reconstruction remained within the range observed in Chile, while lift extended above it (0.200--0.331 against a Chilean maximum of 0.268; SI Appendix, Table~S10). Internal learnability therefore persisted outside Chile but again did not establish matched human fidelity.

Because released French and Brazilian rows place the selected proposal first, original-order agreement is outcome-informed. We therefore averaged equally across the four decoded prompt-by-order cells (Fig.~\ref{fig:country_replication}A). Treatment-balanced agreement was 0.651--0.687 in France and 0.572--0.636 in Brazil. These levels should not be ranked across countries because the instruments contain different proposal catalogues and the released-file order is not a common randomized display. Cross-order proposal-identity consistency ranged from 0.698 to 0.877 (Fig.~\ref{fig:country_replication}B), so inversion changed 12.3--30.2\% of substantive choices. The order-averaged displayed-A rate---the relevant positional statistic under selected-first release order---ranged from 0.36 to 0.64 (SI Appendix, Table~S10): Gemma and Qwen concentrated on visible A, whereas Ministral concentrated on B. Equal weighting across orders is therefore essential.

At the primary DYNAMAP initialization (seed 101), participant-level compression also persisted (Fig.~\ref{fig:country_replication}C--D). After equal weighting across the four factorial cells, artificial variance was 8.6--10.5\% of the French adult human reference and 2.5--4.3\% of the Brazilian reference. The directional left--right separation diagnostic retained 18.2--36.3\% of the French reference and 4.0--11.1\% of the Brazilian reference at this initialization, but it varied across DYNAMAP seeds and between the C10 and C20 cohorts and is not used to rank checkpoints. Compression therefore recurred outside Chile, but its magnitude and accompanying topology did not. The final Results section tests whether these political differences survive explicit verbal scale endpoints.

\subsection{Human benchmarks and initialization robustness bounded the claim}

\begin{figure*}[t]
\centering
\includegraphics[width=0.98\textwidth]{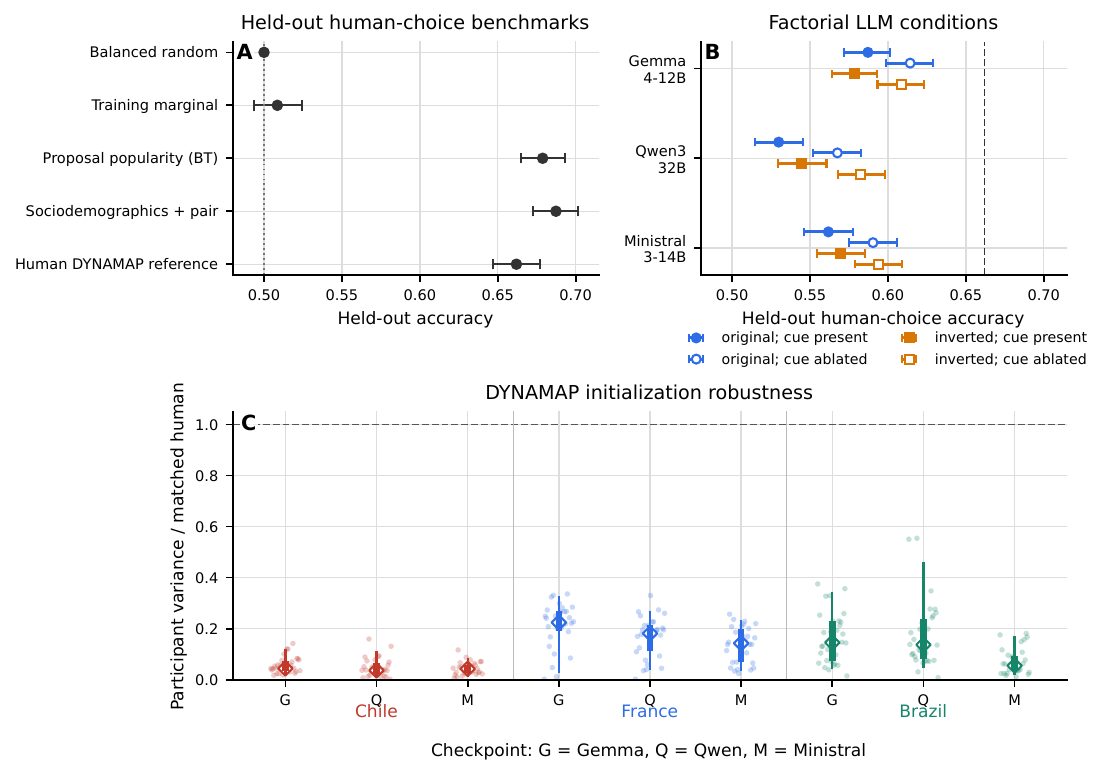}
\caption{Human benchmarks and initialization robustness bound the claim. Panel A compares five human-choice reference predictors trained on decisions 1--18 and evaluated on decisions 19--20; balanced random choice is an analytic 0.500 reference without sampling uncertainty. Panel B places all 12 factorial LLM conditions on the same held-out agreement scale; the human DYNAMAP line is a structural reference rather than a human ceiling. Panel C shows participant-coordinate variance relative to the matched same-seed human fit across 30 DYNAMAP initialization seeds for nine country-by-checkpoint groups; thin ranges span the 5th--95th percentiles, thick ranges the interquartile range, diamonds the median, and faint points individual seeds.}
\label{fig:validity}
\figalttext[Three panels of benchmarks and initialization sensitivity]{LLM held-out agreement is below three fitted human predictors. Across 30 DYNAMAP seeds, every synthetic participant-variance estimate is below the matched human reference.}
\end{figure*}

The same 3,982 held-out Chilean decisions 19--20 placed LLM agreement beside a common set of adult human reference predictors (Fig.~\ref{fig:validity}A--B). Balanced random choice was 0.500 by construction and a training-set marginal A/B rule reached 0.509; Bradley--Terry proposal popularity reached 0.679, a sociodemographic-plus-proposal predictor reached 0.687, and human DYNAMAP reached 0.662. The factorial LLM conditions reached 0.530--0.614. Human DYNAMAP is a structural reference for out-of-sample predictability, not a normative ceiling.

Across 30 independent DYNAMAP initializations, participant-coordinate variance remained below the same-seed human reference in all 270 synthetic fits (Fig.~\ref{fig:validity}C; SI Appendix, Table~S16 and Fig.~S15). This robustness concerns neural-network optimization initialization; it does not quantify LLM-generation, participant-sampling, or proposal-sampling uncertainty.

\subsection{Verbal endpoint anchors changed political direction without restoring population geometry}

\begin{figure*}[!htbp]
\centering
\includegraphics[width=0.98\textwidth]{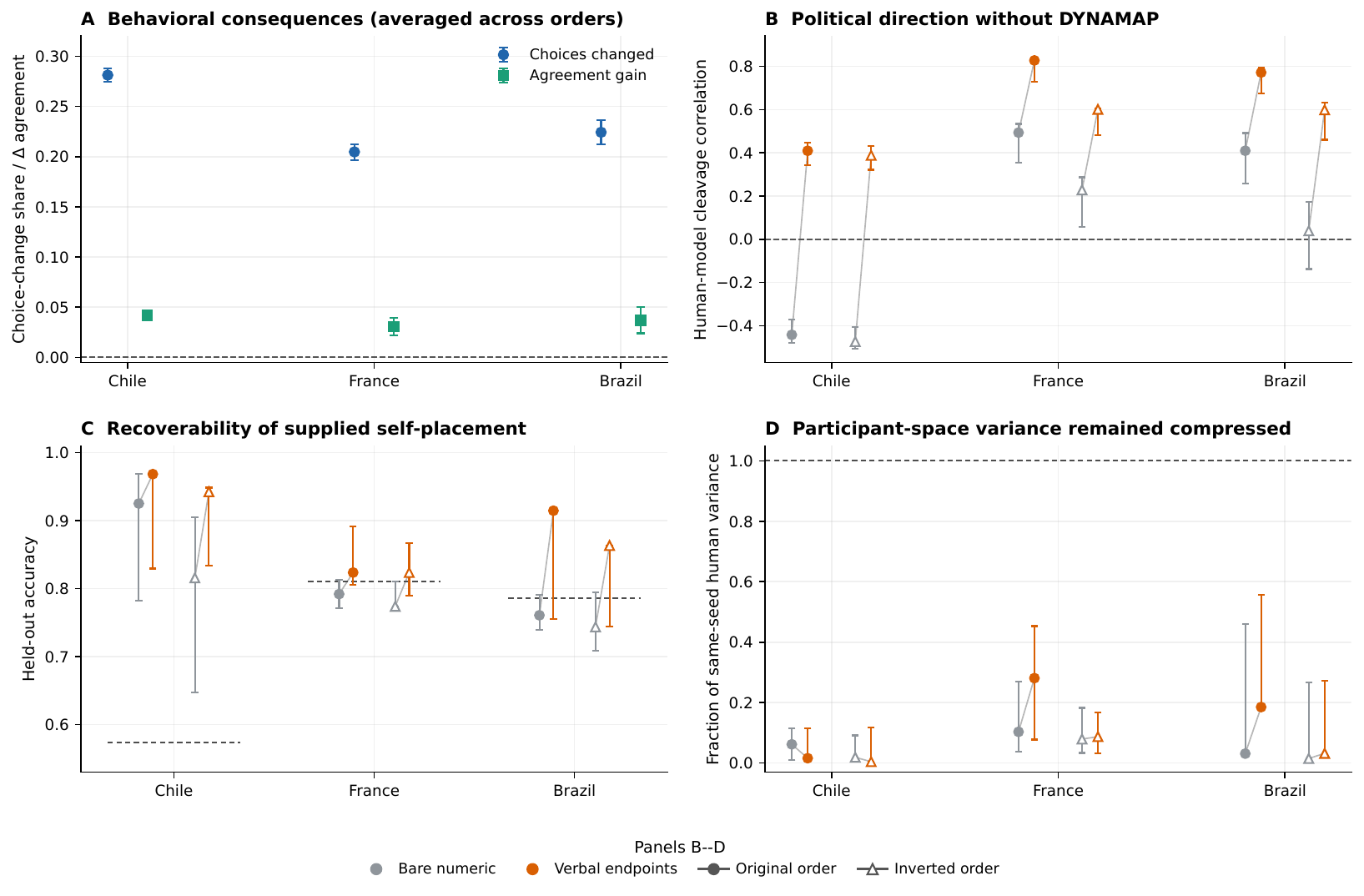}
\caption{Verbal endpoint anchors changed political organization without restoring human population geometry. The outcome-informed probe uses Qwen3-32B at $T=0$ on the confirmed-adult C20 cohorts. Panel A reports order-averaged choice changes and matched-agreement gains with paired participant-bootstrap 95\% intervals; its dashed zero line denotes no agreement gain. Panel B reports human--model Bradley--Terry cleavage correlations for bare and anchored prompts in original (circles) and inverted (triangles) displays; intervals are paired participant-cluster bootstraps. Panels C--D show seed-101 points and 5th--95th percentiles across 30 DYNAMAP initialization seeds. Short dashed segments in panel C are country-specific held-out majority rules; the dashed line in panel D is the same-seed human variance. Anchors made the supplied axis more recoverable and moved cleavage toward the human direction, but did not restore human participant-space variance or remove order sensitivity.}
\label{fig:endpoint_anchor}
\figalttext[Four panels comparing bare numeric and verbally anchored self-placement prompts]{Verbal endpoints change roughly one fifth to more than one quarter of choices, modestly improve average human agreement, move political-cleavage correlations toward the human direction, and increase recoverability of supplied self-placement. Participant-space variance remains far below the human reference and differs between original and inverted proposal order.}
\end{figure*}

The bare-numeric prompts produced different political-cleavage patterns across settings, but those patterns were not stable attributes of a checkpoint. We tested a documented prompt-specification mechanism with Qwen3-32B at $T=0$, the focal condition that made supplied self-placement most recoverable in Chile. For the same adult C20 participants, proposals, model revision, decoding parameters, and original or inverted display, the only intervention was to append verbal scale endpoints: extreme left and extreme right in each prompt's native language. Because this mechanism probe was designed after observing the bare-prompt pattern and uses one checkpoint, it diagnoses sensitivity rather than providing a second model comparison (Fig.~\ref{fig:endpoint_anchor}; SI Appendix, Table~S19 and Figs.~S19--S21).

The wording change was behaviorally consequential. Averaged equally across proposal orders, anchors changed 28.1\% [27.4\%, 28.8\%] of Chilean, 20.5\% [19.7\%, 21.2\%] of French, and 22.4\% [21.3\%, 23.6\%] of Brazilian choices. Matched human agreement increased by only 4.2, 3.1, and 3.7 percentage points, respectively, but the order-specific effects were $+5.0/+3.4$ in Chile, $+12.1/-6.0$ in France, and $+13.2/-5.8$ in Brazil (original/inverted, percentage points). Thus, the French and Brazilian average gains were concentrated in the outcome-informed released original order and reversed after inversion. Anchored cross-order change was 25.5\% in Chile, 37.5\% in France, and 43.1\% in Brazil; the latter two exceeded the 12.3--30.2\% factorial range. A more explicit political scale therefore did not make choices position-invariant.

The independent Bradley--Terry comparison nevertheless showed that the supplied axis was interpreted differently. In Chile, human--model cleavage correlation changed from $-0.442$ to 0.409 in original order and from $-0.474$ to 0.388 after inversion. France moved from 0.494 to 0.828 and from 0.228 to 0.601; Brazil moved from 0.410 to 0.772 and from 0.039 to 0.599. Verbal endpoints thus moved proposal-level political division toward the human direction in every setting and order. Magnitude did not return to equality: anchored synthetic-to-human RMS ratios were 2.66/2.67 in Chile, 1.56/1.20 in France, and 1.43/1.07 in Brazil (original/inverted). Cleavage remains informative as an embedding-independent diagnostic, but its level and direction cannot be treated as a fixed country or checkpoint signature.

DYNAMAP sharpened the distinction between political legibility and population geometry. At the primary initialization (seed 101), anchors increased held-out self-placement accuracy from 0.925 to 0.968 in original-order Chile, 0.792 to 0.824 in France, and 0.761 to 0.915 in Brazil; the inverted-order gains were also positive. In Chile, greater legibility coincided with further compression: participant-coordinate variance fell from 6.2\% and 1.9\% under bare prompts to 1.6\% and 0.4\% with anchors in original and inverted orders. Across anchored orders, variance remained 8.7--28.1\% of the French and 3.2--18.5\% of the Brazilian human reference. The original-order expansions in France and Brazil did not generalize to the inverted display, and one-dimensional separation effects varied across initialization seeds. In original order, the same clarification reversed Chilean cleavage correlation ($-0.442$ to 0.409) while narrowing participant variance (0.062 to 0.016); it strengthened French and Brazilian cleavage correlations (0.494 to 0.828 and 0.410 to 0.772) while expanding variance (0.103 to 0.281 and 0.031 to 0.185). Cross-order choice consistency improved from 0.689 to 0.745 in Chile but fell from 0.759 to 0.625 in France and from 0.737 to 0.569 in Brazil. The one-line intervention therefore had qualitatively different within-setting consequences without identifying why settings differed. Verbal endpoints could repair the direction and recoverability of a prompted political axis without restoring the breadth or positional stability of the human preference space.

\section{Discussion}

Across the same matched decisions, criteria routinely collapsed into one accuracy score came apart. LLM conditions remained below several fitted human predictors while occupying much narrower participant-level spaces. Compression was the stable result: for every focal checkpoint in every setting, and across all 30 primary robustness seeds, synthetic participants spanned less of the latent space than the matched human participants. The endpoint probe preserved this result in all 372 bare and anchored model fits spanning seed 101 and 30 robustness initializations. Political organization was more contingent. Bare numeric prompts yielded an inverted Chilean cleavage and weaker French and Brazilian correspondence, but verbal endpoints moved all three toward the human direction while leaving the participant spaces compressed and position-sensitive. A single accuracy score or model leaderboard would conceal both the robust contraction and the fragility above it.

This separation extends evidence that plausible synthetic averages can coexist with under-dispersion and altered relationships \citep{Bisbee2024}, that demographic personas can be flattened or stereotyped \citep{Wang2025}, and that persona-conditioned behavioral spaces can contract across models and tasks \citep{Ball2025,Xiao2026,Bhattacharyya2026}. Our comparisons hold participants and proposals fixed within each setting, and every one of the 270 focal same-seed refits preserved lower synthetic participant variance than the paired human fit. The strongest claim is consequently specific: these LLM choices compress the breadth of participant-level DYNAMAP geometry across the evaluated checkpoints and settings.

That claim does not extend automatically to aggregate political structure. DYNAMAP participant variance measures how broadly people are distributed after proposal-anchor alignment; Bradley--Terry cleavage measures differences in marginal proposal utilities between political groups. Synthetic proposal utilities were often more dispersed than human utilities, while cleavage could be attenuated, amplified, or directionally misaligned. The endpoint intervention establishes that cleavage is also prompt-sensitive: a one-line clarification changed its direction and magnitude without eliminating geometric compression. Political differentiation can therefore be regular and pronounced without preserving either the same individual choices or a human-like population distribution.

The mechanism probe resolves one ambiguity but not the others. Omitting verbal endpoints contributed to the bare-prompt political mismatch: adding them moved cleavage toward the human direction in all six country-by-order comparisons. This is evidence about specification, not proof that scale interpretation alone generated every difference, because the checkpoint was selected after the original pattern was known and the two prompts were executed in separate deterministic runs. Instrument composition may also matter: the Chilean catalogue represented a 2019 domestic agenda, and delete-one-topic refits show that proposal topology is sensitive to which domains are retained. National political content may be represented unevenly in model training data, a possibility consistent with prior evidence but not testable from these responses alone \citep{Santurkar2023,Atari2023,Lu2025}. A fully crossed translation of one fixed instrument into Spanish, French, and Portuguese would vary language while holding participants and policy content constant; distinguishing training-data representation would additionally require independent corpus evidence or preregistered political-knowledge probes.

The interventions identify two further boundaries. Including self-placement increased its recoverability while slightly reducing matched human agreement in the fixed factorial; adding endpoints later made that supplied cue still more legible and modestly improved order-averaged agreement. Neither result is emergent inference or superior knowledge of the person \citep{Bernardelle2025}. Reversing A/B content and position changed 16.6--31.2\% of Chilean and 12.3--30.2\% of French and Brazilian choices in the factorial, and the anchored probe remained order-sensitive. Temperature-zero decoding therefore did not make substantive preferences position-invariant.

Cross-country differences still do not identify a training-data or country-level mechanism: language, policy instruments, institutions, election contexts, and sample composition vary together. The replication establishes that compression and sensitivity recur across settings; it does not establish why their magnitudes differ. The nonuniform endpoint effects provide a concrete example of why LLM evaluations established in one language or setting need not transfer unchanged to another \citep{Atari2023}. The human samples were voluntary online participants rather than national probability samples, and the personas included age, recorded sex, and self-placement rather than the full biographical and social context that differentiates people. Sparse personas can mechanically restrict model-based heterogeneity because deterministic responses are functions of the prompt-visible profile and displayed pair. Yet participants sharing a profile usually received different proposal sequences, so within-profile coordinate variation also contains item-composition variation; a balanced repeated-item design within identical profiles would be required to identify a persona-based variance ceiling.

These results do not invalidate LLM use in social research. Prediction, data augmentation, instrument stress testing, and evaluations of model representations answer different questions and may be useful in their intended settings \citep{GudinoRosero2024}. The evidence instead limits claims of participant substitution. DYNAMAP reconstruction concerns observed participants and proposals, not cold-start prediction; participant-bootstrap intervals condition on observed policy instruments; and most geometry estimates condition on fitted networks. Model implementations also differed across families. Broader-C10, policy-domain, topic-deletion, and sequential-context analyses in the SI Appendix preserve the central diagnosis while delimiting its generality.

The appropriate unit of validation is therefore the checkpoint--prompt--setting combination: each component moved distinct fidelity criteria in these data, consistent with evidence that survey wording and response-generation method can materially alter LLM survey behavior \citep{Tjuatja2024,Ahnert2026}. We propose the following evaluation standard for simulated publics: compare internal reconstruction with a naive baseline; benchmark matched choices against out-of-sample human predictors; specify scale endpoints explicitly; ablate focal persona information; counterbalance proposal position; and separately measure participant heterogeneity, proposal topology, and political cleavage. Internally regular or politically classifiable responses should not be treated as human substitutes without independent evidence of matched-choice, positional, population-level, and relational fidelity in the specific setting of use.

\section{Materials and methods}

\subsection{Human data and fixed matched design}

The Chilean source table contained 3,007,714 pairwise decisions from 48,119 participant records over 90 policy proposals. We retained participants with one profile row, valid political self-placement and age, and at least 20 eligible decisions after excluding records flagged as nonpolitical. This yielded 31,763 eligible participants and 2,311,464 decisions. With seed 20260821, we drew a fixed frame of 2,000 anonymous participant identifiers (UUIDs) without replacement and retained each participant's first 20 eligible decisions chronologically. The confirmed-adult cohort then restricted that frame to the 1,991 participants with an exact reported age of at least 18, yielding 39,820 decisions. It included 674 participants recorded as female, 1,310 as male, and seven in another category; mean age was 36.4 years (SD, 10.8). The Chilean instrument defined positions 1--4 as left, 5 as center, and 6--10 as right. Binary left--right analyses used 653 left- and 503 right-identifying adults; the 835 midpoint participants were excluded only from those binary contrasts and retained in every other analysis. UUIDs linked records but never entered prompts. Except for the cohort-construction counts in this paragraph, every reported sample size and estimate in the main text and SI Appendix refers to the confirmed-adult cohorts.

\subsection{Model grid, personas, and inference}

The grid crossed three instruction-tuned model families and three checkpoint sizes: Qwen3 (8B, 14B, and 32B), Gemma 4 (E4B, 12B, and 31B), and Ministral 3 (3B, 8B, and 14B). Exact model identifiers, repository revisions, inference engines, and numerical formats are reported in SI Appendix, Table~S1. Open-weight checkpoints were chosen because immutable revisions and local inference settings permit exact replay; this design does not represent hosted proprietary systems. Because families differed in training, architecture, serving software, and precision, cross-family comparisons are descriptive rather than causal effects of parameter count.

The Spanish system message situated the decision in Chile in 2019, prohibited invented attributes and later information, required one choice, and prohibited explanation. The user message supplied age, recorded sex, political self-placement, and the two proposal texts. The response contract admitted only A or B. UUIDs, human choices, coordinates, timestamps, proposal popularity, and prior decisions were excluded. Reasoning was disabled. Each checkpoint was evaluated at $T\in\{0,0.2,0.7\}$ with top-$p=1$ and fixed generation seeds documented in the SI Appendix. Because each condition contains one draw per decision cell, temperature is evaluated as a decoding intervention and not as a proxy for human heterogeneity. Verbatim templates in Spanish, French, and Portuguese appear in the SI Appendix.

\subsection{Factorial interventions and response decoding}

From the completed grid, we selected three temperature-zero checkpoints representing different diagnostic maxima: Gemma 4-12B for matched human agreement, Qwen3-32B for self-placement recoverability, and Ministral 3-14B for DYNAMAP lift. This checkpoint selection was outcome-informed. Before generating the intervention responses, we fixed a $2\times2$ design crossing political self-placement present versus ablated with original versus content-and-position-inverted A/B display. The balanced design comprised 12 cells; nine were newly generated and three exact original-order, self-placement-present cells were reused after verifying identity and completeness. The stored responses were projected onto the same confirmed-adult UUID manifest, so every analyzed cell contained 1,991 participants and 39,820 decisions.

In inverted cells, displayed A contained the original B proposal and displayed B contained the original A proposal. All responses were decoded to original proposal identity before substantive comparisons. The self-placement effect is present minus ablated within checkpoint and order; the position effect is inverted minus original within checkpoint and prompt condition. Behavioral intervals used 2,000 paired bootstrap replicates that resampled adults with their complete 20-decision histories. Classification contrasts were recalculated on the adult coordinate fits with the fixed split and folds described below.

All 12 cells used the same two-dimensional DYNAMAP architecture, primary initialization seed 101, 20 epochs, and decisions 1--18/19--20 training--validation split as the grid. Political-self-placement validation used the same random forest with 2,000 estimators and seed 42. Primary-fit geometry intervals are post-fit: they resample participants or proposal pairs while holding one fitted coordinate system fixed. Optimization sensitivity was assessed separately with repeated DYNAMAP refits.

\subsection{France--Brazil cross-country replication}

The cross-country replication used policy choices collected through an online participation platform during the 2022 French and Brazilian presidential elections \citep{Navarrete2023Data,Navarrete2024}. Cleaning required nonmissing participant, pair, selected-proposal, and timestamp fields; two distinct catalogue-backed proposals; a selected item present in the pair; \texttt{suspicious=0}; the latest valid observation per participant and unordered proposal pair; and the latest usable profile with documented age, sex, and political self-placement. Age was released in bands. To ensure adult-only scope, we excluded the entire 10--19 band because ages 18--19 could not be separated from younger respondents. The primary C20 cohorts therefore comprised 843 confirmed adults and 16,860 rows in France, and 434 confirmed adults and 8,680 rows in Brazil. Decisions 1--18 trained DYNAMAP and 19--20 were held out. A broader confirmed-adult C10 sensitivity included 847 French and 515 Brazilian participants and used decisions 1--8/9--10. Because C10 changes both cohort membership and training volume, it is not a causal history-length intervention.

The same three checkpoints and temperature-zero stateless factorial were rendered in French and Brazilian Portuguese using native proposal text and the 2022 context. French and Brazilian five-point self-placement was grouped as 1--2 left/center-left, 3 center, and 4--5 center-right/right; scales were interpreted only within country. Stored responses were projected onto the confirmed-adult C20 and C10 UUID manifests before analysis; no model inference was repeated for the age restriction. Released human records place the selected proposal first, yielding an outcome-informed released-file order rather than verified historical display position. We therefore decoded every response to proposal identity and used the equal-weight average over all four self-placement-by-order cells as the primary agreement estimate. Cross-order consistency compared proposal identity in paired original and inverted rows with 2,000 participant-bootstrap resamples.

\subsection{Verbal endpoint-anchor mechanism probe}

After inspecting the bare-numeric results, we ran an outcome-informed diagnostic with Qwen3-32B at $T=0$ in Chile, France, and Brazil. It reused the confirmed-adult C20 cohorts and the exact original and inverted decision keys. Model revision, runtime, numerical format, system message, persona fields, proposal text, response contract, and decoding parameters were held fixed. The only prompt change appended verbal endpoints to the self-placement line: ``1 = extrema izquierda; 10 = extrema derecha'' in Chile, ``1 = extr\^{e}me gauche; 5 = extr\^{e}me droite'' in France, and ``1 = extrema esquerda; 5 = extrema direita'' in Brazil. Six new cells were paired with their existing bare-numeric controls and decoded to original proposal identity.

Behavioral and Bradley--Terry contrasts used 2,000 paired UUID-cluster bootstrap replicates. DYNAMAP was refitted independently for bare and anchored responses using the same C20 split and estimator specification. Seed 101 provides the manuscript-compatible point estimates; 30 additional paired initializations refitted every model condition against the same-seed national human reference. These seed distributions quantify optimization sensitivity conditional on fixed model responses. Because the checkpoint and intervention were selected after the earlier pattern was observed, this probe tests a candidate mechanism and does not estimate a general effect of verbal anchoring across checkpoints.

\subsection{Behavioral, classification, and uncertainty estimands}

Matched human agreement equals one when a model selected the substantive proposal chosen by the corresponding human participant. In inverted cells it was computed only after decoding the displayed response to original proposal identity. Original-A rate measures selection of that original proposal identity, whereas displayed-A rate diagnoses positional response concentration. Participant-level response entropy and persistence were calculated on explicitly identified decoded or displayed outcomes. Confidence intervals used 2,000 bootstrap replicates resampling participants with complete histories; no analysis treated decision rows as independent.

For a common Chilean behavioral benchmark, decisions 1--18 trained reference predictors and decisions 19--20 (3,982 adult choices) evaluated balanced random choice, the training-set marginal A/B rule, Bradley--Terry proposal popularity \citep{Bradley1952}, a sociodemographic-plus-proposal-identity classifier, and human DYNAMAP on the same held-out cells. Balanced random accuracy is the analytic value 0.500 and therefore has no sampling interval. Comparable LLM outputs were evaluated on those cells after response decoding where required.

Self-placement recoverability was evaluated after excluding only each country's midpoint category. Random forests with 2,000 trees, entropy splits, and seed 42 predicted binary left versus right self-placement from archived sociodemographic fields, then from those fields plus the two DYNAMAP coordinates. We report a fixed, unstratified 70/30 holdout together with stratified 10-fold cross-validation. The fixed holdout is preserved to avoid post-result redefinition; balanced accuracy, macro-F1, ROC AUC, and stratified CV-10 provide class-imbalance and partition sensitivities. Coordinate gain is the difference between augmented and sociodemographic-only held-out accuracy; it is distinct from absolute classification performance. Because self-placement was supplied in one factorial arm, this analysis measures recoverability of a supplied or ablated cue rather than emergent political inference.

Participant-bootstrap intervals condition on the observed proposal set. As a sensitivity analysis, a crossed Poisson-pigeonhole bootstrap applied product weights to participants, proposal A, and proposal B. This approximates uncertainty under proposal interchangeability but does not turn the observed policies into a probability sample of a defined policy universe.

\subsection{DYNAMAP reconstruction}

Human and synthetic spaces were fitted independently with the same two-dimensional DYNAMAP architecture. DYNAMAP jointly learns participant and proposal coordinates from repeated pairwise choices, with all coordinates constrained to $(0,1)$ by a componentwise sigmoid. Choice prediction uses the elementwise squared participant--proposal displacements as inputs to a flexible shared-parameter neural head, rather than imposing a monotonic Euclidean-distance choice rule. The complete choice function and its parameter sharing are specified in the SI Appendix (\textit{DYNAMAP architecture and validation}). Models minimized binary cross-entropy with Adamax (learning rate 0.01), batch size 64, no epoch shuffling, primary initialization seed 101, and 20 epochs. In C20, decisions 1--18 trained the model and decisions 19--20 formed within-participant validation sets of 3,982 Chilean, 1,686 French, and 868 Brazilian rows. C10 used decisions 1--8/9--10. We report accuracy, the validation-set majority-label accuracy, and their difference (lift). These are deterministic validation-set point diagnostics conditional on a fitted model; initialization sensitivity is reported separately rather than as participant-sampling confidence intervals. They measure reconstruction for observed participants and items, not matched human fidelity or prediction for new participants.

\subsection{Alignment, geometry, and independent validation}

Because independently learned coordinates are arbitrary up to orientation and scale, each synthetic proposal matrix---90 items in Chile, 120 in France, and 67 in Brazil---was mapped to its own country's human proposal matrix with a least-squares similarity Procrustes transformation \citep{Schonemann1966}. The fitted rotation or reflection, scale, and translation were then applied unchanged to participant coordinates; political self-placement did not enter alignment, and spaces were never aligned across countries. We measured left--right participant-centroid distance, total participant covariance trace, and the Pearson correlation between all matched pairwise proposal distances. The first two were normalized to the corresponding human value.

Optimization sensitivity varied only the DYNAMAP initialization across 30 fixed seeds for the confirmed-adult C20 human samples and the three focal temperature-zero, self-placement-present, original-order model conditions in each country. Human and synthetic spaces were refitted at the same seed before alignment, yielding 270 focal synthetic--human comparisons. These distributions quantify sensitivity to DYNAMAP optimization, not uncertainty in model generation or participant sampling. Broader-C10, policy-domain, topic-deletion, and sequential-context analyses are reported in the SI Appendix; the sequential analysis is compared only with its matched local stateless implementation.

To ensure that proposal-level political conclusions did not depend on the DYNAMAP embedding, we fitted L2-regularized Bradley--Terry logistic models without an intercept to aggregate proposal choices. We report proposal-score and rank correlations, pairwise ordering agreement, and synthetic-to-human root-mean-square score dispersion. Political cleavage was estimated by fitting the same model separately to left- and right-identifying participants and defining each proposal's cleavage as centered right-group minus left-group utility. Human and synthetic cleavage vectors were compared by Pearson and Spearman correlation, sign agreement, slope through the origin, and relative root-mean-square magnitude. Two thousand paired participant-cluster bootstrap replicates refitted the human and synthetic models with identical multiplicities. DYNAMAP participant variance, aggregate proposal dispersion, and political cleavage are distinct estimands and need not move together.



\section*{Use of generative AI}

OpenAI Codex and ChatGPT (GPT-5.6 Sol; OpenAI) assisted with code review, consistency checks, and editorial revision during August–September 2026. The authors specified the analyses, reviewed the generated code and text, verified the numerical outputs, and take responsibility for all results and content.

\section*{Ethics and informed consent}

This study is based exclusively on secondary analysis of previously collected anonymous data and involved no new participant recruitment, interaction, or data collection. Participation in the original online platforms was voluntary and uncompensated. The datasets contain session-generated identifiers used to link responses within the platforms, but no names, national identification numbers, email addresses, or other direct personal identifiers were collected.

The French and Brazilian data were collected through the MonProgramme2022 and Brazucracia platforms under ethics-approved protocols, with participants accepting the platforms' consent form, as documented in the original study \citep{Navarrete2024}. The Chilean data were collected through the Chilecracia online civic-participation platform. The secondary analysis reported here was reviewed by the Institutional Ethics Committee of Universidad del Desarrollo, which determined that it was exempt from full ethical review on June 24, 2025.

\section*{Acknowledgments}

The authors have no additional acknowledgments to report.

\section*{Funding}

The authors acknowledge financial support from the Instituto de Data Science, Universidad del Desarrollo. C.C. acknowledges financial support from Agencia Nacional de Investigación y Desarrollo (ANID), FONDECYT Regular grant no. 1262601.

\section*{Author contributions}

A.F.-J.: Conceptualization, Methodology, Investigation, Software, Formal analysis, Data curation, Validation, Visualization, Project administration, Writing---original draft, Writing---review and editing. C.C.: Conceptualization, Methodology, Supervision, Resources, Writing---review and editing.

\section*{Conflicts of interest}

The authors declare a competing interest. A.F.-J. and C.C. are named inventors on a pending U.S. patent application related to the DYNAMAP framework described in this article and assigned to Universidad del Desarrollo.

\section*{Data availability}

The France and Brazil datasets analyzed in this study are publicly available through \href{https://doi.org/10.7910/DVN/8E0EA4}{Harvard Dataverse (DOI: 10.7910/DVN/8E0EA4)} \citep{Navarrete2023Data}. The Chilean dataset was originally collected through a publicly accessible online platform that is no longer active; an archived description of the platform and its data is available through the \href{https://web.archive.org/web/20191229112008/https://chilecracia.org/datos}{Internet Archive}. The \href{https://github.com/adolfofuentesjofre/DYNAMAP}{public DYNAMAP repository} provides a reduced Chilean replication dataset and selected fitted outputs. Cleaned analytical data, prompt materials, validated model responses, environment specifications, and derived outputs for the present study are available from the corresponding author during review, subject to ethical and data-protection considerations, and will be deposited in a public repository upon acceptance. Direct identifiers and private participant-level joins will not be released.

\section*{Code availability}

The DYNAMAP implementation and reduced replication materials are available at \href{https://github.com/adolfofuentesjofre/LLMs_Compress_human_preference_geometry}{github.com/adolfofuentesjofre/LLMs}. Analysis code specific to the present study is available from the corresponding author during review and will accompany the public repository deposit upon acceptance.

\bibliographystyle{oup-plain}
\bibliography{references}

\end{document}


\maketitle
\renewcommand{\thetable}{S\arabic{table}}
\renewcommand{\thefigure}{S\arabic{figure}}

\section{Supplementary methods}

This Supporting Information documents the complete model grid and factorial cells, the country-specific preprocessing and estimands, and the sensitivity analyses that delimit the main-text claims. The prompt-sensitive political organization reported in the main text is supported by the endpoint-anchor diagnostic, classification tables, aggregate proposal-space validation, and political-cleavage analysis below. DYNAMAP (Dynamic System for Mapping of Users and Their Preferences Using Artificial Intelligence) is the neural comparative-choice model used throughout. The SI preserves seven criteria: internal DYNAMAP reconstruction, matched human agreement, self-placement recoverability, position stability, participant geometry, proposal topology, and political cleavage.

\subsection{Chilean fixed sample and unit of analysis}

Data preparation preserved a full integrated choice table, an eligibility-filtered table, a fixed replay frame, and a confirmed-adult analysis manifest. The source contained 3,007,714 pairwise choices from 48,119 anonymous participant identifiers (UUIDs) over 90 policy proposals. Eligibility required one profile row, recorded political self-placement on the original 1--10 scale, age in the archived 0--110 range, at least 20 eligible political decisions, and no record-level exclusion flag. The eligible pool contained 31,763 UUIDs and 2,311,464 choices. Sampling without replacement used NumPy seed 20260821 and drew a fixed frame of 2,000 UUIDs. The confirmed-adult cohort retained the 1,991 frame members with an exact reported age of at least 18 and their first 20 eligible choices in chronological order, yielding exactly 39,820 matched cells. The survey coding defined 1--4 as left, 5 as center, and 6--10 as right. The 835 midpoint participants were excluded only from binary left--right analyses; all 1,991 confirmed adults remained in behavioral, reconstruction, and geometry analyses. The 2,000-person frame is reported only to document cohort construction; unless explicitly labeled as a bootstrap or estimator count, every reported sample size and estimate below refers to a confirmed-adult cohort.

The decision cell is the descriptive comparison unit, but behavioral uncertainty resamples UUIDs and retains all 20 repeated decisions. The coordinate-validation label is defined at the UUID level. No UUID, observed human response, timestamp, learned coordinate, or proposal popularity statistic entered a model prompt.

\subsection{Chilean persona and response contract}

The prompt was in Spanish and fixed across the 27-condition model grid. The system message specified that the respondent was choosing between two public-policy proposals in Chile in 2019, instructed the model to use only the supplied profile, prohibited invented attributes and information after 2019, required a single choice, and prohibited explanation. The user message supplied recorded sex, age, political self-placement, and the exact original texts of options A and B. The response contract admitted only A or B in the archived structured format. Reasoning was disabled.

This profile is sparse. It omits education, income, occupation, religion, locality, family context, life history, and prior behavior. Among the 1,991 Chilean confirmed adults there were 607 distinct recorded sex--age--self-placement profiles, and 87.24\% shared that profile with at least one other participant. Sparse personas can mechanically limit model-based heterogeneity because a deterministic response is a function of the visible profile and displayed pair. At the same time, participants sharing a profile generally received different 20-pair sequences, so within-profile coordinate variation also contains item-composition variation. The present design therefore cannot identify a strict persona-based ceiling or cleanly decompose preference heterogeneity from sequence composition; doing so would require balanced repeated items within identical profiles. The design evaluates a controlled prompt-to-response mapping rather than psychological identification with a person. Downstream self-placement classification measures consistent use of a supplied attribute, not emergent inference about an unobserved orientation.

\subsection{Verbatim prompt templates}

The following templates reproduce the archived messages. Bracketed terms replace row-specific values or proposal text; they were not shown literally. The self-placement-ablated arm deleted only the corresponding self-placement line. The inverted arm exchanged both proposal texts and their A/B positions. The response contract was unchanged.

\paragraph{Chile (Spanish).}
\begin{prompttemplate}
SYSTEM
Estás simulando una decisión individual en un estudio de preferencias de política pública realizado en Chile en 2019.
Adopta únicamente el perfil proporcionado y sitúa tu decisión en ese momento temporal.
No inventes otros rasgos, experiencias o preferencias.
No bases tu decisión en acontecimientos o información posteriores a 2019.
Debes elegir exactamente una de las dos alternativas. No hay empate.
No expliques ni justifiques tu decisión.

USER
Perfil de la persona:
- Edad: [edad]
- Sexo: [sexo]
- Autoubicación política: [valor]/10

Pregunta original:
¿Qué priorizarías?
A: [texto de la propuesta A]
B: [texto de la propuesta B]

Devuelve exactamente uno de estos objetos JSON:
\{"answer":"A"\}
\{"answer":"B"\}
\end{prompttemplate}

\paragraph{France (French).}
\begin{prompttemplate}
SYSTEM
Vous simulez une décision individuelle dans une étude sur les préférences en matière de politiques publiques menée en France lors de l’élection présidentielle de 2022.
Adoptez uniquement le profil fourni et situez votre décision dans ce contexte temporel.
N’inventez pas d’autres caractéristiques, expériences ou préférences.
Ne fondez pas votre décision sur des événements ou des informations postérieurs à 2022.
Vous devez choisir exactement l’une des deux options. Il n’y a pas d’égalité.
N’expliquez ni ne justifiez votre décision.

USER
Profil de la personne :
- Tranche d’âge : [tranche d’âge]
- Sexe : [sexe]
- Auto-positionnement politique : [valeur]/5

Question originale :
Quelle proposition préférez-vous ?
A : [texte de la proposition A]
B : [texte de la proposition B]

Renvoyez exactement l’un de ces objets JSON :
\{"answer":"A"\}
\{"answer":"B"\}
\end{prompttemplate}

\paragraph{Brazil (Brazilian Portuguese).}
\begin{prompttemplate}
SYSTEM
Você está simulando uma decisão individual em um estudo de preferências de políticas públicas realizado no Brasil durante a eleição presidencial de 2022.
Adote apenas o perfil fornecido e situe sua decisão nesse contexto temporal.
Não invente outras características, experiências ou preferências.
Não baseie sua decisão em acontecimentos ou informações posteriores a 2022.
Você deve escolher exatamente uma das duas opções. Não há empate.
Não explique nem justifique sua decisão.

USER
Perfil da pessoa:
- Faixa etária: [faixa etária]
- Sexo: [sexo]
- Autoposicionamento político: [valor]/5

Pergunta original:
Qual proposta você prefere?
A: [texto da proposta A]
B: [texto da proposta B]

Retorne exatamente um destes objetos JSON:
\{"answer":"A"\}
\{"answer":"B"\}
\end{prompttemplate}

\paragraph{Verbal endpoint-anchor variant.}
The later mechanism probe changed only the self-placement line in these templates. The replacement text was \emph{Autoubicaci\'on pol\'itica: [valor]/10 (1 = extrema izquierda; 10 = extrema derecha)} in Chile, \emph{Auto-positionnement politique : [valeur]/5 (1 = extr\^{e}me gauche ; 5 = extr\^{e}me droite)} in France, and \emph{Autoposicionamento pol\'itico: [valor]/5 (1 = extrema esquerda; 5 = extrema direita)} in Brazil. No intermediate categories, substantive interpretation, or additional persona information was added.

\subsection{Model grid and serving provenance}

The grid crosses nine open-weight checkpoints with $T\in\{0,0.2,0.7\}$; Table~\ref{tab:s_models} gives their exact identifiers and serving specifications. Open-weight checkpoints permit immutable revisions and local inference settings to be fixed for exact replay; hosted proprietary systems are outside this design. Top-$p$ was 1.0. Stochastic conditions use the archived run-level sampling seed; temperature-zero runs have no effective sampling seed. Within every checkpoint, repository revision and inference-engine version are identical across temperatures. Qwen3 and Gemma 4 use bfloat16 weights and computation; Ministral 3 uses native FP8 weights with bfloat16 computation. Family and size comparisons therefore describe the deployed checkpoints and are not controlled causal effects of architecture or parameter count. One response was generated per participant--decision cell, so temperature probes decoding sensitivity rather than a respondent-level distribution or human heterogeneity.

\begin{table*}[t]
\caption{Exact checkpoint and serving specifications. Repository revisions are immutable commit identifiers. All three temperatures were completed for each row.}
\label{tab:s_models}
\centering
\scriptsize
\resizebox{\textwidth}{!}{%
\begin{tabular}{lllllll}
\toprule
Family & Size & Model identifier & Revision & Engine & Precision & Quantization \\
\midrule
Qwen3 & 8B & \texttt{Qwen/Qwen3-8B} & \texttt{b968826d9c46dd6066d109eabc6255188de91218} & vLLM 0.21.0 & bfloat16 & none \\
Qwen3 & 14B & \texttt{Qwen/Qwen3-14B} & \texttt{40c069824f4251a91eefaf281ebe4c544efd3e18} & vLLM 0.21.0 & bfloat16 & none \\
Qwen3 & 32B & \texttt{Qwen/Qwen3-32B} & \texttt{9216db5781bf21249d130ec9da846c4624c16137} & vLLM 0.21.0 & bfloat16 & none \\
Gemma 4 & E4B & \texttt{google/gemma-4-E4B-it} & \texttt{ee0ef6023621cff504d758262d4e04895a5af4a2} & vLLM 0.28.0 & bfloat16 & none \\
Gemma 4 & 12B & \texttt{google/gemma-4-12B-it} & \texttt{707f0a3b8a3c7ad586ed01e27eafbad8a27dd0f7} & vLLM 0.28.0 & bfloat16 & none \\
Gemma 4 & 31B & \texttt{google/gemma-4-31B-it} & \texttt{842da3794eaa0b77d5f08bae87a17459d91ff475} & vLLM 0.28.0 & bfloat16 & none \\
Ministral 3 & 3B & \texttt{mistralai/Ministral-3-3B-Instruct-2512} & \texttt{b35d4dfe56c142746f54dbd64f579faab2744308} & vLLM 0.28.0 & bfloat16 & native FP8 weights \\
Ministral 3 & 8B & \texttt{mistralai/Ministral-3-8B-Instruct-2512} & \texttt{5b26027e7b19eeb4b7352e1fed3926375dd2cb4d} & vLLM 0.28.0 & bfloat16 & native FP8 weights \\
Ministral 3 & 14B & \texttt{mistralai/Ministral-3-14B-Instruct-2512} & \texttt{29439f81c2be264d8d393273f99e7db9c0961120} & vLLM 0.28.0 & bfloat16 & native FP8 weights \\
\bottomrule
\end{tabular}
}
\end{table*}

\subsection{DYNAMAP architecture and validation} \label{sec:dynamap_architecture}

For each response source, DYNAMAP jointly learns two-dimensional participant and proposal coordinates from repeated pairwise choices. Let $\mathbf{u}_i \in (0,1)^d$ denote the coordinate of participant $i$ and $\mathbf{v}_a,\mathbf{v}_b \in (0,1)^d$ those of the two proposals, with $d=2$ in all analyses. Trainable participant and proposal embedding tables are transformed componentwise by a sigmoid, constraining all coordinates to the unit square.

The baseline choice function uses a flexible shared-parameter head rather than a monotonic Euclidean-distance rule. For each participant -proposal pair, it first computes elementwise squared displacements
\[
\boldsymbol{\delta}_{ia}
=
(\mathbf{v}_a-\mathbf{u}_i)\odot
(\mathbf{v}_a-\mathbf{u}_i),
\qquad
\boldsymbol{\delta}_{ib}
=
(\mathbf{v}_b-\mathbf{u}_i)\odot
(\mathbf{v}_b-\mathbf{u}_i).
\]

These vectors are not summed into scalar Euclidean distances before entering the choice head. Instead, the same learned transformation is applied to both alternatives:
\[
\mathbf{h}_{ia}
=
\exp(\boldsymbol{\delta}_{ia}\mathbf{W}_1+\mathbf{b}_1),
\qquad
\mathbf{h}_{ib}
=
\exp(\boldsymbol{\delta}_{ib}\mathbf{W}_1+\mathbf{b}_1),
\]
where $\mathbf{W}_1\in\mathbb{R}^{d\times4}$, yielding four hidden components per alternative and eight after concatenation. A second shared transformation produces one scalar per alternative,
\[
r_{ia}=\exp(\mathbf{h}_{ia}\mathbf{w}_2+b_2),
\qquad
r_{ib}=\exp(\mathbf{h}_{ib}\mathbf{w}_2+b_2),
\]
and an unconstrained one-unit dense output layer gives
\[
\widehat P(A\mid i,a,b)
=
\sigma(\alpha_A r_{ia}+\alpha_B r_{ib}+c).
\]
The training target is one when the displayed proposal A was selected and zero when B was selected. Thus, although the intermediate alternative-specific transformations share parameters, the baseline architecture does not structurally require the closer proposal in Euclidean distance to be preferred, nor does it impose exact A/B-swap antisymmetry on the predicted choice probability.

All conditions were optimized with binary cross-entropy and Adamax (learning rate 0.01), batch size 64, 20 epochs, and initialization seed 101. The same architecture and estimation settings were used for human and synthetic response sources within each comparison.

Decisions 1--18 per UUID form the training data and decisions 19--20 form validation (35,838 and 3,982 rows). Both partitions contain the same people and proposal universe. DYNAMAP accuracy is therefore internal reconstruction for observed users and items. The majority baseline is computed separately within each condition's validation labels, and lift is DYNAMAP minus that baseline. Accuracy and lift are deterministic validation-set point diagnostics conditional on a fitted network; the 30-seed analysis separately quantifies optimization sensitivity, and no participant-sampling confidence interval is claimed for these diagnostics.

\subsection{Political self-placement validation}

The Chilean midpoint 5 was omitted from binary validation, leaving 653 UUIDs at 1--4 and 503 at 6--10. The majority class is therefore 0.565 of the binary adult cohort. Random forests used 2,000 trees, entropy splits, and seed 42. The sociodemographic specification used age, recorded sex, and archived region code; the augmented specification added the two DYNAMAP coordinates. We report a fixed, unstratified 70/30 holdout and stratified 10-fold cross-validation. The fixed holdout contains 199 majority-class participants among 347 (0.573). It is retained to avoid redefining the primary estimand after results were observed; CV-10 supplies a stratified partition sensitivity. The human-coordinate reference is 0.674 held-out and 0.676 CV-10, compared with 0.516 for the sociodemographic-only held-out model. The estimand is recoverability of political self-placement from repeated choices, not matched human agreement.

\subsection{Proposal-anchor alignment and geometric diagnostics}

Let $P_H$ be the $90\times2$ human proposal matrix and $P_M$ a model proposal matrix. A similarity Procrustes transformation estimates rotation/reflection $R$, scalar $c$, and translation $t$ by minimizing
\[
\left\lVert P_H-(cP_MR+\mathbf{1}t^\top)\right\rVert_F^2.
\]
The fitted transformation is applied unchanged to model participant coordinates; self-placement labels do not enter alignment \citep{Schonemann1966}. Total coordinate variance is the trace of the covariance matrix across all users. Left--right separation is the Euclidean distance between left and right centroids. Proposal topology is Pearson correlation between all matched pairwise proposal distances. Variance and separation are divided by the corresponding human estimate.

\subsection{Self-placement-by-position factorial intervention}

Three temperature-zero checkpoints were selected from the completed grid to represent its best matched human agreement (Gemma 4-12B), held-out self-placement recoverability (Qwen3-32B), and DYNAMAP lift (Ministral 3-14B). Checkpoint selection was therefore outcome-informed. Before generating the nine new response cells, we fixed a balanced $2\times2$ intervention crossing political self-placement present versus ablated from the persona with original versus inverted proposal display. The analyzed 12 cells contain the same 1,991 confirmed-adult UUIDs and 39,820 decisions each. The stored responses were projected to the adult manifest; no inference was repeated. The three self-placement-present, original-order reference cells and their adult DYNAMAP fits were reused after exact verification.

The position intervention exchanged both proposal content and displayed label: displayed A contained original proposal B and displayed B contained original proposal A. Inverted responses were decoded to original proposal identity before matched human agreement, original-A rate, DYNAMAP, self-placement validation, geometry, or rankings were computed. Displayed-A rate was retained separately as a label-position diagnostic. Cross-order stability is paired agreement between original and inverted responses after this decoding. The self-placement contrast is present minus ablated within checkpoint and display order; the position contrast is inverted minus original within checkpoint and prompt condition.

All DYNAMAP and political-validation settings match the primary grid. Behavioral intervals use 2,000 bootstrap replicates resampling UUIDs with their complete 20-decision histories. Held-out classification contrasts pair the same 347 test UUIDs; CV-10 differences pair folds and are sensitivity summaries rather than independent-population intervals. Holm correction was applied separately to the six focal held-out self-placement contrasts and to the complete family of 42 behavioral self-placement contrasts. The latter comprises matched human agreement, original-A rate, displayed-A rate, decoded entropy, displayed entropy, decoded persistence, and displayed persistence, crossed with three checkpoints and two orders. It is broader because these seven behavioral diagnostics were archived as one family, whereas held-out classification contained only the six focal prompt contrasts. With 2,000 bootstrap replicates, 0.001 is the minimum attainable raw two-sided $P$; its Holm-adjusted bounds are 0.006 in a six-test family and 0.042 in a 42-test family. User-coordinate intervals resample UUIDs while holding coordinates fixed after one DYNAMAP fit, and proposal-distance intervals resample matched proposal pairs. Neither includes uncertainty from repeated neural-network fitting.

\subsection{France--Brazil 2022 cross-country replication}

The French and Brazilian source data were collected through an online platform around the 2022 presidential elections \citep{Navarrete2024}. Production cleaning retained a stable source index and UTC timestamp and required nonmissing UUID, proposal pair, selected item, and time; two distinct catalogue-backed proposals; a selected value equal to one of the pair; \texttt{suspicious=0}; the latest valid observation per UUID and unordered proposal pair; and the latest usable profile with political self-placement 1--5, age 1--7, and documented sex code 1--2. Age was released in bands, so the complete 10--19 category was excluded because ages 18--19 could not be separated from younger respondents. The primary confirmed-adult C20 cohorts contain 843 UUIDs and 16,860 decisions in France and 434 UUIDs and 8,680 decisions in Brazil; all 120 French and 67 Brazilian proposals remain represented. Both training comparison graphs are connected.

The broader confirmed-adult C10 cohorts contain 847 UUIDs and 8,470 rows in France and 515 UUIDs and 5,150 rows in Brazil. Every C20 participant is contained within C10, and the first ten decision identities match exactly for those shared participants. C20 is primary, with decisions 1--18 for training and 19--20 for validation. C10 uses decisions 1--8/9--10. C10 changes cohort composition, history length, training volume, and held-out positions together, so it is a descriptive sensitivity rather than a causal decision-count treatment.

The three focal checkpoints---Gemma 4-12B, Qwen3-32B, and Ministral 3-14B---used the same revisions and temperature-zero, stateless settings as the Chilean factorial. Country-specific prompts used native proposal text, the exact interface question (\emph{Quelle proposition pr\'{e}f\'{e}rez-vous ?} or \emph{Qual proposta voc\^{e} prefere?}), recorded political self-placement, and the 2022 temporal context. Positions 1--2 were grouped as left/center-left, 3 as center, and 4--5 as center-right/right; every scale was interpreted only within country. The $3\times2\times2$ design crossed self-placement present/ablated with original/inverted content order in each country. Across the generated union needed for C20 and C10 before the confirmed-adult restriction, France contributed 210,000 responses and Brazil 118,560, totaling 328,560. Six French catalogue entries were multichoice parent cards whose selected suboption is not recoverable; prompts conservatively used the original parent text rather than constructing an unobserved composite stimulus.

Every valid released human row in both countries records the selected proposal as \texttt{option\_a}. Thus, ``original'' denotes the released file orientation, not reconstructed historical screen position. Model responses were first decoded from visible A/B to proposal identity. Each pair was then represented as lower-ID versus higher-ID proposal before DYNAMAP estimation, preventing the selected-first convention from becoming a perfect human label. Original-order human agreement alone is not interpreted as fidelity. The primary direct-agreement estimand is the equal-weight mean across the four self-placement-cue-by-order cells; paired original/inverted proposal-identity consistency is a separate position-stability estimand.

All 52 French and Brazilian human or synthetic DYNAMAP fits use two dimensions, seed 101, 20 epochs, Adamax learning rate 0.01, batch size 64, and deterministic row order. Self-placement recoverability excludes midpoint 3 and uses random forests with 2,000 trees, entropy splits, seed 42, the fixed unstratified 30\% holdout, and stratified CV-10. In C20 this leaves 597 left and 137 right French participants and 301 left and 89 right Brazilian participants. The held-out majority rules are 179/221 (0.810) in France and 92/117 (0.786) in Brazil. We therefore report raw accuracy together with balanced accuracy (mean class recall), macro-F1, and ROC AUC. Proposal-anchor similarity alignment is performed within country only. UUID-bootstrap intervals use 2,000 resamples. Geometry intervals resample fixed fitted coordinates and therefore omit DYNAMAP-refitting uncertainty.

\subsection{Additional robustness estimands}

All primary and sensitivity estimands reported here were recomputed on the confirmed-adult manifests. Stored LLM responses were projected to those manifests; DYNAMAP and downstream statistical models were refitted on identical human and synthetic UUID sets within each country and cohort. The policy-domain analysis used the archived Chilean proposal taxonomy, retained both endpoints of cross-domain comparisons, and applied explicit minimum-coverage criteria. Delete-one-topic refits held the architecture, split, and primary initialization (seed 101) fixed and were composition diagnostics rather than proposal-population intervals.

Optimization sensitivity varied only DYNAMAP initialization across 30 fixed seeds for the human C20 data and the three focal temperature-zero, self-placement-present, original-order conditions in each country. Human and synthetic spaces were refitted at the same seed before alignment. These distributions measured neural-optimization sensitivity and not LLM-generation, participant-sampling, or proposal-sampling uncertainty.

Embedding-independent proposal analyses used L2-regularized Bradley--Terry logistic models without an intercept, with primary $C=1$ and $C=0.1$ and $C=10$ sensitivities \citep{Bradley1952}. Aggregate proposal summaries compared ranks, scores, pairwise orderings, and root-mean-square score dispersion. Political cleavage was estimated by fitting left and right groups separately, centering utilities within group, and defining cleavage as right-group minus left-group proposal utility. Paired participant-cluster bootstrap replicates resampled complete histories and applied identical multiplicities to human and synthetic responses. Percentile intervals summarize the central 95\% of refitted estimates; for discrete or skewed statistics, the full-sample estimate need not lie inside that central bootstrap interval.

\subsection{Verbal endpoint-anchor diagnostic}

The endpoint diagnostic was specified after the bare-numeric cleavage results were observed and is therefore an outcome-informed mechanism probe. It used only Qwen3-32B at temperature zero, stateless decoding, and the confirmed-adult C20 cohorts: 1,991 Chilean, 843 French, and 434 Brazilian participants. Original- and inverted-order cells were paired one-to-one with the existing bare-numeric controls. The six new inference cells contained 130,720 valid A/B responses in total. Apart from the endpoint text reproduced above, checkpoint revision, runtime, numerical format, system message, persona, temporal context, proposal IDs and text, decoding parameters, output schema, and response parser were unchanged. Inverted responses were decoded to original proposal identity before analysis.

Behavioral effects and order stability used 2,000 paired UUID-cluster bootstrap replicates. Political cleavage used the same $C=1$ Bradley--Terry estimator and 2,000 paired left/right UUID-cluster bootstrap refits as the bare analysis. DYNAMAP used the same two-dimensional architecture, decisions 1--18/19--20 split, 20 epochs, and national proposal-anchor alignment as the primary analysis. Seed 101 is the manuscript-compatible point estimate. Thirty additional initialization seeds refitted bare and anchored original/inverted responses and aligned each model fit to the same-seed human fit, yielding 465 total fits across 15 series when human references and verified reused fits are included. These seed distributions quantify optimization sensitivity conditional on fixed responses; they are not participant-sampling intervals or model-generation uncertainty.

\section{Chilean grid and factorial results}

\subsection{Complete 27-condition results}

Table~\ref{tab:s_all} reports every condition. Dyn. is raw DYNAMAP validation accuracy; Lift subtracts the condition-specific majority rule; Human is matched human agreement; Self-pl. is held-out left--right self-placement accuracy; L--R and Var. are percentages of the human reference; and $r_{dist}$ is proposal-distance correlation. These geometry values use the primary DYNAMAP initialization (seed 101). Exact unrounded estimates, CV-10 accuracy, coordinate gain, choice balance, entropy, alignment diagnostics, and bootstrap intervals accompany the table.

\begin{table*}[!t]
\caption{Complete model grid.}
\label{tab:s_all}
\centering
\scriptsize
\resizebox{\textwidth}{!}{\input{source_tables/table_all_conditions.tex}}
\end{table*}

\begin{figure*}[p]
\centering
\includegraphics[width=0.98\textwidth]{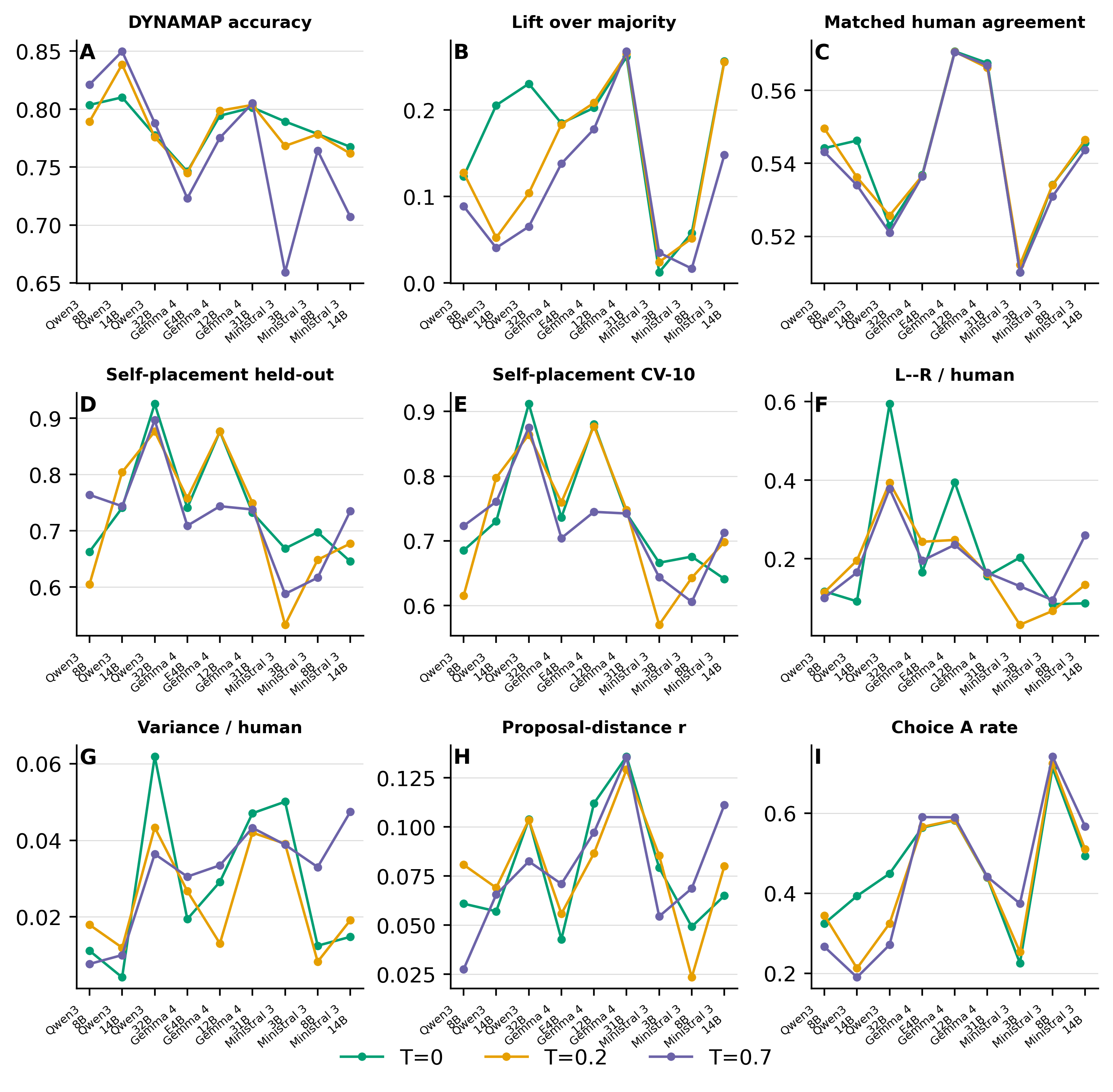}
\caption{Nine-panel view of the complete confirmed-adult grid. Panels show raw DYNAMAP accuracy, lift over majority, matched human agreement, held-out and CV-10 political self-placement accuracy, left--right separation, total variance, proposal-distance correlation, and A-response rate. Lines connect sizes within family and temperature only to aid reading.}
\label{fig:s_grid}
\figalttext[Nine metric panels for 27 model conditions]{Each metric varies nonmonotonically across families and sizes. Temperature lines often cross. All variance and left-right separation points lie far below the human reference.}
\end{figure*}

\begin{figure*}[p]
\centering
\includegraphics[width=0.98\textwidth]{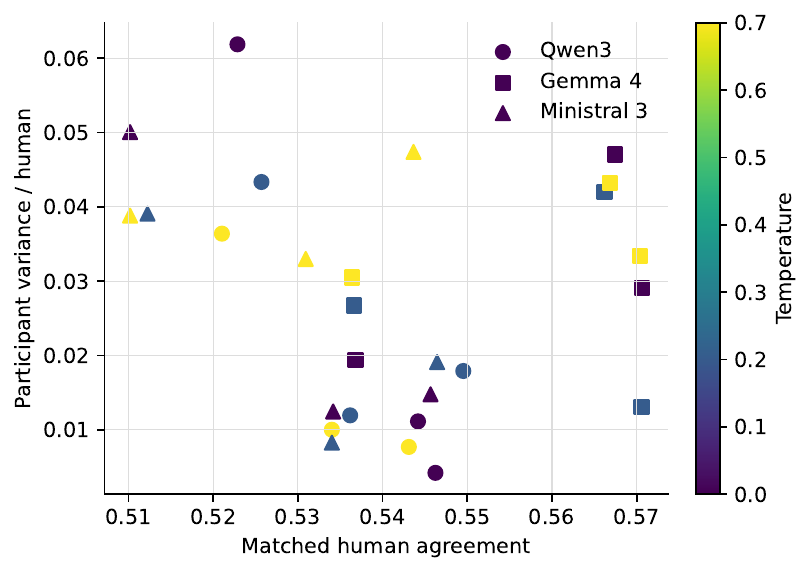}
\caption{Matched human agreement versus participant-coordinate variance relative to the confirmed-adult human reference across the complete Chilean grid. Marker shape denotes model family and color denotes temperature. Every condition remains far below the human variance reference of one.}
\label{fig:s_tradeoff}
\figalttext[Fidelity and participant-variance scatterplot]{Matched human agreement varies within a narrow range while every synthetic participant-coordinate variance remains far below the adult human reference.}
\end{figure*}

The adult-primary criteria have distinct maxima: DYNAMAP accuracy, 0.850 (Qwen3-14B, $T=0.7$); lift, 0.268 (Gemma 4-31B, $T=0.7$); matched human agreement, 0.571 (Gemma 4-12B, $T=0$); held-out self-placement recoverability, 0.925 (Qwen3-32B, $T=0$); CV-10 recoverability, 0.912 (Qwen3-32B, $T=0$); seed-101 left--right separation, 59.4\% of human (Qwen3-32B, $T=0$); seed-101 total variance, 6.19\% of human (Qwen3-32B, $T=0$); and seed-101 proposal-distance correlation, 0.136 (Gemma 4-31B, $T=0$). Figures~\ref{fig:s_grid} and \ref{fig:s_tradeoff} visualize the complete grid and the fidelity--compression tradeoff. These point maxima are descriptive and should not be interpreted as initialization-invariant checkpoint rankings.

\subsection{Factorial intervention results}

\begin{figure*}[p]
\centering
\includegraphics[width=0.98\textwidth]{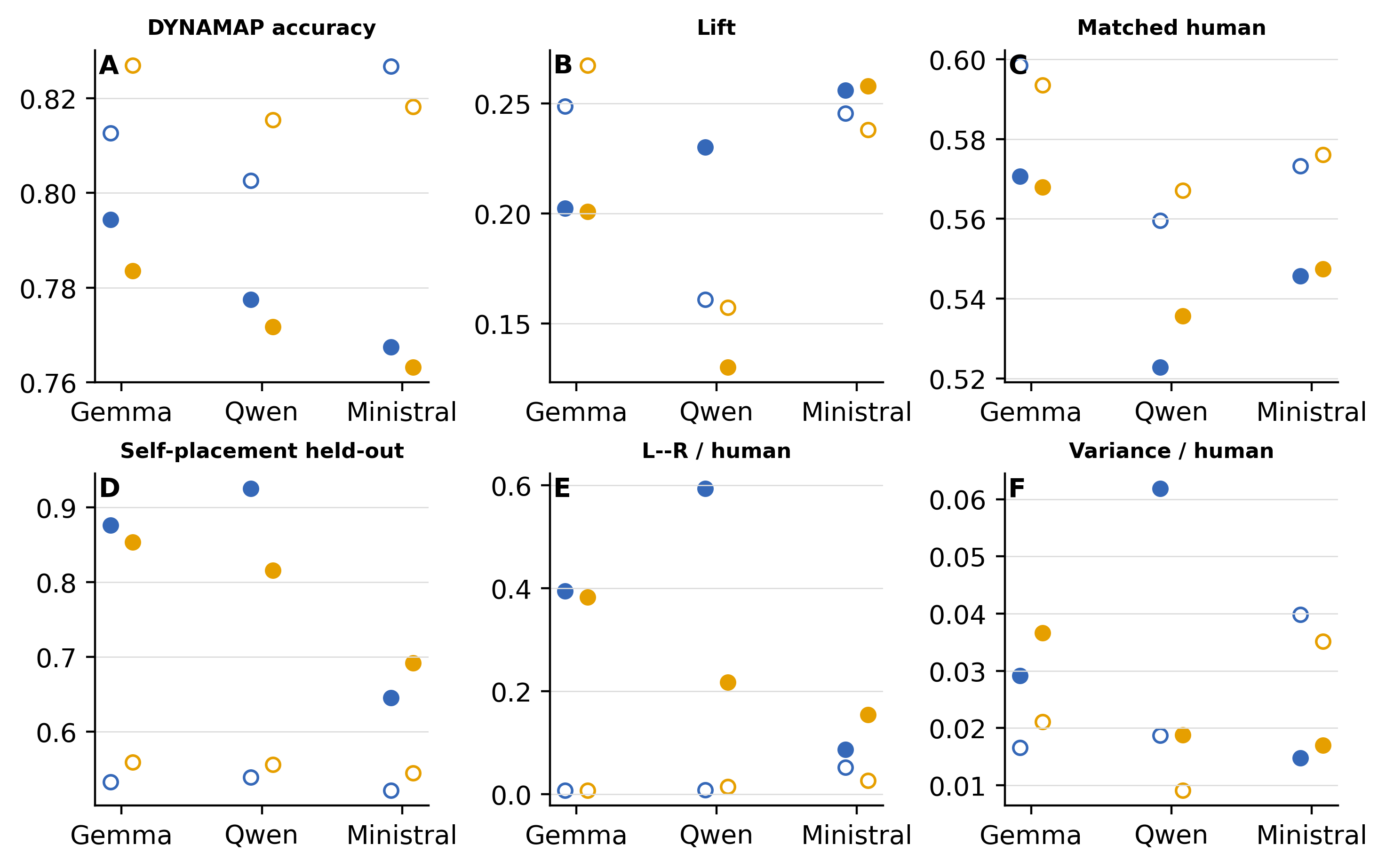}
\caption{Complete confirmed-adult factorial across reconstruction, lift, matched human agreement, self-placement recoverability, left--right separation, and total variance. Filled points retain self-placement in the prompt; open points ablate it. Blue denotes original display and orange inverted display, with inverted responses decoded before substantive metrics. Geometry values use the primary DYNAMAP initialization (seed 101).}
\label{fig:s_confirm_metrics}
\figalttext[Six panels comparing 12 factorial cells]{All cells remain reconstructable, but self-placement classification accuracy and left-right separation drop substantially after self-placement ablation. Matched human agreement often moves in the opposite direction.}
\end{figure*}

The factorial produced 12 complete adult-primary response spaces (Table~\ref{tab:s_confirm_cells}). DYNAMAP validation accuracy ranged from 0.763 to 0.827, its lift over the condition-specific majority rule from 0.130 to 0.267, matched human agreement from 0.523 to 0.598, held-out self-placement recoverability from 0.522 to 0.925, and CV-10 recoverability from 0.519 to 0.912. Figure~\ref{fig:s_confirm_metrics} places these quantities beside the aligned geometric diagnostics. The metrics do not share an optimum.

\begin{table*}[!t]
\caption{Complete confirmed-adult $3\times2\times2$ factorial. Cue denotes whether political self-placement was present or ablated. Order is the displayed proposal order; all substantive metrics for inverted cells use responses decoded to original proposal identity. Dyn. is DYNAMAP validation accuracy; Lift subtracts the validation-majority rule; Human is matched human agreement; Self-pl. and CV-10 are left--right self-placement classification accuracies; L--R and Var. are percentages of the human reference. Geometry values use the primary DYNAMAP initialization (seed 101). The 0.573 majority rule for the fixed adult holdout contextualizes the ablated cells.}
\label{tab:s_confirm_cells}
\centering
\scriptsize
\resizebox{\textwidth}{!}{\input{source_tables/table_confirmatory_cells.tex}}
\end{table*}

\subsubsection{Position stability}

Decoded cross-order agreement was 0.688--0.834 (Table~\ref{tab:s_confirm_position}). Thus, exchanging content and position changed 16.6--31.2\% of substantive choices. Qwen3-32B had 0.689 agreement with the self-placement cue present (95\% CI, 0.684--0.694) and 0.688 when it was ablated (95\% CI, 0.683--0.693). Gemma 4-12B was more stable after ablation (0.834; 95\% CI, 0.831--0.838) than with the cue present (0.779; 95\% CI, 0.774--0.784). Ministral 3-14B agreement was 0.786 with the cue and 0.780 without it. These are paired stability estimates, not agreement with human respondents.

\begin{table*}[t]
\caption{Substantive choice stability across original and inverted A/B displays. Brackets are UUID-bootstrap 95\% confidence intervals for decoded cross-order agreement.}
\label{tab:s_confirm_position}
\centering
\small
\input{source_tables/table_confirmatory_position.tex}
\end{table*}

Position sensitivity also varied across the 20 decision indices in the confirmed-adult histories (machine-readable sequence table). This heterogeneity shows that neither aggregate A-choice rates nor one overall stability coefficient can substitute for row-level and sequence-specific diagnostics.

\begin{figure*}[p]
\centering
\includegraphics[width=0.98\textwidth]{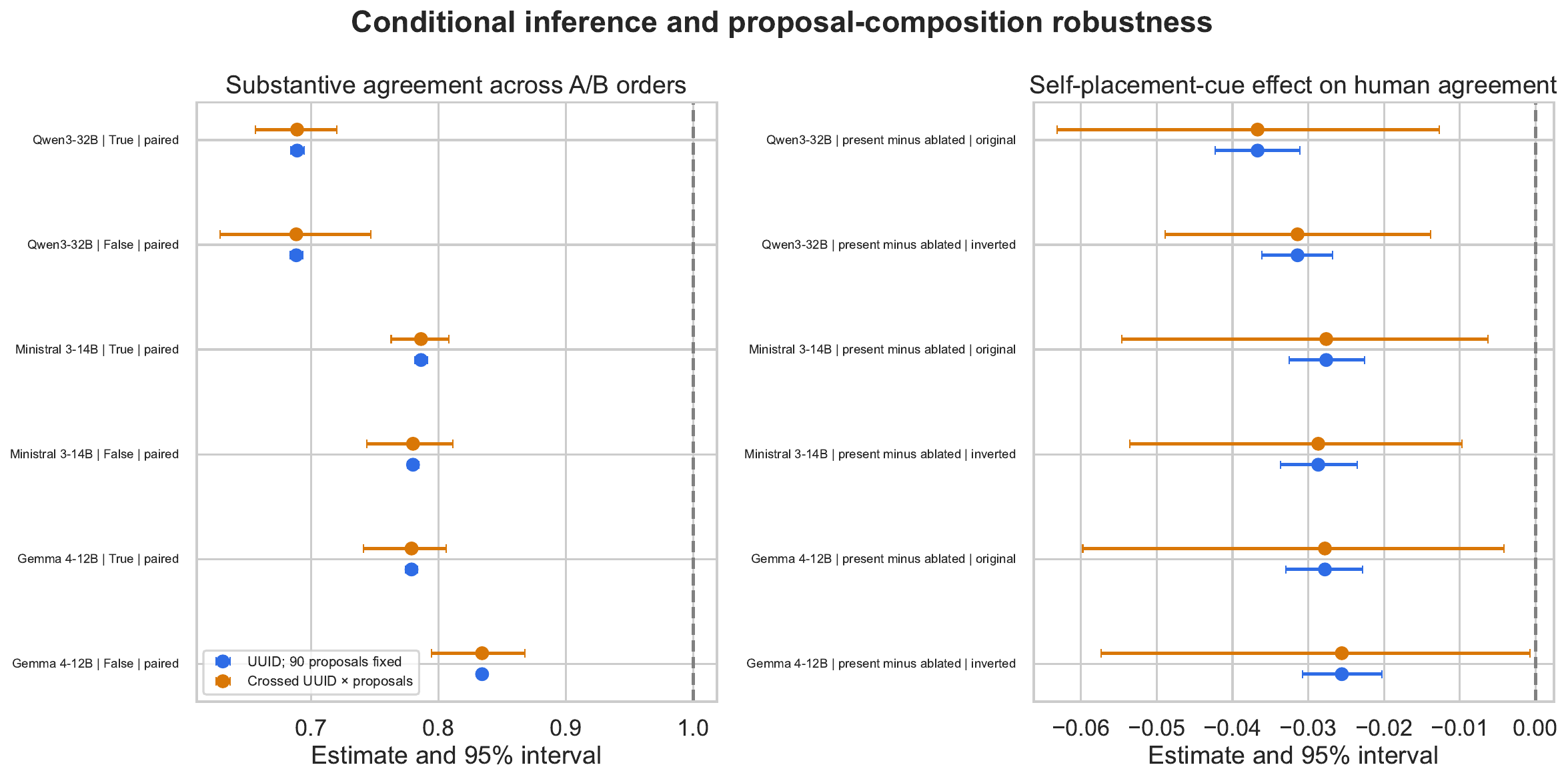}
\caption{Proposal-composition sensitivity for the two factorial contrasts summarized in the intervention results. The left panel shows decoded substantive agreement across original and inverted A/B displays. The right panel shows the effect of including political self-placement on matched human agreement. Blue intervals resample UUID histories while holding the 90 observed proposals fixed; orange intervals use crossed UUID $\times$ proposal-A $\times$ proposal-B Poisson-pigeonhole weights. The crossed intervals are an approximate robustness analysis under proposal interchangeability, not sampling inference for a known policy universe.}
\label{fig:s_crossed_uncertainty}
\figalttext[Two panels comparing UUID-only and crossed intervals]{Crossed proposal-composition intervals are wider than UUID-only intervals, but cross-order agreement remains below one and every self-placement-cue effect on matched human agreement remains below zero.}
\end{figure*}

The UUID bootstrap is conditional on the observed proposal set. Table~\ref{tab:s_crossed_position} adds crossed UUID $\times$ proposal-A $\times$ proposal-B Poisson-pigeonhole intervals. They are generally wider, as expected when proposal composition is allowed to vary under approximate interchangeability; Figure~\ref{fig:s_crossed_uncertainty} displays the position and self-placement-cue contrasts together. This robustness estimand does not turn the 90 proposals into a probability sample from a known universe.

\begin{table*}[t]
\caption{Decoded A/B-order stability with fixed-proposal UUID-only and proposal-composition crossed intervals. Crossed intervals use Poisson-pigeonhole UUID $\times$ proposal-A $\times$ proposal-B weights.}
\label{tab:s_crossed_position}
\centering
\small
\input{source_tables/table_crossed_position_stability.tex}
\end{table*}

\subsubsection{Political self-placement cue effects}

\begin{figure*}[p]
\centering
\includegraphics[width=0.98\textwidth]{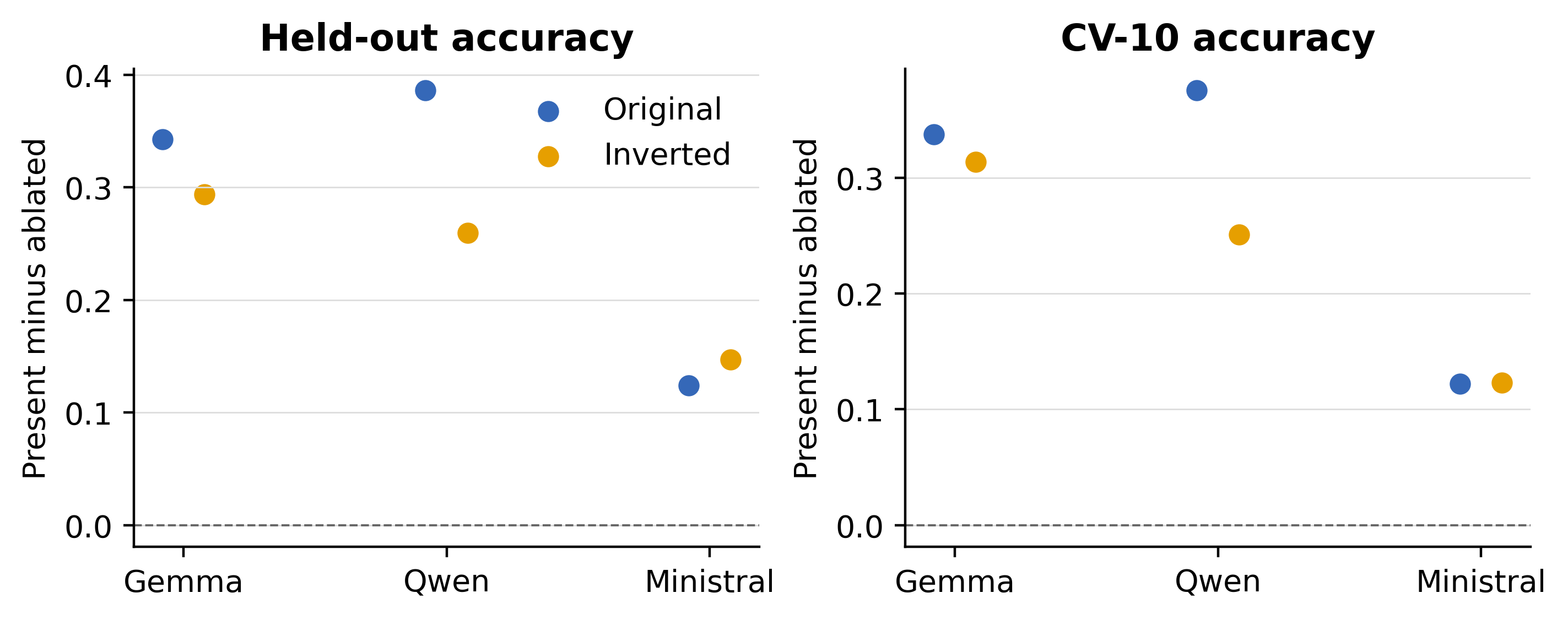}
\caption{Effect of including political self-placement on held-out and CV-10 left--right classification accuracy. Points show present-minus-ablated contrasts separately for original and inverted proposal order; the corresponding uncertainty summaries are reported in Table~\ref{tab:s_confirm_prompt}.}
\label{fig:s_prompt_effects}
\figalttext[Two panels of positive self-placement-cue effects]{Every model and order shows higher political self-placement classification accuracy when self-placement is included in the prompt. The largest held-out increase is for Qwen3-32B in original order.}
\end{figure*}

Including political self-placement increased held-out accuracy by 0.124--0.386 in all six checkpoint-by-order comparisons (Table~\ref{tab:s_confirm_prompt}). Effects were 0.343 and 0.294 for Gemma 4-12B under original and inverted orders, 0.124 and 0.147 for Ministral 3-14B, and 0.386 and 0.259 for Qwen3-32B. Paired-fold CV-10 differences were also positive (0.122--0.375; Fig.~\ref{fig:s_prompt_effects}), but folds are not independent populations.

Matched human agreement moved oppositely. Including self-placement reduced agreement by 0.026--0.037 across all six comparisons, and each primary UUID-bootstrap interval excluded zero. Crossed proposal-composition intervals retain this direction (Table~\ref{tab:s_crossed_ideology}). This intervention therefore strengthens recoverability of the supplied category while slightly reducing correspondence with the matched person's observed choice.

\begin{table*}[t]
\caption{Effects of including political self-placement, computed as present minus ablated within each checkpoint and A/B order. Classification intervals pair held-out test UUIDs; human-agreement intervals resample paired UUID histories.}
\label{tab:s_confirm_prompt}
\centering
\small
\input{source_tables/table_confirmatory_prompt.tex}
\end{table*}

\begin{table*}[t]
\caption{Matched-human-agreement effect of including political self-placement (present minus ablated), with primary UUID-only and crossed proposal-composition intervals. The UUID-only intervals are the same archived bootstrap realization used in Table~\ref{tab:s_confirm_prompt}; the crossed intervals additionally vary proposal composition. All six crossed intervals lie below zero.}
\label{tab:s_crossed_ideology}
\centering
\small
\input{source_tables/table_crossed_ideology_effects.tex}
\end{table*}

\subsubsection{Behavioral concentration, rankings, and geometry}

Across adult-primary factorial cells, aggregate decoded entropy ranged from 0.926 to 1.000 bits and the majority-choice concentration from 0.504 to 0.658. These are descriptive response-concentration diagnostics, not subgroup effect estimates.

Aggregate proposal ordering is evaluated in the focal confirmed-adult analysis reported below. That analysis places the temperature-zero, self-placement-present, original-order checkpoints from all three countries on one Bradley--Terry estimator, refits adult human and model rankings in paired UUID-bootstrap samples, and propagates rank uncertainty. We do not use descriptive rankings from the intervention cells as an additional model-selection criterion.

\begin{figure*}[p]
\centering
\includegraphics[width=0.98\textwidth]{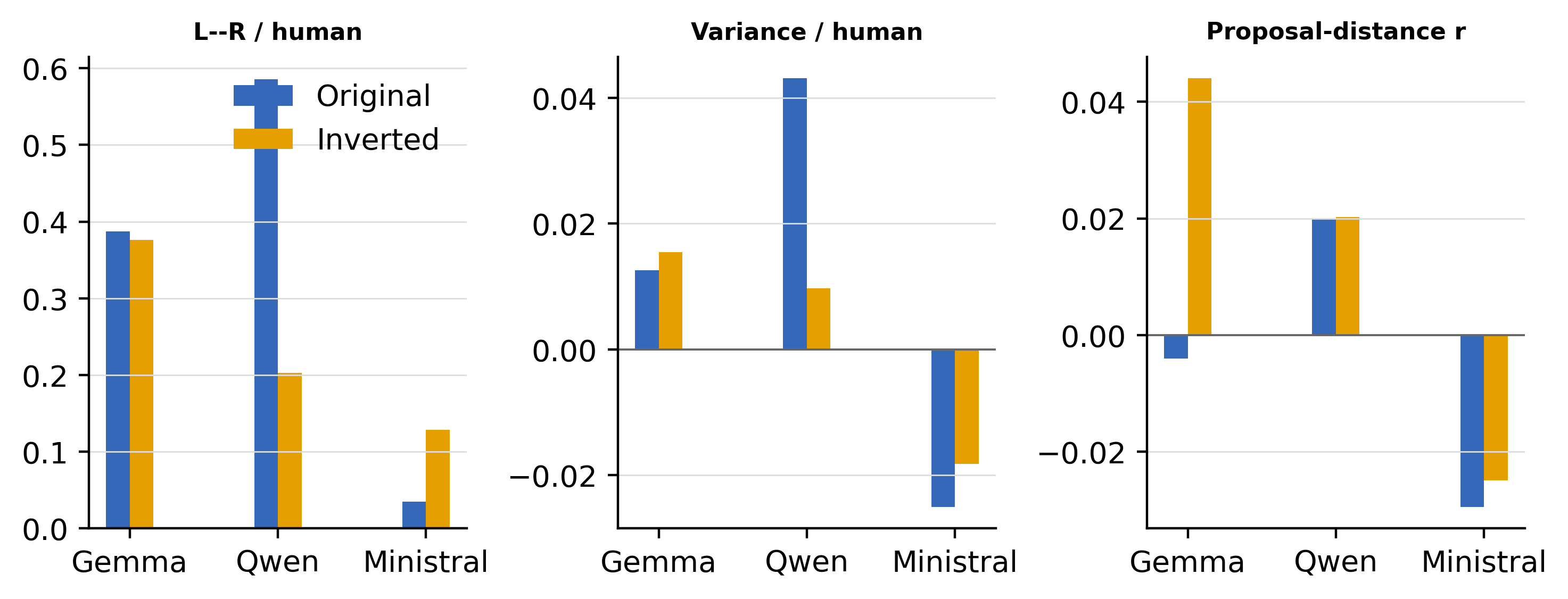}
\caption{Contribution of the political self-placement cue to aligned latent geometry, shown as present minus ablated. Left--right separation and total variance are in human-normalized units; proposal-distance correlations compare aligned proposal topology. Bars are post-fit contrasts and do not include repeated DYNAMAP refitting uncertainty.}
\label{fig:s_geometry_ablation}
\figalttext[Three panels of geometry contrasts]{The self-placement cue increases left-right separation in every model and order, while effects on total variance and proposal-distance correlation vary in sign and magnitude.}
\end{figure*}

Including political self-placement consistently increased left--right centroid separation, but effects on variance and proposal-distance correlation varied by model and order (Fig.~\ref{fig:s_geometry_ablation}). Geometry intervals are conditional post-fit diagnostics. In particular, their narrowness must not be interpreted as incorporating DYNAMAP refitting uncertainty.

\section{Human benchmarks and policy-instrument sensitivity}

\subsection{Held-out human-choice benchmarks}

\begin{figure*}[p]
\centering
\includegraphics[width=0.98\textwidth]{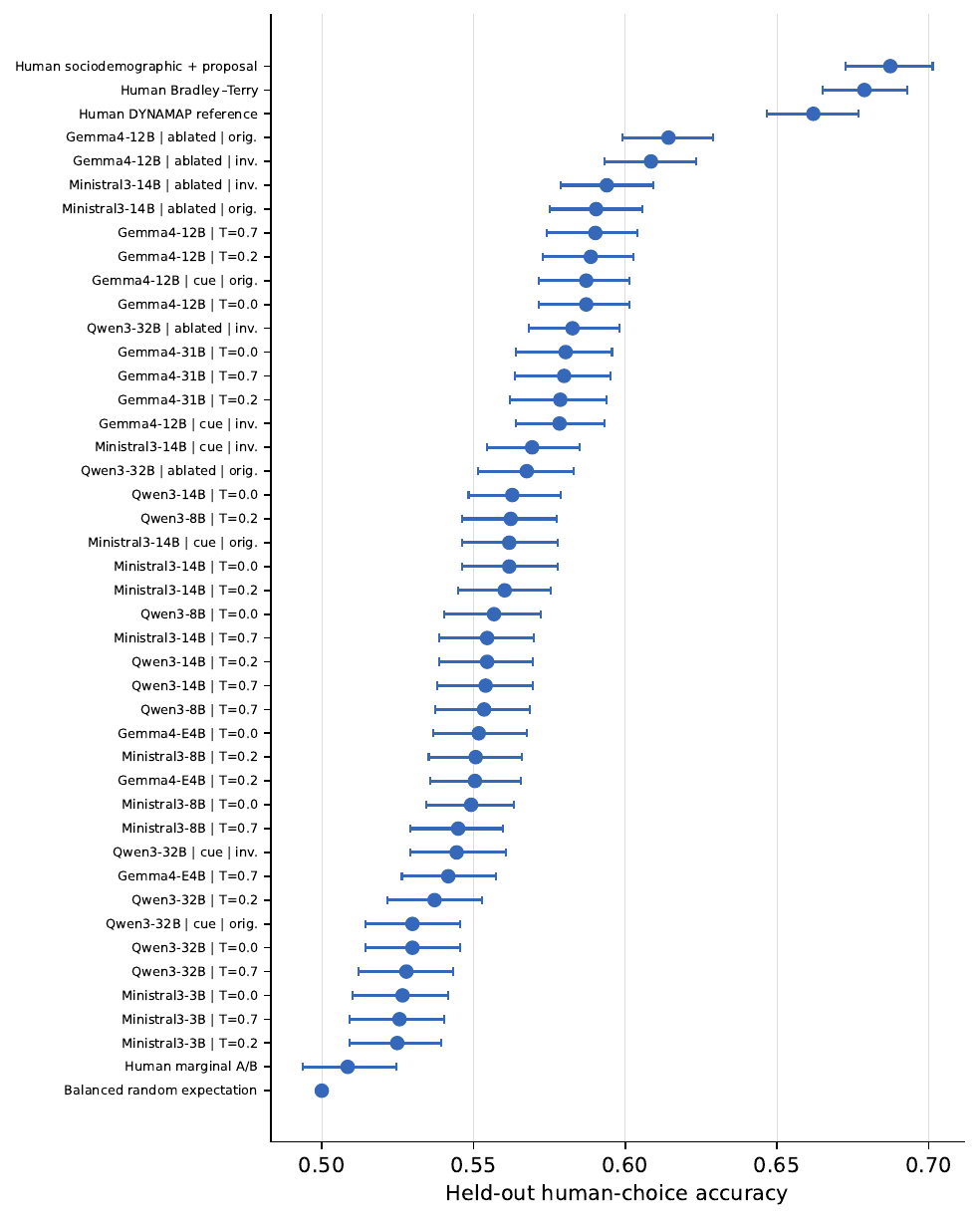}
\caption{Held-out human-choice benchmarks. Human predictors and LLM conditions are evaluated on the identical held-out choice set. The human DYNAMAP line is a structural reference for human predictability out of sample, not a human ceiling.}
\label{fig:s_benchmark_ladder}
\figalttext[Held-out benchmarks for human and LLM predictors]{Three fitted human predictors exceed all factorial and comparable grid LLM conditions on the same held-out choices.}
\end{figure*}

All benchmark predictors were trained on decisions 1--18 and evaluated on the same 3,982 confirmed-adult decisions 19--20. The references contextualize matched human agreement rather than define a normative performance ceiling (Table~\ref{tab:s_benchmark_ladder}; Fig.~\ref{fig:s_benchmark_ladder}). Balanced random choice equals 0.500 by construction and therefore has no sampling interval. The factorial LLM range was 0.530--0.614 and the comparable 27-condition grid range was 0.510--0.571, below the fitted proposal-popularity, sociodemographic-plus-pair, and human-DYNAMAP predictors. Human DYNAMAP is an out-of-sample structural reference, not a maximum that an LLM ought to attain.

\begin{table*}[t]
\caption{Held-out human-choice benchmarks evaluated on the common decision set. Confidence intervals resample UUID histories.}
\label{tab:s_benchmark_ladder}
\centering
\small
\input{source_tables/table_human_benchmark_ladder.tex}
\end{table*}

\subsection{Policy-domain heterogeneity}

The expert taxonomy is the archived Chilean proposal-level field \texttt{topic\_name}. All 90 proposal identifiers matched. The only normalizations were typography (capitalization or whitespace), with no semantic merging. Both topic endpoints are retained for every decision, so cross-topic pairs are not assigned arbitrarily to one domain. The confirmed-adult sample contains 3,367 within-domain and 36,453 between-domain decisions. Five domains satisfied the minimum coverage criteria of at least 250 decisions, 200 UUIDs, and five proposals: Econom\'{i}a, Educaci\'{o}n, Modernizaci\'{o}n del Estado, Salud, and Vivienda y Transporte (Table~\ref{tab:s_topic_coverage}).

\begin{figure*}[p]
\centering
\includegraphics[width=0.98\textwidth]{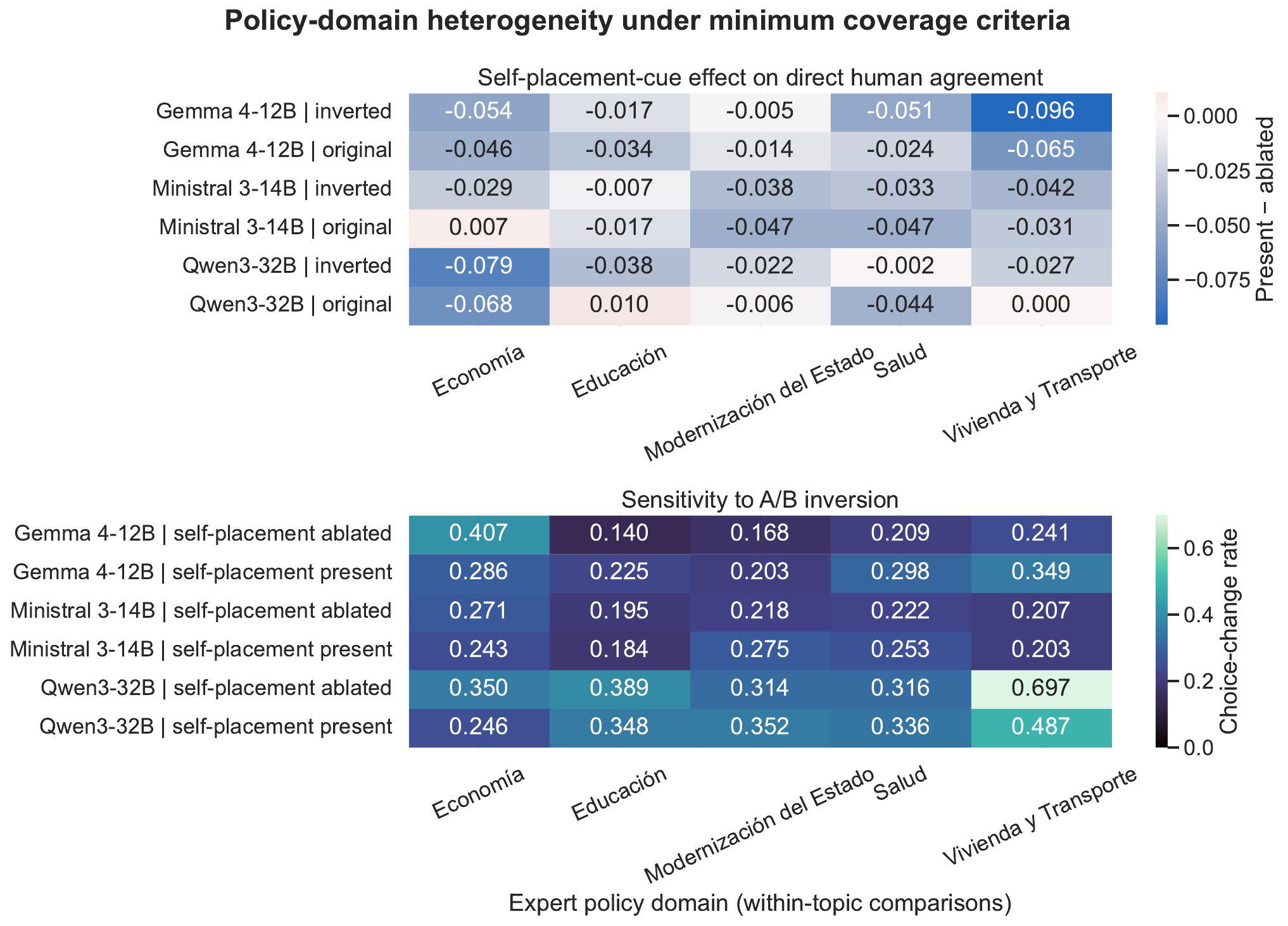}
\caption{Within-topic heterogeneity in behavior. The display includes only the five domains meeting the minimum coverage criteria.}
\label{fig:s_topics}
\figalttext[Topic heterogeneity heatmap]{Behavioral fidelity, self-placement-cue effects, and position sensitivity differ across expert-coded policy topics, with low-coverage strata marked.}
\end{figure*}

\begin{figure*}[p]
\centering
\includegraphics[width=0.98\textwidth]{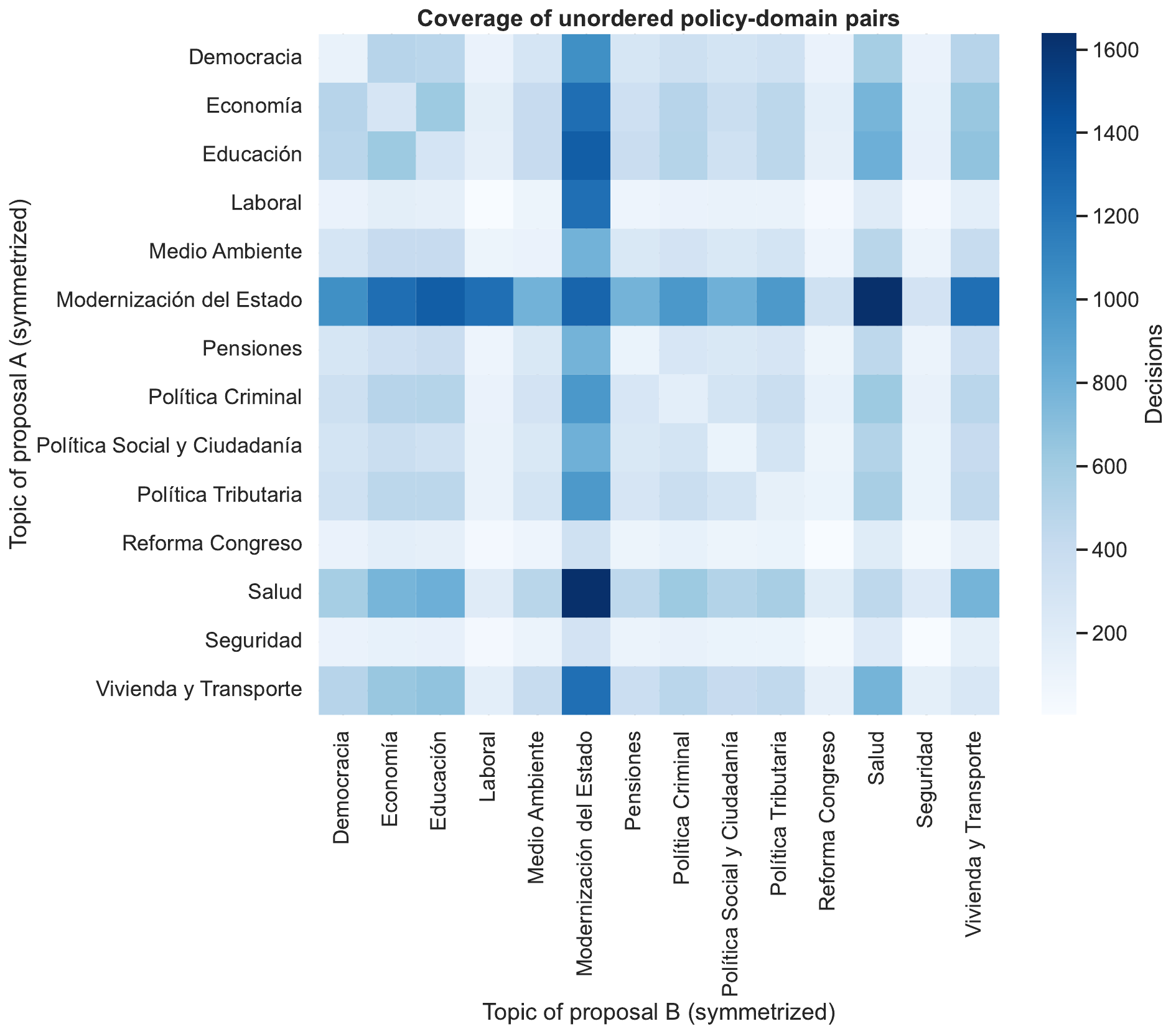}
\caption{Coverage of symmetrized, unordered policy-domain pairs. This display retains both endpoints rather than assigning between-domain decisions to one category.}
\label{fig:s_topic_pairs}
\figalttext[Topic-pair coverage heatmap]{A heatmap displays sample coverage for pairs of expert-coded policy topics.}
\end{figure*}

Within-topic direct fidelity, self-placement-cue effects, and A/B instability vary across these strata (Fig.~\ref{fig:s_topics}). Figure~\ref{fig:s_topic_pairs} documents coverage across all 105 unordered topic pairs. The complete within-topic table, sample sizes, Holm correction, low-coverage flags, and exploratory ridge model with both endpoints are preserved in the machine-readable source tables. Effects are descriptive of policy domains in this Chilean instrument; their magnitude is not an index of cultural importance.

\begin{table*}[t]
\caption{Coverage of the expert-coded proposal topics. The minimum display criteria required at least 250 decisions, 200 UUIDs, and five proposals.}
\label{tab:s_topic_coverage}
\centering
\scriptsize
\input{source_tables/table_topic_coverage_compact.tex}
\end{table*}

\subsection{DYNAMAP sensitivity to topic composition}

\begin{figure*}[p]
\centering
\includegraphics[width=0.98\textwidth]{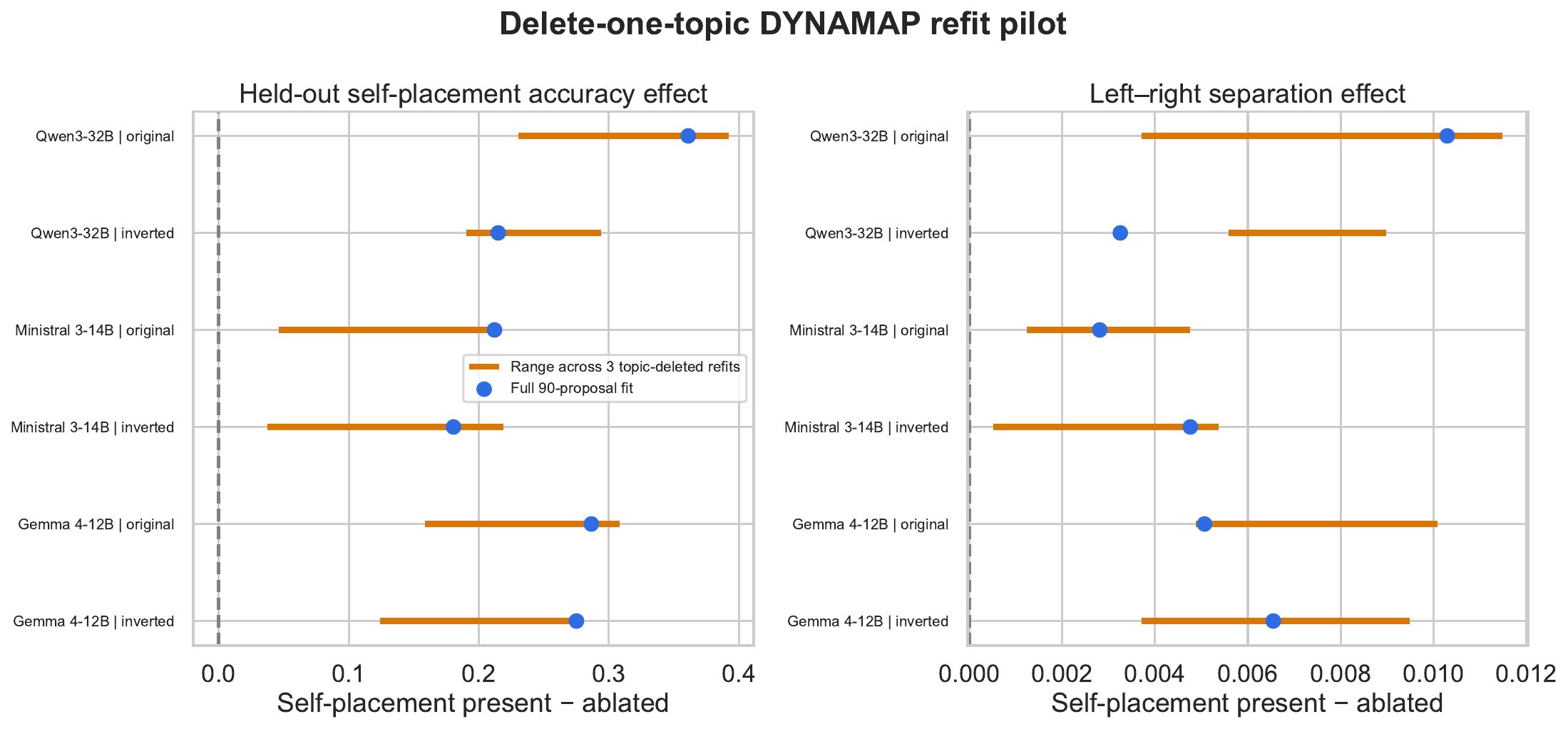}
\caption{Delete-one-topic DYNAMAP refit sensitivity pilot. The 39 refits hold architecture, seed, and temporal split fixed. Proposal-distance topology is comparatively sensitive to topic composition; the display is diagnostic rather than a population confidence interval.}
\label{fig:s_delete_topic}
\figalttext[Topic-deletion DYNAMAP sensitivity plot]{Removing one of three high-exposure policy topics changes some geometry metrics, especially proposal-distance correlations, while central contrast directions generally persist.}
\end{figure*}

Thirty-nine delete-one-topic refits remove Modernizaci\'{o}n del Estado, Salud, or Educaci\'{o}n from the adult human space and each of the 12 adult factorial conditions, using the same architecture, seed, and temporal split. These were the three eligible within-topic domains with the largest confirmed-adult decision counts (1,300, 450, and 293, respectively); selection used coverage only and not DYNAMAP outcomes. Directions of the central contrasts generally persist, but proposal-distance correlations are especially sensitive to which policy domains are retained (Fig.~\ref{fig:s_delete_topic}). This is a sensitivity pilot, not an interval for a population of topic compositions.

\section{France--Brazil cross-country replication}

\subsection{Primary C20 results}

The replication contains 24 complete checkpoint-by-cue-by-order response cells. The stored inference generated 328,560 responses over the union needed for C20 and C10 before the confirmed-adult restriction; the primary confirmed-adult C20 analysis uses 306,480 of those responses. Decision keys join one-to-one to the fixed human rows, inverted displays exchange both proposal identity and position, and all decoded responses return to a common proposal-identity representation before analysis. The 52 human and synthetic DYNAMAP fits use the same architecture and country-specific preprocessing described above. Table~\ref{tab:s_crosscountry_all_cells} reports all 24 primary C20 cells underlying the ranges in the main text.

\begin{table*}[p]
\caption{All 24 primary C20 cross-country cells. Cue indicates whether political self-placement was present or ablated. Dyn. acc. is DYNAMAP validation accuracy; Majority is the condition-specific held-out majority-label accuracy; Lift is their difference; Matched human is decoded agreement with the corresponding human choice; and Displayed A is the fraction of responses selecting the visible A position. Under the selected-first released order, the per-cell Displayed A is determined by matched agreement; the order mean is therefore the positional statistic, with 0.5 indicating no letter-position concentration. No treatment averaging is applied to the other columns.}
\label{tab:s_crosscountry_all_cells}
\centering
\scriptsize
\resizebox{\textwidth}{!}{\input{source_tables/table_crosscountry_c20_all_cells.tex}}
\end{table*}

Self-placement classification requires separate treatment of class imbalance. Table~\ref{tab:s_crosscountry_classification} reports the augmented sociodemographic-plus-coordinate classifier in every C20 cell. The held-out majority rule is 0.810 in France (179/221) and 0.786 in Brazil (92/117). French raw accuracy of 0.774--0.810 coexists with balanced accuracy of 0.480--0.582, macro-F1 of 0.438--0.595, and ROC AUC of 0.470--0.638. Brazilian results are heterogeneous: raw accuracy is 0.744--0.846, balanced accuracy 0.487--0.742, macro-F1 0.457--0.756, and ROC AUC 0.420--0.807. Raw accuracy alone therefore does not establish strong self-placement recoverability.

\begin{table*}[p]
\caption{Political self-placement classification in all 24 confirmed-adult primary C20 cells. Models use archived sociodemographic fields plus two DYNAMAP coordinates. Bal. acc. is the mean of left- and right-class recall; macro-F1 weights the two classes equally. Held-out sample sizes are 221 in France and 117 in Brazil; corresponding majority-rule accuracies are 0.810 and 0.786.}
\label{tab:s_crosscountry_classification}
\centering
\scriptsize
\resizebox{\textwidth}{!}{\input{source_tables/table_crosscountry_c20_classification.tex}}
\end{table*}

Table~\ref{tab:s_crosscountry_c20} reports treatment-balanced adult C20 estimates. DYNAMAP reconstruction is 0.761--0.793 after averaging the four treatment cells within checkpoint and country, and lift is 0.233--0.273. The corresponding agreement estimates are 0.651--0.687 in France and 0.572--0.636 in Brazil. The national adult human DYNAMAP references are 0.773 reconstruction and 0.262 lift in France and 0.778 reconstruction and 0.229 lift in Brazil. These equal-weight estimates are the primary fidelity summaries because original-order agreement alone is outcome-informed under the selected-first released-file convention.

\begin{table*}[p]
\caption{Primary confirmed-adult C20 treatment-balanced results. Estimates give equal weight to self-placement present/ablated and original/inverted cells. Human-reference rows report national DYNAMAP point diagnostics; matched-agreement confidence intervals resample complete UUID histories 2,000 times. Self-pl. is held-out left--right self-placement classification accuracy; $\Delta$ coord. is its gain over the sociodemographic model. For the human-reference rows, Self-pl. and $\Delta$ coord. are computed from the human DYNAMAP coordinates under the same random forest, fixed holdout, and sociodemographic baseline as the synthetic rows. L--R and Var. are fractions of the corresponding national human geometry; $r_{dist}$ is the within-country proposal-distance correlation; and $\rho_{BT}$ is the Bradley--Terry rank correlation with the national human ranking. Human-reference rows in Panel B are 1.000 by definition. French and Brazilian C20 majority rules are 0.810 and 0.786, respectively; raw self-placement accuracy should be read with Table~\ref{tab:s_crosscountry_classification}.}
\label{tab:s_crosscountry_c20}
\centering
\footnotesize
\textit{Panel A. Reconstruction, matched human agreement, and self-placement recoverability}\par\smallskip
\resizebox{\textwidth}{!}{\input{source_tables/table_crosscountry_c20_performance.tex}}
\medskip

\textit{Panel B. Participant geometry and proposal-space fidelity}\par\smallskip
\input{source_tables/table_crosscountry_c20_geometry.tex}
\end{table*}

\begin{figure*}[p]
\centering
\includegraphics[width=0.98\textwidth]{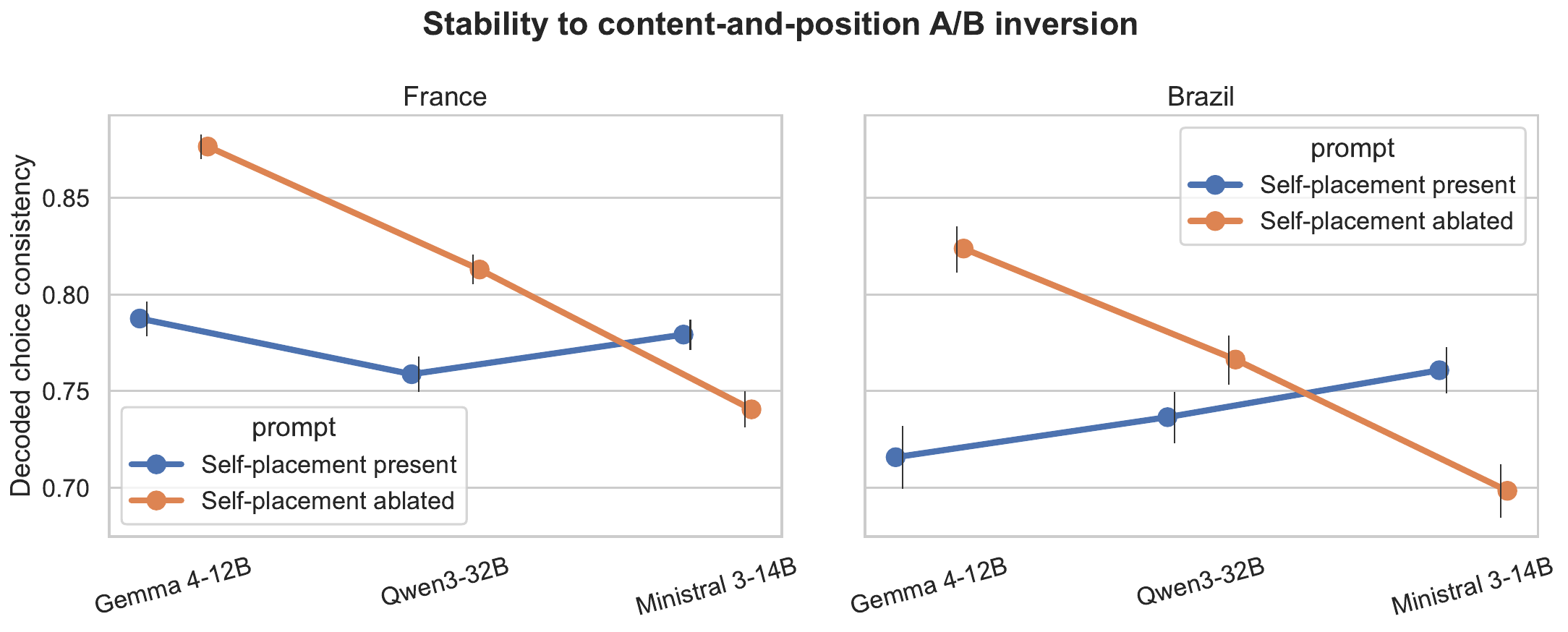}
\caption{Decoded proposal-identity consistency under original versus inverted content-and-position displays in France and Brazil. Points and UUID-bootstrap 95\% intervals are shown separately by checkpoint and self-placement-cue condition. This is position stability, not agreement with human respondents.}
\label{fig:s_crosscountry_position}
\figalttext[Two panels of France and Brazil position stability]{Identity consistency varies from roughly 0.70 to 0.88 across countries, checkpoints, and self-placement-cue conditions. Cue ablation increases consistency for Gemma and Qwen but not for Ministral.}
\end{figure*}

Paired decoded identity consistency ranged from 0.698 to 0.877 (Table~\ref{tab:s_crosscountry_ab}; Fig.~\ref{fig:s_crosscountry_position}), corresponding to substantive change rates of 0.123--0.302. Order-specific matched agreement differed sharply across checkpoints. The order-mean Displayed-A column in Table~\ref{tab:s_crosscountry_all_cells} showed opposite checkpoint-specific letter-position concentration. Because released original order is selected-first, the per-cell displayed-A rate is algebraically determined by matched agreement; the order mean is the positional statistic. These contrasts are not intrinsic policy-preference effects. They reinforce the decision to emphasize treatment-balanced levels and paired identity consistency rather than a single order-specific cell.

\begin{table*}[p]
\caption{Proposal-identity stability under content-and-position A/B inversion in C20. Consistency compares paired decoded substantive choices; change is one minus consistency. Confidence intervals resample UUID histories 2,000 times.}
\label{tab:s_crosscountry_ab}
\centering
\scriptsize
\resizebox{\textwidth}{!}{\input{source_tables/table_crosscountry_ab_stability.tex}}
\end{table*}

\begin{figure*}[p]
\centering
\includegraphics[width=0.98\textwidth]{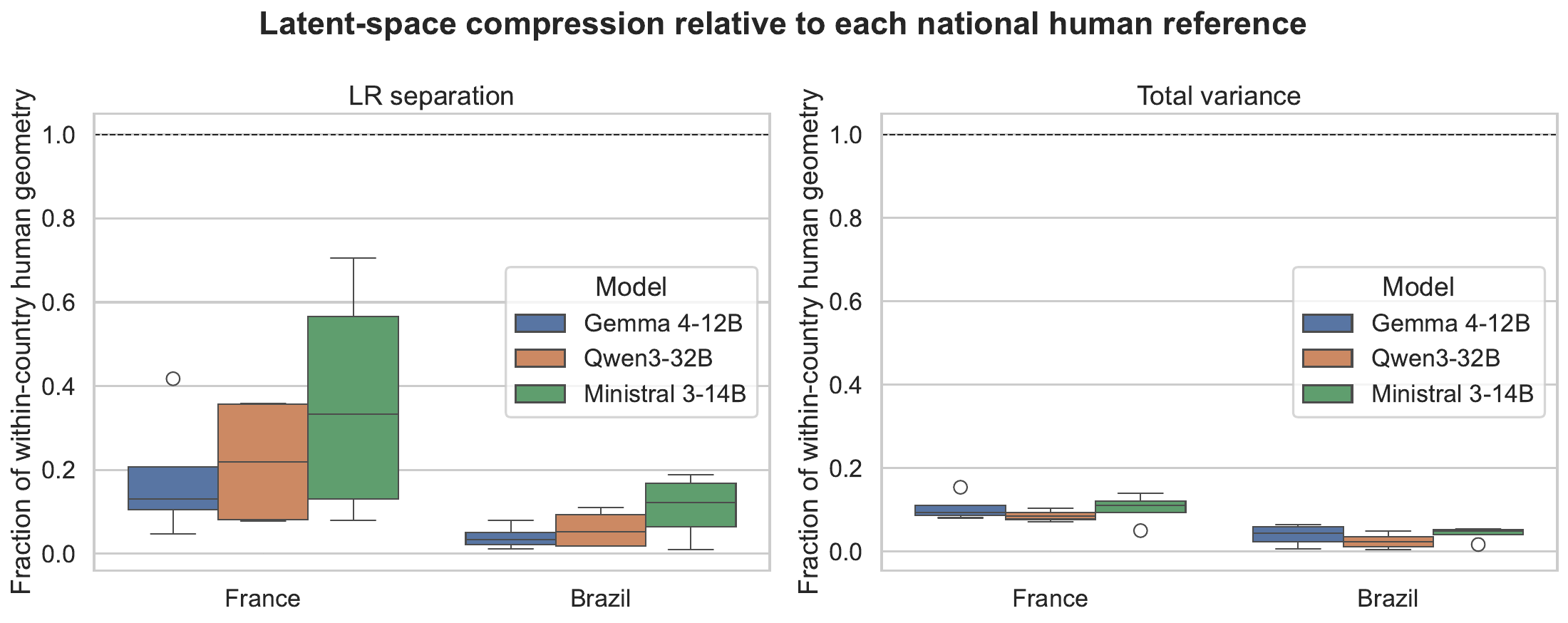}
\caption{Latent-space geometry relative to each national human reference. Boxplots summarize the four C20 self-placement-cue-by-order cells within checkpoint and country. The dashed line marks the national human value. France and Brazil were aligned separately; no point implies a shared DYNAMAP coordinate system across countries.}
\label{fig:s_crosscountry_geometry}
\figalttext[Two panels of national geometry fractions]{Synthetic total variance is far below one in every French and Brazilian condition. Left-right separation varies more, including two high French cells, but is consistently small in Brazil.}
\end{figure*}

Treatment-balanced coordinate variance is 0.086--0.105 of the French adult human reference and 0.025--0.043 of the Brazilian adult reference. Left--right separation is more context-variable: 0.182--0.363 of human in France and 0.040--0.111 in Brazil. These summaries average four separately fitted spaces and are not common cross-country coordinates. Figure~\ref{fig:s_crosscountry_geometry} shows the distribution of the four cell-specific fractions behind each treatment-balanced mean. Geometry intervals in the source tables condition on a single fitted network and must not be read as retraining uncertainty.

Aggregate proposal fidelity is incomplete even where matched human agreement is comparatively high. Treatment-balanced proposal-distance correlations are 0.241--0.311 in France and 0.141--0.178 in Brazil. Treatment-balanced Bradley--Terry rankings and summaries are included in the accompanying Source Data. Because averaging four fitted utility vectors is not the same estimand as refitting on one common fixed condition, inferential interpretation uses the paired analysis reported below rather than these descriptive treatment-balanced ranks.

\begin{figure*}[p]
\centering
\includegraphics[width=0.98\textwidth]{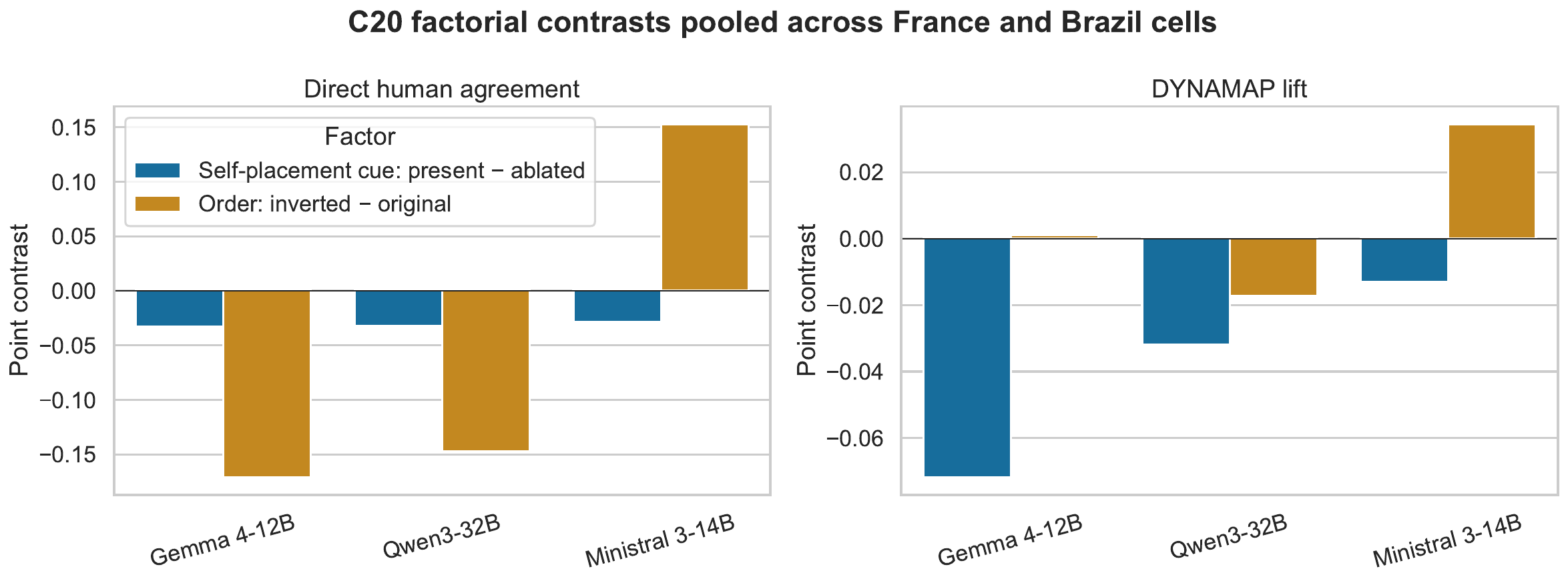}
\caption{Descriptive C20 factorial point contrasts pooled across French and Brazilian cells. Blue bars are self-placement present minus ablated and orange bars inverted minus original. The figure does not display uncertainty and does not override country- and order-specific UUID-bootstrap contrasts in the machine-readable tables.}
\label{fig:s_crosscountry_factorial}
\figalttext[Two panels of pooled point contrasts]{Average self-placement-cue contrasts are negative for matched human agreement and DYNAMAP lift. Average order contrasts are negative for Gemma and Qwen agreement but positive for Ministral agreement and lift.}
\end{figure*}

The pooled point contrasts in Fig.~\ref{fig:s_crosscountry_factorial} summarize direction rather than uncertainty: averaging across the French and Brazilian cells, including political self-placement reduces agreement and DYNAMAP lift for all three checkpoints, while the average inverted-minus-original contrast differs in sign for Ministral. Country-, order-, and prompt-specific tables should be used for substantive interpretation because pooling can conceal interactions.

Detailed sequence, subgroup, proposal, response-entropy, persistence, and cleavage diagnostics are provided as machine-readable Supplementary Source Data. These strata are descriptive; sparse region groups are not interpreted as stable subgroup effects.

\subsection{Broader C10 sensitivity}

\begin{figure*}[p]
\centering
\includegraphics[width=0.98\textwidth]{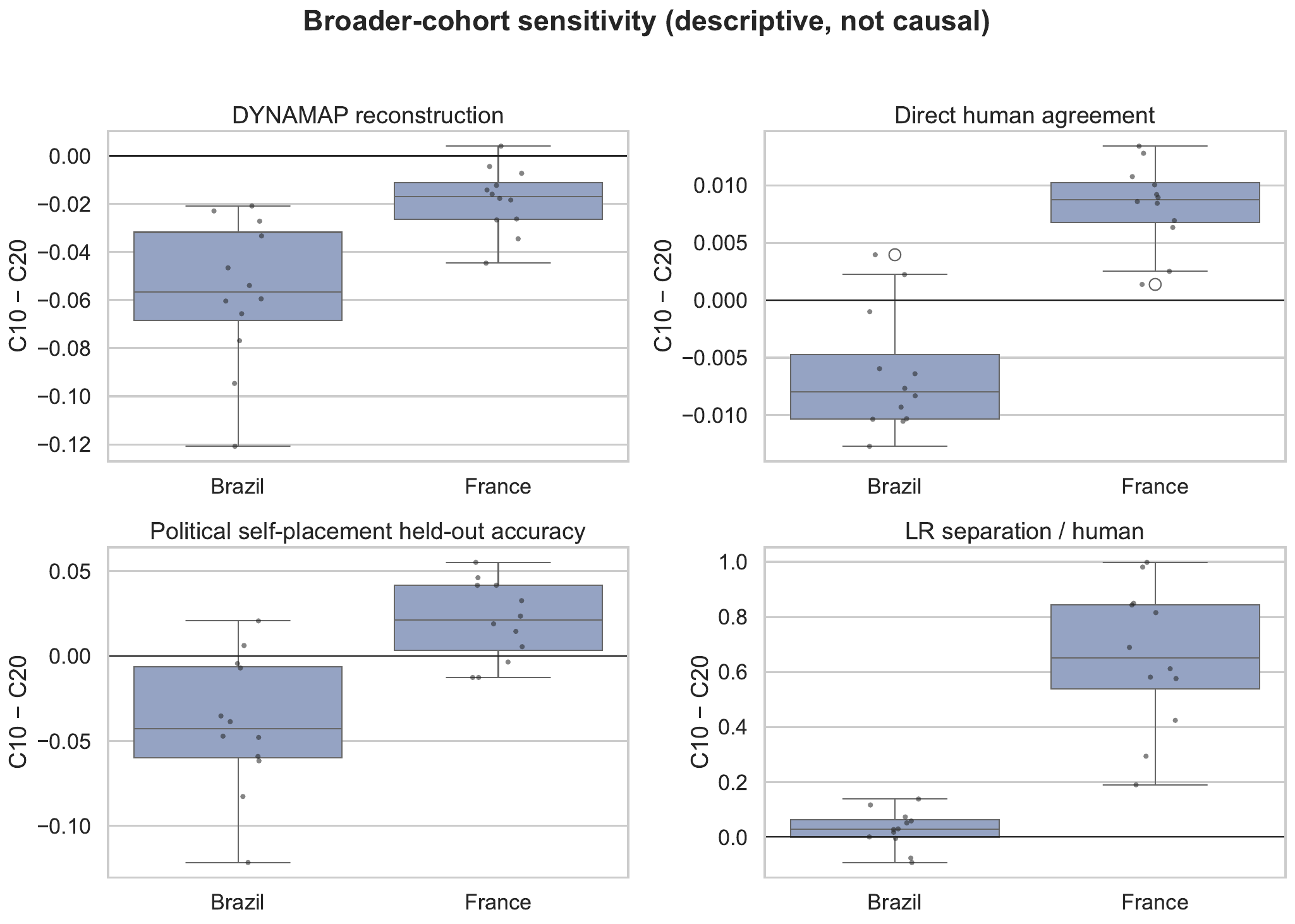}
\caption{Broader-cohort sensitivity, shown as C10 minus C20 across the 12 treatment cells per country. The panels display reconstruction, matched human agreement, self-placement held-out accuracy, and normalized left--right separation. Differences combine cohort composition, history length, training volume, and holdout position.}
\label{fig:s_crosscountry_c10}
\figalttext[Four panels comparing broader cohorts]{C10 reconstruction is generally lower. Matched human agreement changes are small and opposite in direction by country, while self-placement classification accuracy and left-right separation are more sensitive to cohort definition.}
\end{figure*}

The broader C10 specification modestly lowers treatment-balanced DYNAMAP reconstruction relative to C20 for every checkpoint and country. Treatment-balanced agreement differs by less than 0.014 in absolute value, decreasing in Brazil and increasing in France. Political recoverability and especially left--right separation are less stable across cohort definitions. Because participant eligibility, history length, training volume, and held-out position all change, Table~\ref{tab:s_crosscountry_c10} and Fig.~\ref{fig:s_crosscountry_c10} are descriptive sensitivity analyses, not factorial estimates.

\begin{table*}[p]
\caption{Treatment-balanced confirmed-adult C10 and C20 sensitivity. C20 is primary. Cohort differences are descriptive because C10 and C20 do not differ only in decision count. Raw self-placement accuracy must be read against its cohort-specific majority rule; C20 rules are 0.810 in France and 0.786 in Brazil (Table~\ref{tab:s_crosscountry_classification}).}
\label{tab:s_crosscountry_c10}
\centering
\footnotesize
\textit{Panel A. Reconstruction, matched human agreement, and self-placement recoverability}\par\smallskip
\input{source_tables/table_crosscountry_c10_c20_performance.tex}
\medskip

\textit{Panel B. Participant geometry and proposal topology}\par\smallskip
\input{source_tables/table_crosscountry_c10_c20_geometry.tex}
\end{table*}

\section{Additional sensitivity and independent validation}

\subsection{Temperature, response balance, and entropy}

Temperature means do not support monotonic fidelity improvement in the adult-primary grid. From $T=0$ to $T=0.7$, mean matched human agreement changes from 0.542 to 0.540, mean held-out self-placement recoverability from 0.743 to 0.726, mean DYNAMAP lift from 0.170 to 0.109, mean proposal-distance correlation from 0.078 to 0.079, and mean seed-101 variance fraction from 0.028 to 0.031. The variance increase is small relative to the missing human variance and does not show that temperature recovers human heterogeneity. Choice-A rates and aggregate response entropy vary substantially across conditions, demonstrating that similar raw accuracy can coexist with different response concentrations.

\subsection{Exploratory boundary check: sequential context}

The main persona-replay design supplies each policy choice without prior model responses. This boundary check tests one sequential-context alternative within a single Chilean checkpoint and implementation: a local, quantized Qwen3-8B Q4\_K\_M replay with temperature-zero decoding. Decision 1 used a byte-identical stateless message; decisions 2--20 prefixed an oldest-to-newest JSONL history containing only the model's previous alternatives and its own A/B choices. The quantization, inference engine, and response protocol differ from the primary grid. Absolute levels therefore cannot be compared with the bfloat16/vLLM grid; only sequential versus matched local stateless differences identify the effect within this replay. The analysis does not generalize across model families, scales, temperatures, countries, or the factorial and is not a general evaluation of memory or agents.

After projection to the confirmed-adult cohort, the stored replay contributed 39,820 valid decisions from 1,991 adults. Eleven memory-free first responses differed across otherwise identical requests (0.55\% of the adult cohort), reflecting run-to-run computational reproducibility rather than the sequential-context treatment. The primary paired analysis therefore contains the 1,980 confirmed adults with exact first-choice reproduction; all 1,991 confirmed adults form the robustness cohort. The original response records were preserved unchanged.

\begin{table*}[!b]
\caption{Single-checkpoint deterministic Qwen3-8B sequential-context boundary check. Behavioral intervals resample matched UUID histories.}
\label{tab:s_memory}
\centering
\small
\begin{tabular}{lrrrrrr}
\toprule
Cohort & UUIDs & Changed & Stateless agreement & Sequential agreement & Difference & 95\% CI \\
\midrule
Primary adult exact first-choice match & 1,980 & 0.1838 & 0.5293 & 0.4901 & $-0.0393$ & [$-0.0438$, $-0.0350$] \\
All-adult robustness & 1,991 & 0.1844 & 0.5293 & 0.4901 & $-0.0392$ & [$-0.0434$, $-0.0346$] \\
\bottomrule
\end{tabular}
\end{table*}

In the primary confirmed-adult cohort, sequential context changed 18.38\% of choices and reduced matched human agreement from 0.5293 to 0.4901 (Table~\ref{tab:s_memory}). Although raw DYNAMAP reconstruction rose from 0.853 to 0.978, its lift over the condition-specific majority rule fell from 0.057 to $-0.011$, indicating that greater internal predictability reflected response concentration rather than restored human-like structure.

\begin{figure*}[t]
\centering
\includegraphics[width=0.98\textwidth]{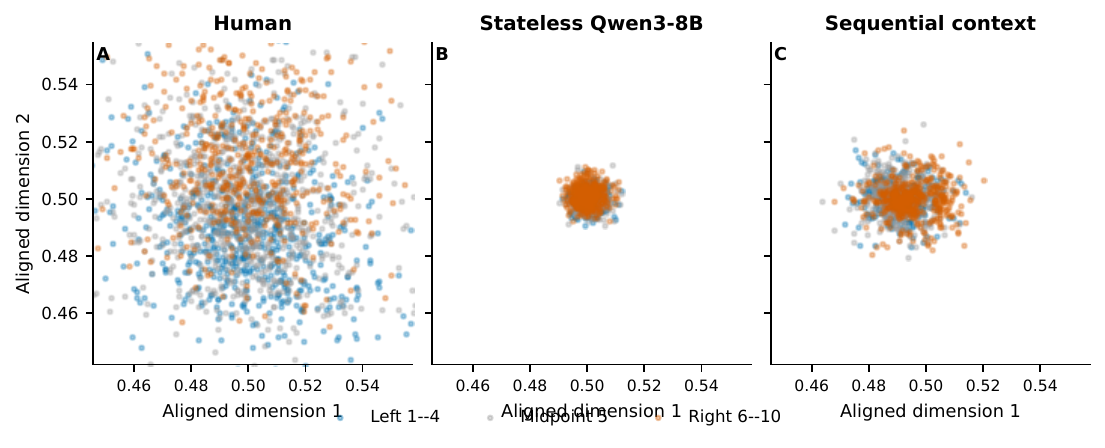}
\caption{Proposal-anchored geometry for the primary 1,980-UUID confirmed-adult sequential-context boundary check. The human, stateless, and sequential spaces use identical axes. Sequential context changes local concentration but does not restore the breadth of the human space.}
\label{fig:s_memory_geometry}
\figalttext[Three aligned preference maps]{Human coordinates form a broad cloud. Stateless and sequential-context Qwen3-8B coordinates remain narrowly concentrated near the center.}
\end{figure*}

Geometric recovery was likewise limited (Fig.~\ref{fig:s_memory_geometry}). In the primary paired cohort, left--right separation was 6.5\% of the matched human value under stateless replay and 18.2\% with sequential context; total coordinate variance was 1.6\% and 6.6\% of human variance, respectively; and proposal-distance correlation changed from 0.066 to 0.081. The all-adult fits again remained compressed despite some movement (18.0\% versus 32.2\% left--right separation; 0.9\% versus 3.4\% total variance). These fit-sensitive differences do not show restoration of human geometry. They apply only to the paired local replay and cannot be used to rank the sequential run against the primary grid.

\subsection{Confirmed-adult cohort construction and provenance}

All reported estimands use confirmed-adult cohorts. The Chilean source frame contained 2,000 sampled participants; the confirmed-adult cohort contained the 1,991 members whose exact reported age was at least 18. France retained 843 of 873 C20 participants and Brazil 434 of 452; because those releases record age categorically, the entire 10--19 band was excluded conservatively rather than attempting to distinguish ages 18--19. The broader C10 sensitivity likewise retained 847 French and 515 Brazilian confirmed adults. Stored model responses were projected onto these fixed adult manifests, and all human references, synthetic DYNAMAP fits, classifiers, geometric alignments, and independent proposal-space estimators were recomputed on matched adult UUID sets. No language-model inference was repeated. Original frame sizes are reported only to document cohort construction, not as alternative results.

\subsection{Thirty-seed DYNAMAP initialization sensitivity}

We varied only the DYNAMAP initialization seed across 30 fixed values specified before refitting the confirmed-adult primary C20 human cohorts and the three focal temperature-zero, self-placement-present, original-order checkpoints. This produced 360 fits: four response sources by three countries by 30 seeds. Every model space was aligned to the human space fitted with the same seed. The resulting distributions measure optimization sensitivity; they do not quantify LLM-generation, participant-sampling, or proposal-sampling uncertainty.

\begin{figure*}[p]
\centering
\includegraphics[width=0.98\textwidth]{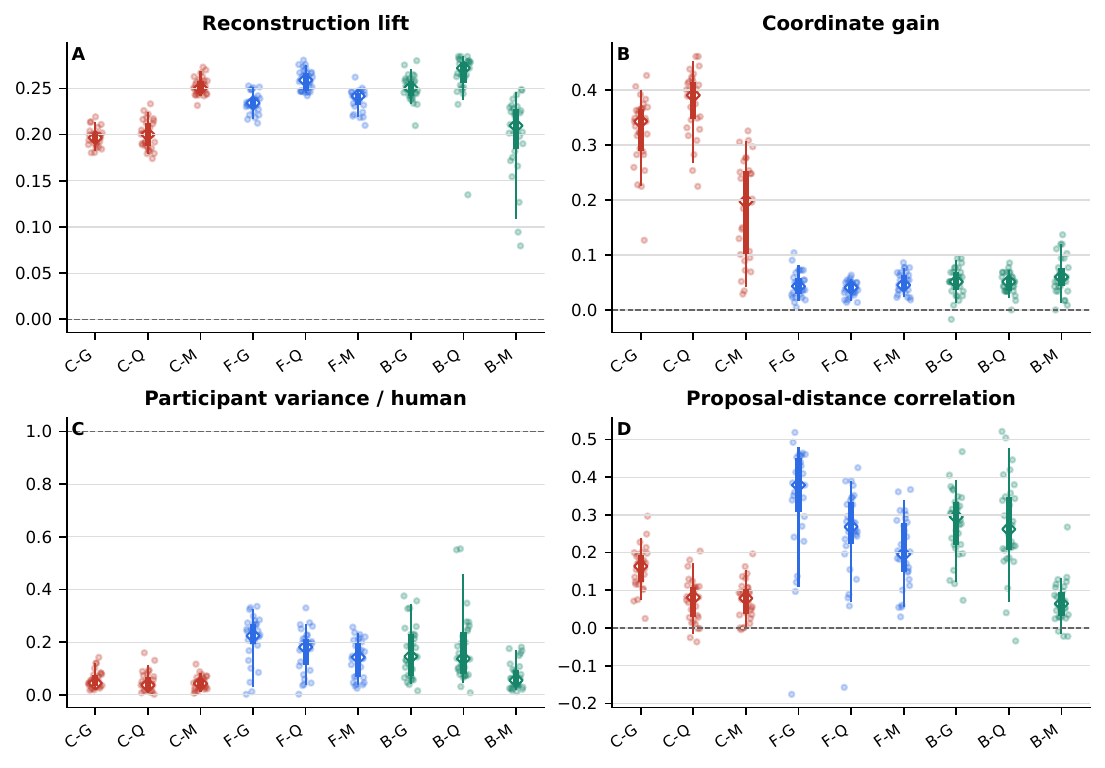}
\caption{Optimization sensitivity across 30 DYNAMAP initialization seeds for the three focal checkpoints and matched human references in each country. The distributions support robust positive internal reconstruction and participant-variance compression, while some one-dimensional and political-grouping metrics vary appreciably with initialization.}
\label{fig:s_multiseed}
\figalttext[Multi-seed DYNAMAP sensitivity panels]{Across countries and checkpoints, reconstruction lift stays positive and artificial total variance stays below the human reference for every seed. Coordinate gain is consistently positive in Chile and France but crosses zero in Brazil.}
\end{figure*}

Reconstruction lift was positive in all 270 artificial fits, and artificial participant-coordinate variance was below the same-seed adult human reference in all 270 (Table~\ref{tab:s_multiseed}; Fig.~\ref{fig:s_multiseed}). Coordinate gain over sociodemographics was positive in all 90 Chilean and all 90 French model fits and in 87 of 90 Brazilian fits. Precise model rankings were less stable than the central directional results. Occasional left--right ratios above one in Chile or France coexist with uniformly compressed total variance and do not imply recovery of the full two-dimensional topology.

\begin{table*}[p]
\caption{DYNAMAP initialization sensitivity across 30 seeds. Entries are seed means with 5th and 95th percentiles. Each artificial fit is normalized or aligned to the same-seed human fit. Lift is reconstruction accuracy minus the held-out majority rule; $\Delta$ coord. is held-out self-placement classification accuracy minus the sociodemographic baseline.}
\label{tab:s_multiseed}
\centering
\footnotesize
\textit{Panel A. Reconstruction lift and incremental self-placement recoverability}\par\smallskip
\input{source_tables/table_robustness_dynamap_multiseed_performance.tex}
\medskip

\textit{Panel B. Participant-variance and proposal-distance fidelity}\par\smallskip
\input{source_tables/table_robustness_dynamap_multiseed_geometry.tex}
\end{table*}

\subsection{Bradley--Terry aggregate proposal-space validation}

Regularized Bradley--Terry rankings used pairwise logistic models without an intercept \citep{Bradley1952}. The primary regularization was L2 $C=1$, with $C=0.1$ and $C=10$ point-estimate sensitivities. The analysis used the same confirmed-adult C20 cohorts and the same three temperature-zero, self-placement-present, original-order checkpoints as the multi-seed analysis. Within each country and checkpoint, 2,000 paired UUID-cluster bootstrap replicates refit both the human and synthetic rankings. Intervals therefore represent participant-cohort sampling uncertainty conditional on the observed proposal universe; they are not model-generation or country-level uncertainty.

\begin{figure*}[p]
\centering
\includegraphics[width=0.98\textwidth]{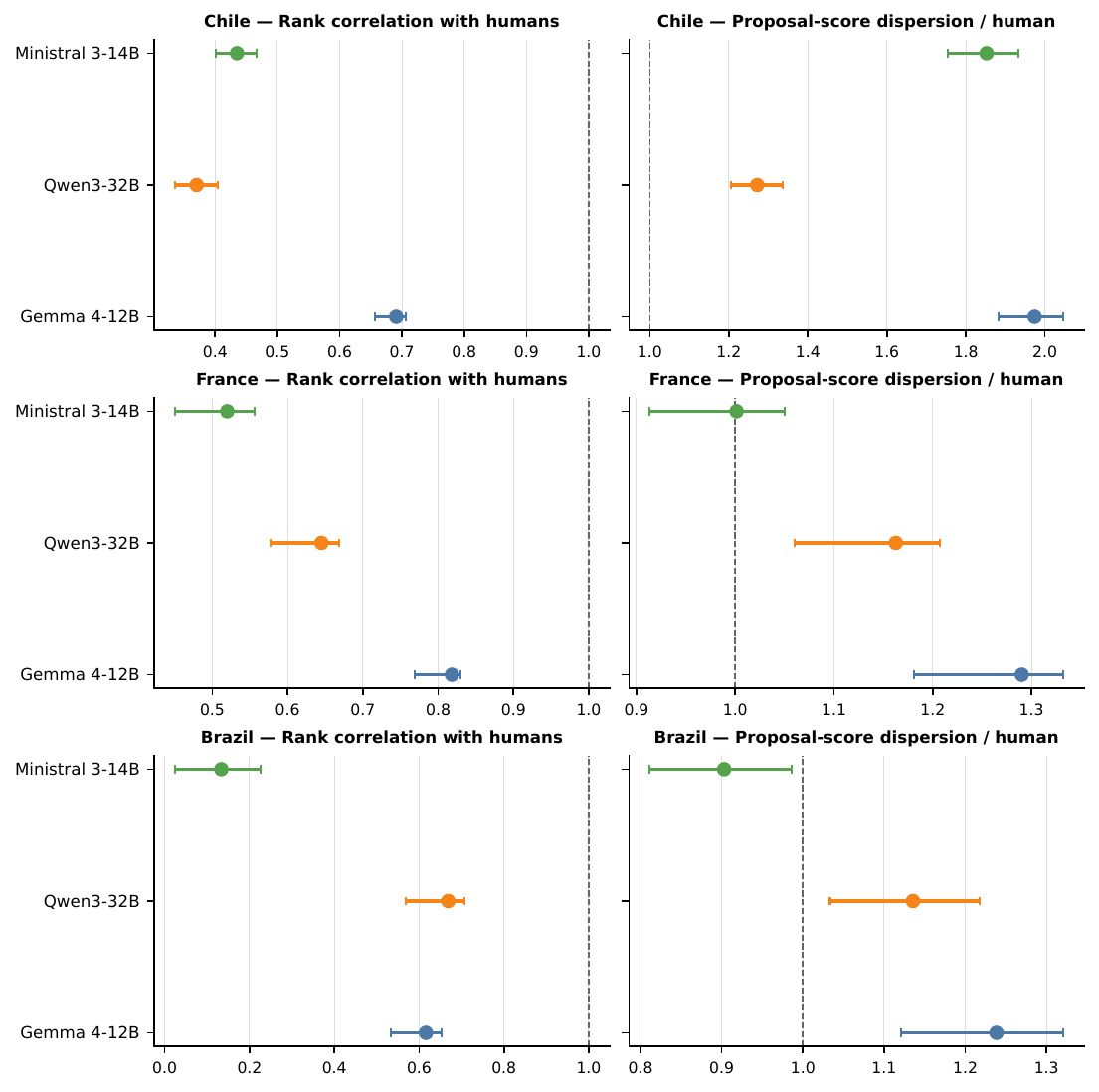}
\caption{Bradley--Terry proposal-space agreement and dispersion in the three focal temperature-zero conditions. Intervals are paired UUID-cluster bootstrap 95\% intervals. A dispersion ratio above one denotes steeper aggregate proposal utilities than in the matched human cohort, not greater participant diversity.}
\label{fig:s_bt_primary}
\figalttext[Bradley-Terry agreement and dispersion panels]{Adult human proposal-rank agreement is strongest for Gemma in Chile and France and for Qwen in Brazil, while Brazil Ministral is weak. Eight of nine proposal-score dispersion point ratios exceed the human reference of one.}
\end{figure*}

Spearman adult human-rank correlation ranged from 0.370 to 0.691 in Chile, 0.520 to 0.818 in France, and 0.133 to 0.669 in Brazil (Table~\ref{tab:s_bt_primary}; Fig.~\ref{fig:s_bt_primary}). Gemma ranked proposals most similarly to humans in Chile and France, Qwen did so in Brazil, and the Brazilian Ministral association was weak. Pairwise ordering agreement ranged from 0.626 to 0.747, 0.679 to 0.808, and 0.543 to 0.735, respectively. Aggregate proposal-score dispersion was not uniformly compressed: eight of nine artificial-to-human point ratios exceeded one, while Brazil Ministral was below one; the France Ministral point estimate was approximately one. This does not contradict participant-level DYNAMAP compression. Bradley--Terry dispersion measures between-proposal marginal-utility gradients; DYNAMAP variance measures between-person latent heterogeneity after proposal-anchor alignment. Aggregate rank recovery likewise does not establish person-level fidelity.

\begin{table*}[p]
\caption{Bradley--Terry aggregate proposal-space validation. Estimates use L2 $C=1$ and 2,000 paired UUID-cluster bootstrap refits of the human and artificial rankings. Dispersion is the ratio of synthetic to human root-mean-square proposal scores.}
\label{tab:s_bt_primary}
\centering
\footnotesize
\textit{Panel A. Human rank and pairwise-order recovery}\par\smallskip
\input{source_tables/table_robustness_bradley_terry_ordering.tex}
\medskip

\textit{Panel B. Aggregate proposal-score dispersion}\par\smallskip
\input{source_tables/table_robustness_bradley_terry_dispersion.tex}
\end{table*}

\subsection{Political cleavage without DYNAMAP embeddings}

To test left--right structure without using DYNAMAP coordinates, we fit separate Bradley--Terry utilities among left- and right-identified participants. Chile uses self-placement 1--4 versus 6--10; France and Brazil use 1--2 versus 4--5. The midpoint is excluded only from this contrast. Utilities are centered within group, and proposal-level cleavage is right-group minus left-group utility. Positive values therefore indicate relatively greater right-group preference. The estimator, regularization sensitivities, and paired UUID-cluster bootstrap were specified before estimation. Each bootstrap resamples left and right UUIDs separately and applies the same multiplicities to matched human and model responses; intervals condition on the observed proposal universe and fixed model outputs.

\begin{figure*}[p]
\centering
\includegraphics[width=0.98\textwidth]{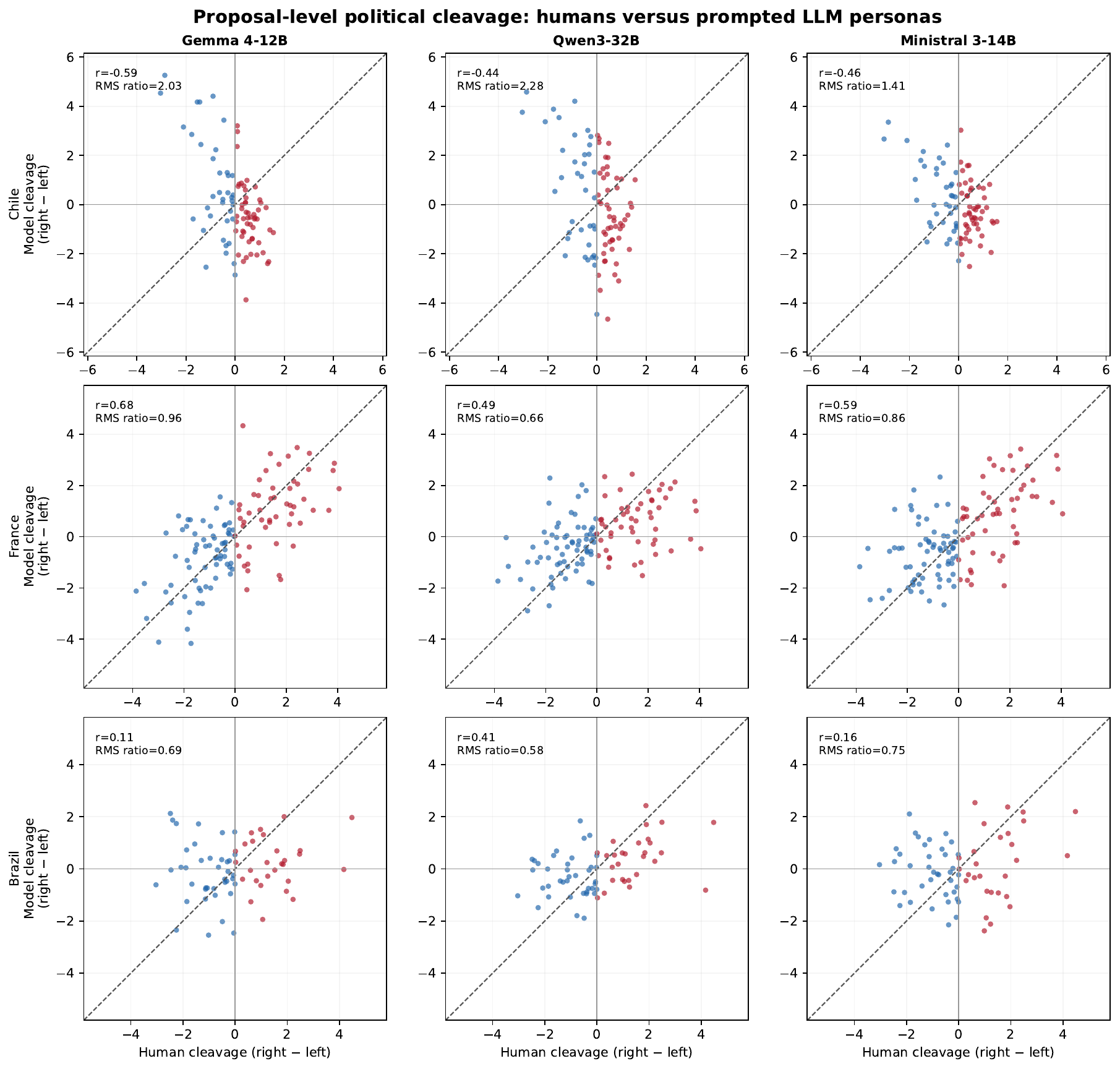}
\caption{Proposal-level human and synthetic political cleavage under the bare-numeric self-placement prompt, without DYNAMAP embeddings. Each panel compares centered right-minus-left Bradley--Terry utilities for one country and checkpoint. The diagonal denotes equality; point color denotes the sign of human cleavage. These baseline patterns motivated the later single-checkpoint endpoint-anchor probe in Figs.~\ref{fig:s_endpoint_behavior}--\ref{fig:s_endpoint_dynamap}.}
\label{fig:s_bt_cleavage}
\figalttext[Nine proposal-level cleavage scatterplots]{France shows positive human-model cleavage alignment, Brazil mostly weak positive alignment and attenuated magnitude, and Chile negative alignment with synthetic magnitude exceeding the human reference.}
\end{figure*}

Under the bare-numeric self-placement prompt, the embedding-independent comparison does not support universal attenuation of political cleavage (Table~\ref{tab:s_bt_cleavage}; Fig.~\ref{fig:s_bt_cleavage}). In France, adult human--model cleavage correlations were positive for all checkpoints ($r=0.494$--$0.676$), while model RMS magnitude was 0.66--0.96 of human. In Brazil, magnitude was also attenuated (0.58--0.75), but correspondence was weak for Gemma and Ministral, with 95\% intervals that included zero; Qwen had the strongest positive correlation ($r=0.410$). In Chile, all three cleavage vectors were negatively associated with the adult human vector ($r=-0.592$ to $-0.442$, intervals excluding zero) and their RMS magnitude was 1.41--2.28 times the human value. Under the instrument-defined Chilean grouping of 1--4 as left and 6--10 as right, this is a directional mismatch between synthetic and human proposal-level cleavage. The $C=0.1$ and $C=10$ sensitivities preserve the qualitative diagnosis. The endpoint-anchor section below shows that the Qwen3-32B component of this pattern is prompt-sensitive and should not be read as an invariant country signature.

This result is compatible with uniformly narrow participant-level DYNAMAP variance because the estimands differ. DYNAMAP variance measures the breadth of individual latent positions after proposal-anchor alignment; Bradley--Terry cleavage measures a between-group difference in marginal proposal utilities. A narrow synthetic population can still encode a steep, rotated, or reversed group contrast. The result describes persona-conditioned differentiation under the tested self-placement prompt; it is neither emergent inference of political orientation nor evidence that participant-level compression implies uniformly attenuated cleavage.

\begin{table*}[p]
\caption{Embedding-independent Bradley--Terry political-cleavage validation. Pearson correlation and sign agreement assess whether the same proposals divide human and synthetic left/right groups in the same direction. RMS ratio and slope assess magnitude; a ratio of one and slope of one denote equality with the matched adult human cleavage vector. Brackets are equal-tailed percentile intervals from paired UUID-cluster bootstraps. For discrete or skewed statistics, a full-sample estimate need not lie at the center of the percentile interval; the France--Ministral sign-agreement estimate is 0.750 [0.633, 0.758].}
\label{tab:s_bt_cleavage}
\centering
\scriptsize
\resizebox{\textwidth}{!}{\input{source_tables/table_robustness_bradley_terry_cleavage.tex}}
\end{table*}

\clearpage
\section{Verbal endpoint-anchor mechanism probe}

\begin{figure*}[p]
\centering
\includegraphics[width=0.98\textwidth]{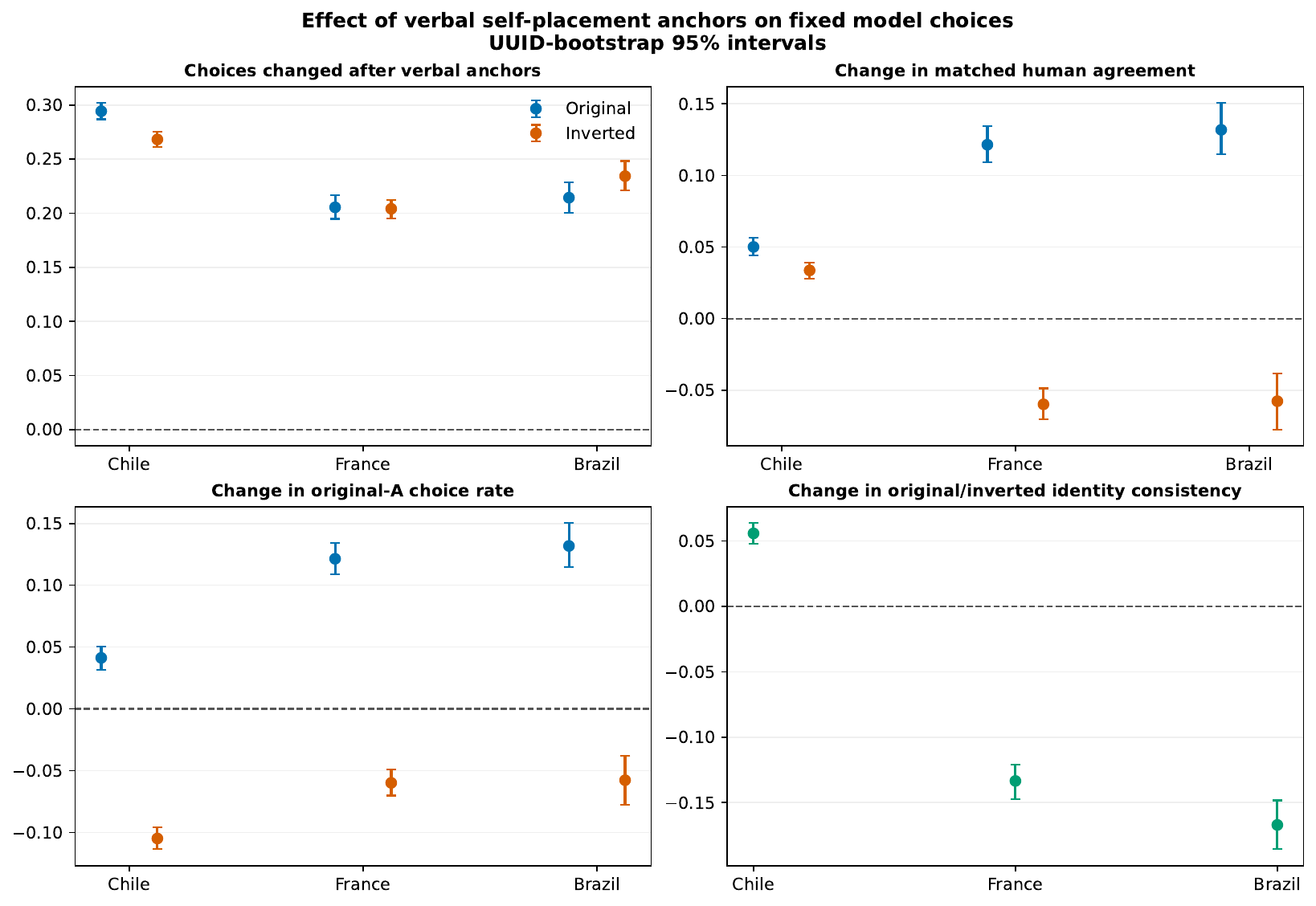}
\caption{Paired behavioral effects of verbal self-placement endpoints. The upper panels report the share of choices changed and the change in matched human agreement separately by proposal order. The lower panels report changes in original-A rate and cross-order proposal-identity consistency. Intervals are paired UUID-cluster bootstrap 95\% intervals.}
\label{fig:s_endpoint_behavior}
\figalttext[Four panels of endpoint-anchor behavioral effects]{Verbal endpoints change about one fifth to three tenths of choices. Human agreement rises in original order in all settings but falls after inversion in France and Brazil. Cross-order consistency improves only in Chile.}
\end{figure*}

The verbal endpoint intervention changed a substantial share of fixed Qwen3-32B choices while producing only modest order-averaged gains in matched human agreement (Table~\ref{tab:s_endpoint_anchor}; Fig.~\ref{fig:s_endpoint_behavior}). Anchors changed 28.1\% [27.4\%, 28.8\%] of choices in Chile, 20.5\% [19.7\%, 21.2\%] in France, and 22.4\% [21.3\%, 23.6\%] in Brazil. Agreement increased by 4.2 [3.7, 4.7], 3.1 [2.2, 4.0], and 3.7 [2.4, 5.0] percentage points, respectively, after equal weighting across original and inverted displays. Cross-order proposal-identity consistency increased from 0.689 to 0.745 in Chile but decreased from 0.759 to 0.625 in France and from 0.737 to 0.569 in Brazil. Thus, verbal clarification did not remove the position intervention and intensified it in two settings.

\begin{table*}[p]
\caption{Endpoint-anchor diagnostic for Qwen3-32B at $T=0$. Changed and $\Delta$ agreement compare anchored with bare responses within the same participant--decision cells. Cleavage is Pearson correlation between adult human and model Bradley--Terry right-minus-left proposal utilities. Self-pl. accuracy and participant variance use the primary DYNAMAP initialization (seed 101); variance is normalized to the same-seed national human fit. Original and inverted displays are reported separately because the anchor effect remained position-dependent.}
\label{tab:s_endpoint_anchor}
\centering
\scriptsize
\resizebox{\textwidth}{!}{\input{source_tables/table_endpoint_anchor_diagnostic.tex}}
\end{table*}

\begin{figure*}[p]
\centering
\includegraphics[width=0.98\textwidth]{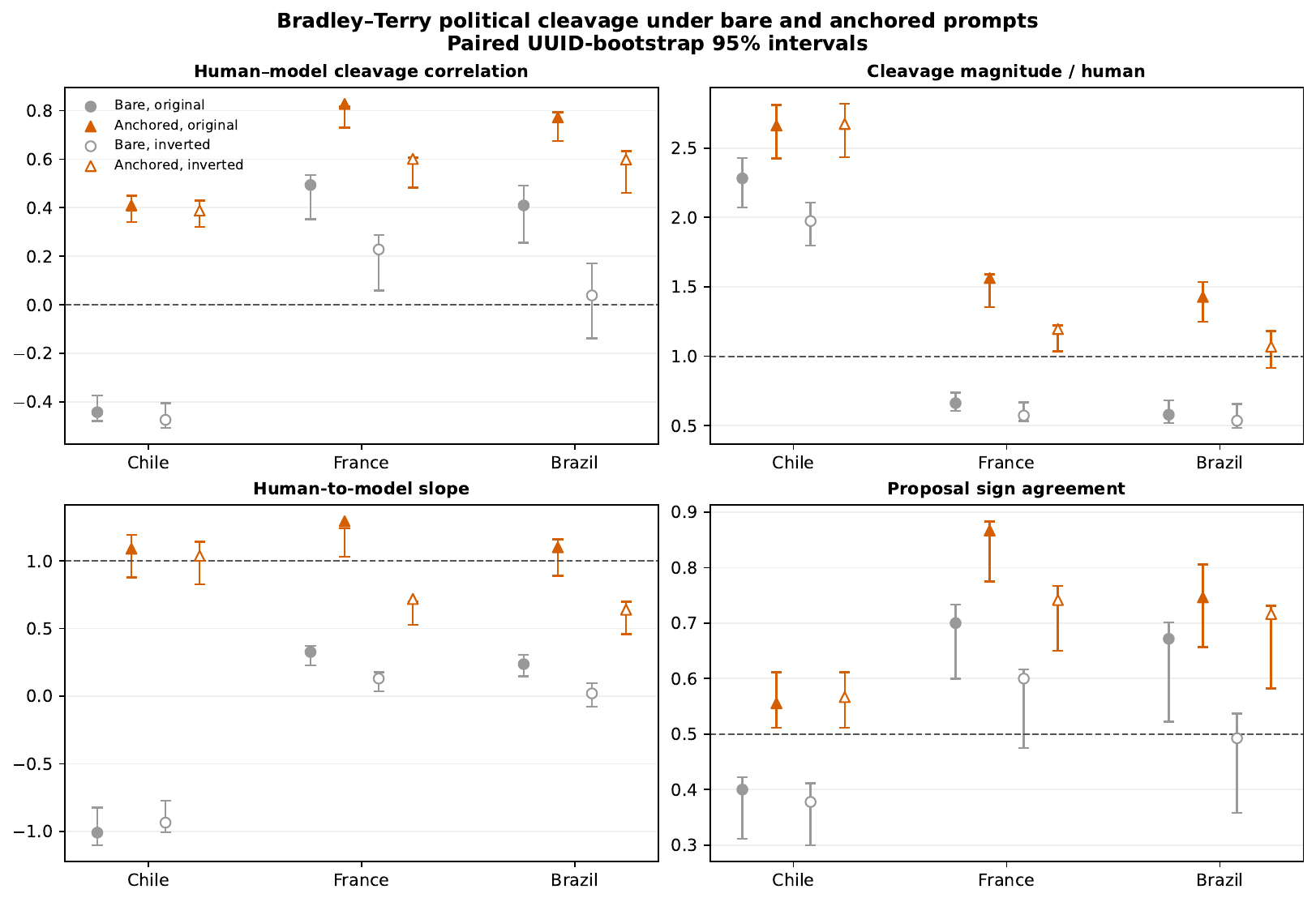}
\caption{Bradley--Terry political cleavage under bare numeric and verbally anchored self-placement. Panels compare human--model correlation, synthetic-to-human RMS magnitude, slope through the origin, and proposal sign agreement. Filled markers denote original order and hollow markers inverted order; intervals are paired UUID-cluster bootstrap 95\% intervals.}
\label{fig:s_endpoint_cleavage}
\figalttext[Four panels of endpoint-anchor cleavage metrics]{Anchored prompts move political-cleavage correlations and slopes toward the human direction in every country and order, while magnitude becomes amplified in Chile and in some French and Brazilian cells.}
\end{figure*}

The embedding-independent cleavage estimates changed more sharply than average agreement (Fig.~\ref{fig:s_endpoint_cleavage}). Human--model cleavage correlation moved from negative to positive in Chile under both orders ($-0.442$ to 0.409 and $-0.474$ to 0.388). France moved from 0.494 to 0.828 in original order and from 0.228 to 0.601 after inversion; Brazil moved from 0.410 to 0.772 and from 0.039 to 0.599. The direction became more human-aligned, but magnitude was not restored to equality: the RMS ratio increased from 2.28 to 2.66 and from 1.97 to 2.67 in Chile, from 0.66 to 1.56 and 0.57 to 1.20 in France, and from 0.58 to 1.43 and 0.54 to 1.07 in Brazil. The bare-prompt country pattern therefore cannot be interpreted as an invariant model or national signature.

\begin{figure*}[p]
\centering
\includegraphics[width=0.98\textwidth]{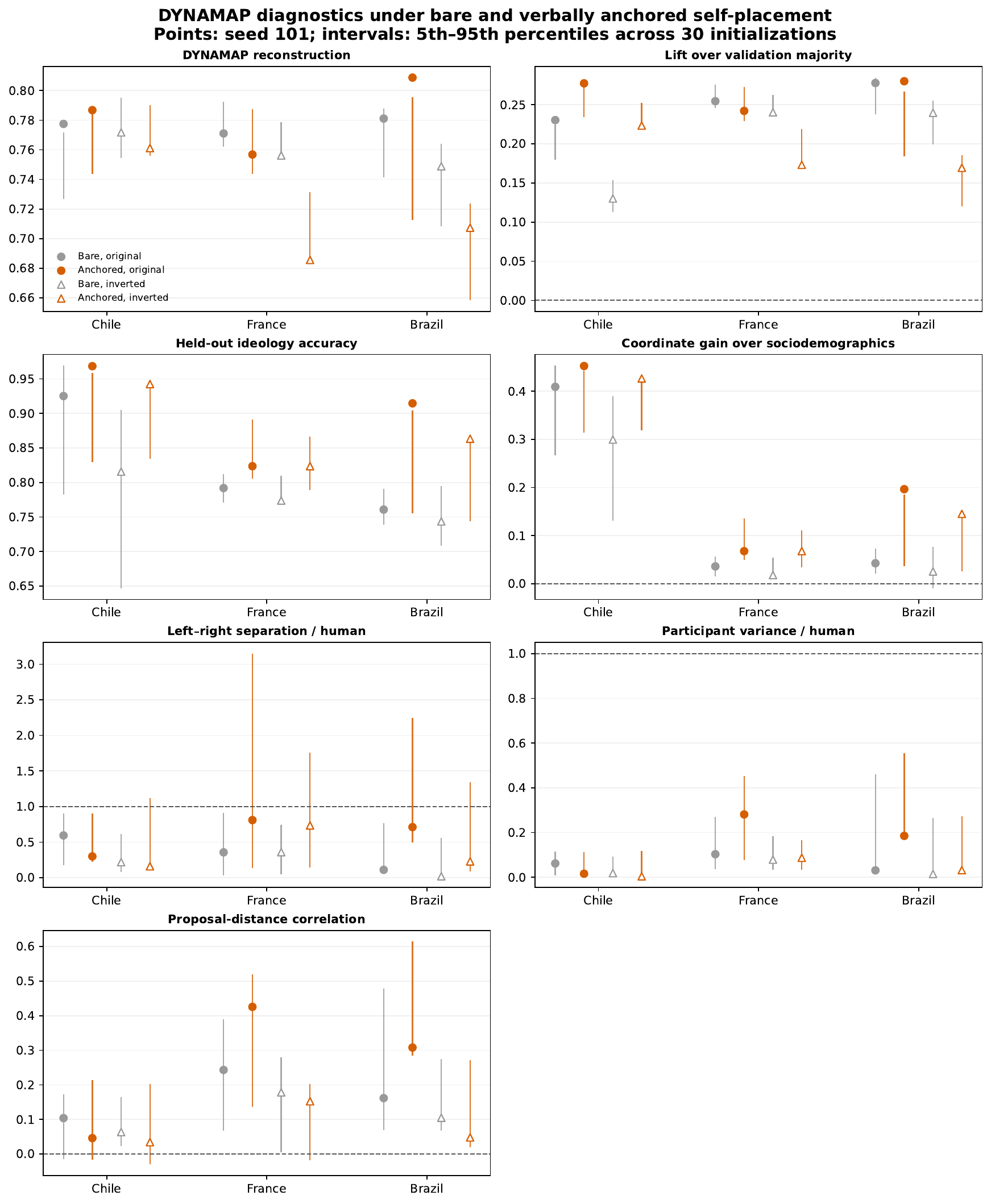}
\caption{DYNAMAP diagnostics under bare numeric and verbally anchored self-placement. Points show seed 101; vertical ranges show the 5th--95th percentiles across 30 additional initialization seeds. Bare and anchored original/inverted conditions use identical adult cohorts, splits, architecture, and same-seed human references. The dashed value one in normalized panels denotes the human reference.}
\label{fig:s_endpoint_dynamap}
\figalttext[Seven DYNAMAP panels for endpoint-anchor sensitivity]{Verbal endpoints usually increase self-placement recoverability and sometimes expand left-right separation or variance, but all seed-101 participant-variance estimates remain far below the human reference and several effects differ by proposal order or initialization.}
\end{figure*}

The DYNAMAP refits reached the same substantive boundary (Fig.~\ref{fig:s_endpoint_dynamap}). At seed 101, anchors increased held-out self-placement accuracy in every country and order. In original order it rose from 0.925 to 0.968 in Chile, 0.792 to 0.824 in France, and 0.761 to 0.915 in Brazil; inverted-order gains computed from the displayed three-decimal values were 0.126, 0.050, and 0.119. Participant-coordinate variance nevertheless remained below the same-seed human reference in all six anchored fits: 0.016 and 0.004 in Chile, 0.281 and 0.087 in France, and 0.185 and 0.032 in Brazil for original and inverted displays. In Chile, the anchored values were reductions from bare-prompt fractions of 0.062 and 0.019: greater recoverability coincided with a narrower participant space. France and Brazil expanded under original order, but the smaller inverted-order estimates and broad initialization distributions show that this was not a position-invariant recovery of population geometry. Left--right separation increased robustly in Brazil, was frequently positive but initialization-sensitive in France, and showed no stable repair in Chile. The intervention therefore improved interpretation of the supplied axis without recovering the full two-dimensional human structure.

\clearpage
\section{Reproducibility materials}

The accompanying reproducibility package separates stored model responses from statistical estimation and manuscript rendering. It provides the exact confirmed-adult cohort manifests; model-condition dictionary; bare and endpoint-anchored prompts and response-decoding rules; complete condition-level metrics; participant-cluster bootstrap summaries; DYNAMAP initialization sensitivities; Bradley--Terry aggregate-ranking and political-cleavage estimates; machine-readable source data for every displayed table and figure; and executable analysis notebooks. Exact model revisions, inference engines, numerical formats, decoding settings, and random seeds are retained. Public materials exclude direct identifiers and private participant-level joined records.

\section{Scope of robustness analyses}

The completed robustness analyses address A/B position, the political self-placement cue, verbal endpoint specification at one focal checkpoint, participant and proposal composition, selected policy domains, adult cohort construction, DYNAMAP initialization for the focal checkpoints, and participant-cluster uncertainty for Bradley--Terry estimates. They do not provide probability-sample or calibrated population estimates, proposal-superpopulation inference, categorical-region split sensitivity, repeated DYNAMAP fitting for every grid and treatment-balanced condition, or generalization of the endpoint effect across checkpoints. These boundaries limit the scope of inference; they are not unreported completed analyses.

\bibliographystyle{oup-plain}
\nocite{Abeliuk2026,Argyle2023,Atari2023,Ball2025,Bernardelle2025,Bhattacharyya2026,Bisbee2024,Brinkmann2023,GudinoRosero2024,Lu2025,Navarrete2024,Santurkar2023,Wang2025,Xiao2026}
\bibliography{references}

%% file: source_tables/table_all_conditions.tex
\begin{tabular}{llrrrrrrrr}
\toprule
Family & Size & $T$ & Dyn. & Lift & Human & Self-pl. & L--R (\%) & Var. (\%) & $r_{dist}$ \\
\midrule
Qwen3 & 8B & 0.0 & 0.804 & 0.123 & 0.544 & 0.663 & 11.7 & 1.1 & 0.061 \\
Qwen3 & 8B & 0.2 & 0.789 & 0.127 & 0.550 & 0.605 & 11.5 & 1.8 & 0.081 \\
Qwen3 & 8B & 0.7 & 0.821 & 0.089 & 0.543 & 0.764 & 10.1 & 0.8 & 0.028 \\
Qwen3 & 14B & 0.0 & 0.810 & 0.205 & 0.546 & 0.741 & 9.2 & 0.4 & 0.057 \\
Qwen3 & 14B & 0.2 & 0.839 & 0.052 & 0.536 & 0.804 & 19.6 & 1.2 & 0.069 \\
Qwen3 & 14B & 0.7 & 0.850 & 0.040 & 0.534 & 0.744 & 16.5 & 1.0 & 0.066 \\
Qwen3 & 32B & 0.0 & 0.777 & 0.230 & 0.523 & 0.925 & 59.4 & 6.2 & 0.104 \\
Qwen3 & 32B & 0.2 & 0.776 & 0.104 & 0.526 & 0.876 & 39.4 & 4.3 & 0.104 \\
Qwen3 & 32B & 0.7 & 0.788 & 0.065 & 0.521 & 0.896 & 37.8 & 3.6 & 0.082 \\
Gemma 4 & E4B & 0.0 & 0.746 & 0.185 & 0.537 & 0.741 & 16.6 & 1.9 & 0.043 \\
Gemma 4 & E4B & 0.2 & 0.745 & 0.183 & 0.537 & 0.758 & 24.3 & 2.7 & 0.056 \\
Gemma 4 & E4B & 0.7 & 0.723 & 0.138 & 0.536 & 0.709 & 19.5 & 3.1 & 0.071 \\
Gemma 4 & 12B & 0.0 & 0.794 & 0.202 & 0.571 & 0.876 & 39.4 & 2.9 & 0.112 \\
Gemma 4 & 12B & 0.2 & 0.799 & 0.208 & 0.571 & 0.876 & 24.8 & 1.3 & 0.087 \\
Gemma 4 & 12B & 0.7 & 0.775 & 0.178 & 0.570 & 0.744 & 23.5 & 3.3 & 0.097 \\
Gemma 4 & 31B & 0.0 & 0.801 & 0.261 & 0.567 & 0.732 & 15.6 & 4.7 & 0.136 \\
Gemma 4 & 31B & 0.2 & 0.804 & 0.265 & 0.566 & 0.749 & 16.2 & 4.2 & 0.129 \\
Gemma 4 & 31B & 0.7 & 0.805 & 0.268 & 0.567 & 0.738 & 16.4 & 4.3 & 0.135 \\
Ministral 3 & 3B & 0.0 & 0.789 & 0.012 & 0.510 & 0.669 & 20.3 & 5.0 & 0.079 \\
Ministral 3 & 3B & 0.2 & 0.768 & 0.024 & 0.512 & 0.533 & 3.2 & 3.9 & 0.085 \\
Ministral 3 & 3B & 0.7 & 0.659 & 0.035 & 0.510 & 0.588 & 13.0 & 3.9 & 0.054 \\
Ministral 3 & 8B & 0.0 & 0.779 & 0.058 & 0.534 & 0.697 & 8.4 & 1.2 & 0.049 \\
Ministral 3 & 8B & 0.2 & 0.778 & 0.051 & 0.534 & 0.648 & 6.7 & 0.8 & 0.023 \\
Ministral 3 & 8B & 0.7 & 0.764 & 0.017 & 0.531 & 0.617 & 9.5 & 3.3 & 0.069 \\
Ministral 3 & 14B & 0.0 & 0.767 & 0.256 & 0.546 & 0.646 & 8.6 & 1.5 & 0.065 \\
Ministral 3 & 14B & 0.2 & 0.762 & 0.255 & 0.546 & 0.677 & 13.4 & 1.9 & 0.080 \\
Ministral 3 & 14B & 0.7 & 0.707 & 0.148 & 0.544 & 0.735 & 25.9 & 4.7 & 0.111 \\
\bottomrule
\end{tabular}

%% file: source_tables/table_confirmatory_cells.tex
\begin{tabular}{lllrrrrrrrr}
\toprule
Model & Cue & Order & Dyn. & Lift & Human & Self-pl. & CV-10 & L--R (\%) & Var. (\%) & $r_{dist}$ \\
\midrule
Gemma 4-12B & Present & Original & 0.794 & 0.202 & 0.571 & 0.876 & 0.881 & 39.4 & 2.9 & 0.112 \\
Gemma 4-12B & Present & Inverted & 0.784 & 0.201 & 0.568 & 0.853 & 0.855 & 38.3 & 3.7 & 0.142 \\
Gemma 4-12B & Ablated & Original & 0.813 & 0.249 & 0.598 & 0.533 & 0.543 & 0.7 & 1.7 & 0.116 \\
Gemma 4-12B & Ablated & Inverted & 0.827 & 0.267 & 0.594 & 0.559 & 0.542 & 0.7 & 2.1 & 0.098 \\
Qwen3-32B & Present & Original & 0.777 & 0.230 & 0.523 & 0.925 & 0.912 & 59.4 & 6.2 & 0.104 \\
Qwen3-32B & Present & Inverted & 0.772 & 0.130 & 0.536 & 0.816 & 0.815 & 21.8 & 1.9 & 0.063 \\
Qwen3-32B & Ablated & Original & 0.803 & 0.161 & 0.560 & 0.539 & 0.536 & 0.8 & 1.9 & 0.084 \\
Qwen3-32B & Ablated & Inverted & 0.815 & 0.157 & 0.567 & 0.556 & 0.564 & 1.5 & 0.9 & 0.043 \\
Ministral 3-14B & Present & Original & 0.767 & 0.256 & 0.546 & 0.646 & 0.641 & 8.6 & 1.5 & 0.065 \\
Ministral 3-14B & Present & Inverted & 0.763 & 0.258 & 0.547 & 0.692 & 0.687 & 15.4 & 1.7 & 0.062 \\
Ministral 3-14B & Ablated & Original & 0.827 & 0.246 & 0.573 & 0.522 & 0.519 & 5.2 & 4.0 & 0.094 \\
Ministral 3-14B & Ablated & Inverted & 0.818 & 0.238 & 0.576 & 0.545 & 0.564 & 2.6 & 3.5 & 0.087 \\
\bottomrule
\end{tabular}

%% file: source_tables/table_confirmatory_position.tex
\begin{tabular}{llrr}
\toprule
Model & Self-placement cue & Cross-order agreement [95\% CI] & Choice changed (\%) \\
\midrule
Gemma 4-12B & Present & 0.779 [0.774, 0.784] & 22.1 \\
Gemma 4-12B & Ablated & 0.834 [0.831, 0.838] & 16.6 \\
Qwen3-32B & Present & 0.689 [0.684, 0.694] & 31.1 \\
Qwen3-32B & Ablated & 0.688 [0.683, 0.693] & 31.2 \\
Ministral 3-14B & Present & 0.786 [0.782, 0.791] & 21.4 \\
Ministral 3-14B & Ablated & 0.780 [0.776, 0.785] & 22.0 \\
\bottomrule
\end{tabular}

%% file: source_tables/table_crossed_position_stability.tex
\begin{tabular}{llccc}
\toprule
Model & Self-placement cue & Agreement & UUID-only CI & UUID $\times$ proposal CI \\
\midrule
Gemma 4-12B & present & 0.779 & [0.774, 0.784] & [0.741, 0.806] \\
Gemma 4-12B & ablated & 0.834 & [0.831, 0.838] & [0.795, 0.868] \\
Qwen3-32B & present & 0.689 & [0.684, 0.694] & [0.656, 0.720] \\
Qwen3-32B & ablated & 0.688 & [0.683, 0.693] & [0.628, 0.747] \\
Ministral 3-14B & present & 0.786 & [0.782, 0.791] & [0.763, 0.808] \\
Ministral 3-14B & ablated & 0.780 & [0.776, 0.785] & [0.744, 0.811] \\
\bottomrule
\end{tabular}

%% file: source_tables/table_confirmatory_prompt.tex
\begin{tabular}{llrrrr}
\toprule
Model & Order & Held-out $\Delta$ & CV-10 $\Delta$ & Human agreement $\Delta$ & 95\% CI \\
\midrule
Gemma 4-12B & Original & 0.343 & 0.337 & -0.028 & [-0.033, -0.023] \\
Gemma 4-12B & Inverted & 0.294 & 0.314 & -0.026 & [-0.031, -0.020] \\
Qwen3-32B & Original & 0.386 & 0.375 & -0.037 & [-0.042, -0.031] \\
Qwen3-32B & Inverted & 0.259 & 0.251 & -0.031 & [-0.036, -0.027] \\
Ministral 3-14B & Original & 0.124 & 0.122 & -0.028 & [-0.033, -0.023] \\
Ministral 3-14B & Inverted & 0.147 & 0.123 & -0.029 & [-0.034, -0.024] \\
\bottomrule
\end{tabular}

%% file: source_tables/table_crossed_ideology_effects.tex
\begin{tabular}{llccc}
\toprule
Model & Order & Present $-$ ablated & UUID-only CI & UUID $\times$ proposal CI \\
\midrule
Gemma 4-12B & original & -0.028 & [-0.033, -0.023] & [-0.060, -0.004] \\
Gemma 4-12B & inverted & -0.026 & [-0.031, -0.020] & [-0.057, -0.001] \\
Qwen3-32B & original & -0.037 & [-0.042, -0.031] & [-0.063, -0.013] \\
Qwen3-32B & inverted & -0.031 & [-0.036, -0.027] & [-0.049, -0.014] \\
Ministral 3-14B & original & -0.028 & [-0.033, -0.023] & [-0.055, -0.006] \\
Ministral 3-14B & inverted & -0.029 & [-0.034, -0.024] & [-0.054, -0.010] \\
\bottomrule
\end{tabular}

%% file: source_tables/table_human_benchmark_ladder.tex
\begin{tabular}{p{0.52\linewidth}cc}
\toprule
Predictor & Held-out accuracy & UUID bootstrap 95\% CI \\
\midrule
Balanced random choice & 0.500 & --- \\
Marginal A/B rule & 0.509 & [0.494, 0.524] \\
Bradley--Terry proposal popularity & 0.679 & [0.665, 0.693] \\
Sociodemographic + proposal identity & 0.687 & [0.672, 0.701] \\
Human DYNAMAP out of sample & 0.662 & [0.647, 0.677] \\
\bottomrule
\end{tabular}

%% file: source_tables/table_topic_coverage_compact.tex
\begin{tabular}{lrrrrrc}
\toprule
Topic & Decisions & UUIDs & Proposal A & Proposal B & Distinct proposals & Meets criteria \\
\midrule
Democracia & 124 & 121 & 6 & 6 & 6 & No \\
Economía & 280 & 267 & 8 & 8 & 8 & Yes \\
Educación & 293 & 275 & 8 & 8 & 8 & Yes \\
Laboral & 3 & 3 & 2 & 2 & 2 & No \\
Medio Ambiente & 116 & 113 & 5 & 5 & 5 & No \\
Modernización del Estado & 1300 & 970 & 17 & 17 & 17 & Yes \\
Pensiones & 107 & 106 & 5 & 5 & 5 & No \\
Política Criminal & 166 & 159 & 6 & 6 & 6 & No \\
Política Social y Ciudadanía & 107 & 104 & 5 & 5 & 5 & No \\
Política Tributaria & 147 & 140 & 6 & 6 & 6 & No \\
Reforma Congreso & 7 & 7 & 2 & 2 & 2 & No \\
Salud & 450 & 403 & 10 & 10 & 10 & Yes \\
Seguridad & 6 & 6 & 2 & 2 & 2 & No \\
Vivienda y Transporte & 261 & 244 & 8 & 8 & 8 & Yes \\
\bottomrule
\end{tabular}

%% file: source_tables/table_crosscountry_c20_all_cells.tex
\begin{tabular}{llllrrrrrr}
\toprule
Country & Checkpoint & Cue & Order & Dyn. acc. & Majority & Lift & Matched human & Displayed A & Displayed-A, order mean \\
\midrule
France & Gemma 4-12B & Present & Original & 0.775 & 0.558 & 0.217 & 0.778 & 0.778 & 0.595 \\
France & Gemma 4-12B & Present & Inverted & 0.751 & 0.551 & 0.200 & 0.587 & 0.413 & 0.595 \\
France & Gemma 4-12B & Ablated & Original & 0.816 & 0.561 & 0.254 & 0.733 & 0.733 & 0.542 \\
France & Gemma 4-12B & Ablated & Inverted & 0.807 & 0.546 & 0.262 & 0.650 & 0.350 & 0.542 \\
France & Qwen3-32B & Present & Original & 0.771 & 0.517 & 0.254 & 0.702 & 0.702 & 0.567 \\
France & Qwen3-32B & Present & Inverted & 0.756 & 0.516 & 0.240 & 0.569 & 0.431 & 0.567 \\
France & Qwen3-32B & Ablated & Original & 0.817 & 0.531 & 0.286 & 0.735 & 0.735 & 0.562 \\
France & Qwen3-32B & Ablated & Inverted & 0.828 & 0.533 & 0.295 & 0.610 & 0.390 & 0.562 \\
France & Ministral 3-14B & Present & Original & 0.730 & 0.501 & 0.230 & 0.634 & 0.634 & 0.482 \\
France & Ministral 3-14B & Present & Inverted & 0.769 & 0.505 & 0.264 & 0.671 & 0.329 & 0.482 \\
France & Ministral 3-14B & Ablated & Original & 0.759 & 0.508 & 0.251 & 0.533 & 0.533 & 0.382 \\
France & Ministral 3-14B & Ablated & Inverted & 0.792 & 0.506 & 0.286 & 0.768 & 0.232 & 0.382 \\
\addlinespace
Brazil & Gemma 4-12B & Present & Original & 0.757 & 0.528 & 0.229 & 0.719 & 0.719 & 0.636 \\
Brazil & Gemma 4-12B & Present & Inverted & 0.741 & 0.518 & 0.222 & 0.447 & 0.553 & 0.636 \\
Brazil & Gemma 4-12B & Ablated & Original & 0.825 & 0.515 & 0.310 & 0.709 & 0.709 & 0.569 \\
Brazil & Gemma 4-12B & Ablated & Inverted & 0.842 & 0.512 & 0.331 & 0.572 & 0.428 & 0.569 \\
Brazil & Qwen3-32B & Present & Original & 0.781 & 0.503 & 0.278 & 0.706 & 0.706 & 0.583 \\
Brazil & Qwen3-32B & Present & Inverted & 0.749 & 0.509 & 0.240 & 0.539 & 0.461 & 0.583 \\
Brazil & Qwen3-32B & Ablated & Original & 0.810 & 0.518 & 0.291 & 0.733 & 0.733 & 0.583 \\
Brazil & Qwen3-32B & Ablated & Inverted & 0.768 & 0.502 & 0.266 & 0.567 & 0.433 & 0.583 \\
Brazil & Ministral 3-14B & Present & Original & 0.742 & 0.503 & 0.238 & 0.510 & 0.510 & 0.467 \\
Brazil & Ministral 3-14B & Present & Inverted & 0.768 & 0.518 & 0.250 & 0.575 & 0.425 & 0.467 \\
Brazil & Ministral 3-14B & Ablated & Original & 0.755 & 0.535 & 0.220 & 0.465 & 0.465 & 0.363 \\
Brazil & Ministral 3-14B & Ablated & Inverted & 0.778 & 0.501 & 0.276 & 0.739 & 0.261 & 0.363 \\
\bottomrule
\end{tabular}

%% file: source_tables/table_crosscountry_c20_classification.tex
\begin{tabular}{llllrrrr}
\toprule
Country & Checkpoint & Cue & Order & Accuracy & Bal. acc. & Macro-F1 & ROC AUC \\
\midrule
France & Gemma 4-12B & Present & Original & 0.796 & 0.501 & 0.464 & 0.602 \\
France & Gemma 4-12B & Present & Inverted & 0.774 & 0.487 & 0.455 & 0.519 \\
France & Gemma 4-12B & Ablated & Original & 0.792 & 0.498 & 0.462 & 0.483 \\
France & Gemma 4-12B & Ablated & Inverted & 0.778 & 0.490 & 0.457 & 0.490 \\
France & Qwen3-32B & Present & Original & 0.792 & 0.507 & 0.481 & 0.594 \\
France & Qwen3-32B & Present & Inverted & 0.774 & 0.532 & 0.531 & 0.638 \\
France & Qwen3-32B & Ablated & Original & 0.805 & 0.525 & 0.507 & 0.611 \\
France & Qwen3-32B & Ablated & Inverted & 0.810 & 0.518 & 0.490 & 0.470 \\
France & Ministral 3-14B & Present & Original & 0.792 & 0.525 & 0.515 & 0.600 \\
France & Ministral 3-14B & Present & Inverted & 0.810 & 0.582 & 0.595 & 0.636 \\
France & Ministral 3-14B & Ablated & Original & 0.778 & 0.480 & 0.438 & 0.547 \\
France & Ministral 3-14B & Ablated & Inverted & 0.778 & 0.480 & 0.438 & 0.524 \\
\addlinespace
Brazil & Gemma 4-12B & Present & Original & 0.769 & 0.504 & 0.469 & 0.540 \\
Brazil & Gemma 4-12B & Present & Inverted & 0.769 & 0.518 & 0.498 & 0.420 \\
Brazil & Gemma 4-12B & Ablated & Original & 0.744 & 0.487 & 0.457 & 0.601 \\
Brazil & Gemma 4-12B & Ablated & Inverted & 0.803 & 0.540 & 0.519 & 0.714 \\
Brazil & Qwen3-32B & Present & Original & 0.761 & 0.542 & 0.540 & 0.674 \\
Brazil & Qwen3-32B & Present & Inverted & 0.744 & 0.502 & 0.484 & 0.633 \\
Brazil & Qwen3-32B & Ablated & Original & 0.786 & 0.573 & 0.580 & 0.677 \\
Brazil & Qwen3-32B & Ablated & Inverted & 0.778 & 0.538 & 0.529 & 0.478 \\
Brazil & Ministral 3-14B & Present & Original & 0.846 & 0.742 & 0.756 & 0.807 \\
Brazil & Ministral 3-14B & Present & Inverted & 0.829 & 0.658 & 0.686 & 0.776 \\
Brazil & Ministral 3-14B & Ablated & Original & 0.769 & 0.533 & 0.524 & 0.602 \\
Brazil & Ministral 3-14B & Ablated & Inverted & 0.786 & 0.544 & 0.535 & 0.591 \\
\bottomrule
\end{tabular}

%% file: source_tables/table_crosscountry_c20_performance.tex
\begin{tabular}{llccccc}
\toprule
Country & Checkpoint & Dyn. & Lift & Matched human [95\% CI] & Self-pl. & $\Delta$ coord. \\
\midrule
France & Human reference & 0.773 & 0.262 & --- & 0.774 & 0.018 \\
France & Gemma 4-12B & 0.787 & 0.233 & 0.687 [0.675, 0.699] & 0.785 & 0.029 \\
France & Qwen3-32B & 0.793 & 0.269 & 0.654 [0.642, 0.666] & 0.795 & 0.040 \\
France & Ministral 3-14B & 0.763 & 0.258 & 0.651 [0.640, 0.663] & 0.790 & 0.034 \\
Brazil & Human reference & 0.778 & 0.229 & --- & 0.821 & 0.103 \\
Brazil & Gemma 4-12B & 0.791 & 0.273 & 0.612 [0.594, 0.630] & 0.771 & 0.053 \\
Brazil & Qwen3-32B & 0.777 & 0.269 & 0.636 [0.619, 0.653] & 0.767 & 0.049 \\
Brazil & Ministral 3-14B & 0.761 & 0.246 & 0.572 [0.557, 0.589] & 0.808 & 0.090 \\
\bottomrule
\end{tabular}

%% file: source_tables/table_crosscountry_c20_geometry.tex
\begin{tabular}{llcccc}
\toprule
Country & Checkpoint & L--R/human & Var./human & $r_{dist}$ & $\rho_{BT}$ \\
\midrule
France & Human reference & 1.000 & 1.000 & 1.000 & 1.000 \\
France & Gemma 4-12B & 0.182 & 0.105 & 0.311 & 0.742 \\
France & Qwen3-32B & 0.219 & 0.086 & 0.241 & 0.588 \\
France & Ministral 3-14B & 0.363 & 0.103 & 0.243 & 0.556 \\
Brazil & Human reference & 1.000 & 1.000 & 1.000 & 1.000 \\
Brazil & Gemma 4-12B & 0.040 & 0.040 & 0.178 & 0.465 \\
Brazil & Qwen3-32B & 0.059 & 0.025 & 0.143 & 0.549 \\
Brazil & Ministral 3-14B & 0.111 & 0.043 & 0.141 & 0.310 \\
\bottomrule
\end{tabular}

%% file: source_tables/table_crosscountry_ab_stability.tex
\begin{tabular}{lllcccrr}
\toprule
Country & Checkpoint & Self-placement cue & Consistency & UUID 95\% CI & Change & UUIDs & Rows \\
\midrule
France & Gemma 4-12B & present & 0.788 & [0.778, 0.796] & 0.212 & 843 & 16860 \\
France & Gemma 4-12B & ablated & 0.877 & [0.870, 0.883] & 0.123 & 843 & 16860 \\
France & Qwen3-32B & present & 0.759 & [0.749, 0.768] & 0.241 & 843 & 16860 \\
France & Qwen3-32B & ablated & 0.813 & [0.805, 0.821] & 0.187 & 843 & 16860 \\
France & Ministral 3-14B & present & 0.779 & [0.771, 0.787] & 0.221 & 843 & 16860 \\
France & Ministral 3-14B & ablated & 0.741 & [0.731, 0.750] & 0.259 & 843 & 16860 \\
Brazil & Gemma 4-12B & present & 0.716 & [0.699, 0.732] & 0.284 & 434 & 8680 \\
Brazil & Gemma 4-12B & ablated & 0.824 & [0.812, 0.835] & 0.176 & 434 & 8680 \\
Brazil & Qwen3-32B & present & 0.737 & [0.723, 0.750] & 0.263 & 434 & 8680 \\
Brazil & Qwen3-32B & ablated & 0.766 & [0.753, 0.779] & 0.234 & 434 & 8680 \\
Brazil & Ministral 3-14B & present & 0.761 & [0.749, 0.773] & 0.239 & 434 & 8680 \\
Brazil & Ministral 3-14B & ablated & 0.698 & [0.685, 0.712] & 0.302 & 434 & 8680 \\
\bottomrule
\end{tabular}

%% file: source_tables/table_crosscountry_c10_c20_performance.tex
\begin{tabular}{lllcccc}
\toprule
Country & Checkpoint & Cohort & Dyn. & Lift & Human & Self-pl. \\
\midrule
France & Gemma 4-12B & C10 & 0.769 & 0.225 & 0.697 & 0.816 \\
France & Gemma 4-12B & C20 & 0.787 & 0.233 & 0.687 & 0.785 \\
France & Qwen3-32B & C10 & 0.770 & 0.248 & 0.664 & 0.812 \\
France & Qwen3-32B & C20 & 0.793 & 0.269 & 0.654 & 0.795 \\
France & Ministral 3-14B & C10 & 0.750 & 0.236 & 0.656 & 0.804 \\
France & Ministral 3-14B & C20 & 0.763 & 0.258 & 0.651 & 0.790 \\
Brazil & Gemma 4-12B & C10 & 0.732 & 0.223 & 0.602 & 0.737 \\
Brazil & Gemma 4-12B & C20 & 0.791 & 0.273 & 0.612 & 0.771 \\
Brazil & Qwen3-32B & C10 & 0.735 & 0.218 & 0.631 & 0.743 \\
Brazil & Qwen3-32B & C20 & 0.777 & 0.269 & 0.636 & 0.767 \\
Brazil & Ministral 3-14B & C10 & 0.691 & 0.171 & 0.568 & 0.746 \\
Brazil & Ministral 3-14B & C20 & 0.761 & 0.246 & 0.572 & 0.808 \\
\bottomrule
\end{tabular}

%% file: source_tables/table_crosscountry_c10_c20_geometry.tex
\begin{tabular}{lllccc}
\toprule
Country & Checkpoint & Cohort & L--R/human & Var./human & $r_{dist}$ \\
\midrule
France & Gemma 4-12B & C10 & 0.929 & 0.239 & 0.354 \\
France & Gemma 4-12B & C20 & 0.182 & 0.105 & 0.311 \\
France & Qwen3-32B & C10 & 0.859 & 0.181 & 0.280 \\
France & Qwen3-32B & C20 & 0.219 & 0.086 & 0.241 \\
France & Ministral 3-14B & C10 & 0.938 & 0.212 & 0.299 \\
France & Ministral 3-14B & C20 & 0.363 & 0.103 & 0.243 \\
Brazil & Gemma 4-12B & C10 & 0.090 & 0.075 & 0.188 \\
Brazil & Gemma 4-12B & C20 & 0.040 & 0.040 & 0.178 \\
Brazil & Qwen3-32B & C10 & 0.102 & 0.069 & 0.148 \\
Brazil & Qwen3-32B & C20 & 0.059 & 0.025 & 0.143 \\
Brazil & Ministral 3-14B & C10 & 0.101 & 0.075 & 0.126 \\
Brazil & Ministral 3-14B & C20 & 0.111 & 0.043 & 0.141 \\
\bottomrule
\end{tabular}

%% file: source_tables/table_robustness_dynamap_multiseed_performance.tex
\begin{tabular}{llrr}
\toprule
Country & Checkpoint & Lift, mean [P5, P95] & $\Delta$ coord., mean [P5, P95] \\
\midrule
Chile & Gemma 4-12B & 0.197 [0.182, 0.214] & 0.326 [0.226, 0.400] \\
Chile & Qwen3-32B & 0.200 [0.179, 0.224] & 0.377 [0.267, 0.453] \\
Chile & Ministral 3-14B & 0.252 [0.242, 0.268] & 0.185 [0.042, 0.307] \\
France & Gemma 4-12B & 0.234 [0.217, 0.251] & 0.046 [0.016, 0.082] \\
France & Qwen3-32B & 0.259 [0.246, 0.276] & 0.039 [0.016, 0.057] \\
France & Ministral 3-14B & 0.238 [0.219, 0.249] & 0.049 [0.023, 0.077] \\
Brazil & Gemma 4-12B & 0.250 [0.233, 0.271] & 0.051 [0.012, 0.090] \\
Brazil & Qwen3-32B & 0.263 [0.238, 0.285] & 0.048 [0.021, 0.073] \\
Brazil & Ministral 3-14B & 0.200 [0.109, 0.246] & 0.062 [0.012, 0.120] \\
\bottomrule
\end{tabular}

%% file: source_tables/table_robustness_dynamap_multiseed_geometry.tex
\begin{tabular}{llrr}
\toprule
Country & Checkpoint & Var./human, mean [P5, P95] & $r_{dist}$, mean [P5, P95] \\
\midrule
Chile & Gemma 4-12B & 0.056 [0.021, 0.120] & 0.159 [0.073, 0.239] \\
Chile & Qwen3-32B & 0.048 [0.009, 0.114] & 0.073 [-0.015, 0.173] \\
Chile & Ministral 3-14B & 0.044 [0.012, 0.083] & 0.075 [0.000, 0.155] \\
France & Gemma 4-12B & 0.212 [0.028, 0.328] & 0.345 [0.108, 0.479] \\
France & Qwen3-32B & 0.169 [0.037, 0.271] & 0.252 [0.068, 0.390] \\
France & Ministral 3-14B & 0.137 [0.036, 0.233] & 0.201 [0.055, 0.339] \\
Brazil & Gemma 4-12B & 0.164 [0.043, 0.344] & 0.277 [0.121, 0.392] \\
Brazil & Qwen3-32B & 0.175 [0.047, 0.459] & 0.271 [0.070, 0.478] \\
Brazil & Ministral 3-14B & 0.071 [0.019, 0.172] & 0.069 [-0.016, 0.132] \\
\bottomrule
\end{tabular}

%% file: source_tables/table_robustness_bradley_terry_ordering.tex
\begin{tabular}{llrr}
\toprule
Country & Checkpoint & Spearman $\rho$ [95\% CI] & Pairwise agreement [95\% CI] \\
\midrule
Chile & Gemma 4-12B & 0.691 [0.656, 0.706] & 0.747 [0.732, 0.754] \\
Chile & Qwen3-32B & 0.370 [0.336, 0.404] & 0.626 [0.614, 0.639] \\
Chile & Ministral 3-14B & 0.435 [0.401, 0.466] & 0.651 [0.638, 0.661] \\
France & Gemma 4-12B & 0.818 [0.769, 0.829] & 0.808 [0.783, 0.815] \\
France & Qwen3-32B & 0.645 [0.578, 0.669] & 0.730 [0.705, 0.741] \\
France & Ministral 3-14B & 0.520 [0.451, 0.556] & 0.679 [0.654, 0.694] \\
Brazil & Gemma 4-12B & 0.616 [0.533, 0.654] & 0.728 [0.693, 0.744] \\
Brazil & Qwen3-32B & 0.669 [0.568, 0.706] & 0.735 [0.698, 0.755] \\
Brazil & Ministral 3-14B & 0.133 [0.025, 0.225] & 0.543 [0.505, 0.575] \\
\bottomrule
\end{tabular}

%% file: source_tables/table_robustness_bradley_terry_dispersion.tex
\begin{tabular}{llr}
\toprule
Country & Checkpoint & RMS dispersion / human [95\% CI] \\
\midrule
Chile & Gemma 4-12B & 1.974 [1.883, 2.047] \\
Chile & Qwen3-32B & 1.272 [1.206, 1.336] \\
Chile & Ministral 3-14B & 1.852 [1.755, 1.932] \\
France & Gemma 4-12B & 1.290 [1.181, 1.332] \\
France & Qwen3-32B & 1.163 [1.060, 1.207] \\
France & Ministral 3-14B & 1.001 [0.914, 1.050] \\
Brazil & Gemma 4-12B & 1.238 [1.121, 1.321] \\
Brazil & Qwen3-32B & 1.136 [1.033, 1.218] \\
Brazil & Ministral 3-14B & 0.903 [0.811, 0.986] \\
\bottomrule
\end{tabular}

%% file: source_tables/table_robustness_bradley_terry_cleavage.tex
\begin{tabular}{llrrrr}
\toprule
Country & Checkpoint & Pearson $r$ [95\% CI] & Sign agreement [95\% CI] & RMS ratio [95\% CI] & Slope [95\% CI] \\
\midrule
Chile & Gemma 4-12B & -0.592 [-0.620, -0.523] & 0.333 [0.267, 0.389] & 2.033 [1.849, 2.172] & -1.204 [-1.290, -1.020] \\
Chile & Qwen3-32B & -0.442 [-0.478, -0.376] & 0.400 [0.311, 0.422] & 2.282 [2.072, 2.428] & -1.009 [-1.104, -0.820] \\
Chile & Ministral 3-14B & -0.456 [-0.491, -0.383] & 0.378 [0.322, 0.444] & 1.409 [1.274, 1.531] & -0.643 [-0.709, -0.516] \\
France & Gemma 4-12B & 0.676 [0.573, 0.709] & 0.733 [0.667, 0.792] & 0.964 [0.862, 1.018] & 0.652 [0.514, 0.697] \\
France & Qwen3-32B & 0.494 [0.347, 0.531] & 0.700 [0.600, 0.733] & 0.662 [0.603, 0.738] & 0.327 [0.224, 0.369] \\
France & Ministral 3-14B & 0.593 [0.461, 0.603] & 0.750 [0.633, 0.758] & 0.857 [0.759, 0.925] & 0.508 [0.370, 0.521] \\
Brazil & Gemma 4-12B & 0.113 [-0.009, 0.270] & 0.567 [0.478, 0.657] & 0.693 [0.626, 0.791] & 0.078 [-0.007, 0.192] \\
Brazil & Qwen3-32B & 0.410 [0.258, 0.491] & 0.672 [0.522, 0.701] & 0.579 [0.524, 0.686] & 0.237 [0.149, 0.308] \\
Brazil & Ministral 3-14B & 0.156 [0.000, 0.274] & 0.522 [0.418, 0.597] & 0.753 [0.661, 0.863] & 0.117 [0.000, 0.215] \\
\bottomrule
\end{tabular}

%% file: source_tables/table_endpoint_anchor_diagnostic.tex
\begin{tabular}{llrrrrr}
\toprule
Country & Order & Changed & $\Delta$ agree. & Cleavage $r$ & Self-pl. acc. & Var./human \\
\midrule
Chile & Original & 0.294 & +0.050 & -0.442$\rightarrow$0.409 & 0.925$\rightarrow$0.968 & 0.062$\rightarrow$0.016 \\
Chile & Inverted & 0.268 & +0.034 & -0.474$\rightarrow$0.388 & 0.816$\rightarrow$0.942 & 0.019$\rightarrow$0.004 \\
France & Original & 0.205 & +0.121 & 0.494$\rightarrow$0.828 & 0.792$\rightarrow$0.824 & 0.103$\rightarrow$0.281 \\
France & Inverted & 0.204 & -0.060 & 0.228$\rightarrow$0.601 & 0.774$\rightarrow$0.824 & 0.079$\rightarrow$0.087 \\
Brazil & Original & 0.214 & +0.132 & 0.410$\rightarrow$0.772 & 0.761$\rightarrow$0.915 & 0.031$\rightarrow$0.185 \\
Brazil & Inverted & 0.234 & -0.058 & 0.039$\rightarrow$0.599 & 0.744$\rightarrow$0.863 & 0.015$\rightarrow$0.032 \\
\bottomrule
\end{tabular}